%% file: main.tex
\documentclass[runningheads]{llncs}

\usepackage{eccv}

\usepackage{eccvabbrv}

\usepackage{graphicx}
\usepackage{booktabs}
\usepackage{comment}
\usepackage{amsmath,amssymb}
\usepackage{color}
\usepackage{url}
\usepackage{booktabs}
\usepackage{pifont}
\usepackage{multirow}
\usepackage{xspace}
\usepackage{multicol}
\usepackage{makecell}
\usepackage{epsfig}
\usepackage{subcaption}
\usepackage{array}
\usepackage[export]{adjustbox}
\usepackage[accsupp]{axessibility}  

\usepackage{pgfplots}
\pgfplotsset{width=10cm,compat=1.9}

\definecolor{Seaborn1}{HTML}{1f77b4}
\definecolor{Seaborn2}{HTML}{ff7f0e}
\definecolor{Seaborn3}{HTML}{2ca02c}
\definecolor{Seaborn4}{HTML}{d62728}
\definecolor{Seaborn5}{HTML}{9467bd}
\definecolor{Seaborn6}{HTML}{8c564b}
\definecolor{Seaborn7}{HTML}{e377c2}
\definecolor{Seaborn8}{HTML}{7f7f7f}
\definecolor{Seaborn9}{HTML}{bcbd22}
\definecolor{Seaborn10}{HTML}{17becf}

\pgfplotsset{
compat=1.9,
legend image code/.code={
        \draw[mark repeat=4,mark phase=1]
        plot coordinates {
            (0cm,0.02cm)
            (0.1cm,0.02cm)        
            (0.2cm,0.02cm)
            (0.3cm,0.02cm)      
        };%
    }
}

\usepackage[pagebackref,breaklinks,colorlinks,citecolor=eccvblue]{hyperref}

\usepackage{orcidlink}

\newcommand{\M}[1]{\mathbf{#1}}
\newcommand{\dataset}[1]{{\fontfamily{cmtt}\selectfont #1}\xspace }

\newcommand{\ETH}{\dataset{ETH3D}}
\newcommand{\ROTUNDA}{\dataset{ROTUNDA}}
\newcommand{\VITUS}{\dataset{CATHEDRAL}}

\newcommand{\Phototourism}{\dataset{Phototourism}}

\newcommand{\F}{{$\M F$}}
\newcommand{\Fk}{{$\M F \lambda$}}
\newcommand{\Fkk}{{$\M F \lambda_1\lambda_2$}}
\def\gb{Gr{\"o}bner basis\xspace}

\begin{document}

\input{main_paper}

\clearpage

\input{supplementary}

\clearpage
\bibliographystyle{splncs04}  
\bibliography{main}  

\end{document}

%% file: main_paper.tex

\title{Are Minimal Radial Distortion Solvers Necessary for Relative Pose Estimation?} 

\titlerunning{Are Minimal Radial Distortion Solvers Necessary?}

\author{Charalambos Tzamos\inst{1} \and
Viktor Kocur\inst{2} \and
Yaqing Ding\inst{1} \and
Torsten Sattler\inst{3} \and
Zuzana Kukelova\inst{1}}

\authorrunning{C.~Tzamos et al.}

\institute{Visual Recognition Group, Faculty of Electrical Engineering, Czech Technical University in Prague \\
\email{\{tzamocha, yaqing.ding, kukelzuz\}@fel.cvut.cz} \and
Faculty of Mathematics, Physics and Informatics, Comenius University Bratislava \\
\email{viktor.kocur@fmph.uniba.sk} \and
Czech Institute of Informatics, Robotics and Cybernetics, Czech Technical University in Prague \\
\email{torsten.sattler@cvut.cz}}

\maketitle

\begin{abstract} 
  Estimating the relative pose between two cameras is a fundamental step in many applications such as Structure-from-Motion. The common approach to relative pose estimation is to apply a minimal solver inside a RANSAC loop. Highly efficient solvers exist for pinhole cameras. Yet, (nearly) all cameras exhibit radial distortion. Not modeling radial distortion leads to (significantly) worse results. However, minimal radial distortion solvers are significantly more complex than pinhole solvers, both in terms of run-time and implementation efforts. This  paper compares radial distortion solvers with a simple-to-implement approach that combines an efficient pinhole solver with sampled radial distortion parameters. Extensive experiments on multiple datasets and RANSAC variants show that this simple approach performs similarly or better than the most accurate minimal distortion solvers at faster run-times while being significantly more accurate than faster non-minimal solvers. We clearly show that complex radial distortion solvers are not necessary in practice. Code and benchmark are available at {\url{https://github.com/kocurvik/rd}}.     
  \keywords{Relative Pose Estimation \and Radial Distortion  \and RANSAC}
\end{abstract}

\section{Introduction}
\label{sec:introduction}
\noindent Estimating the relative pose of two cameras, \ie, estimating the relative rotation, translation, and potentially internal calibration parameters of both cameras, is a fundamental problem in computer vision. 
Relative pose solvers are core components of Structure-from-Motion (SfM)~\cite{wu2011visualsfm,Schoenberger2016CVPR} and localization pipelines~\cite{sattler2016efficient,svarm2016city,sarlin2019coarse} and play an important role in robotics~\cite{mur2015orb,mur2017orb} and autonomous driving~\cite{scaramuzza2011visual}.


A predominant way to estimate the relative pose of two cameras 
is based on 2D-2D point correspondences between the two images. Due to noise and the presence of outliers, 
robust estimation algorithms, such as RANdom SAmple Consensus (RANSAC)~\cite{Fischler-Bolles-ACM-1981}, or its more modern variants~\cite{raguram2013usac,barath2017graph}, are used for the estimation.
Inside RANSAC two different steps are performed: 
(i) Estimating the camera geometry from a (small or minimal) sample of correspondences and classifying all correspondences into inliers and outliers \wrt the obtained camera model.
(ii) Local optimization (LO) of the camera model parameters 
on (a subset of) the inliers to better account for noise in the 2D point positions~\cite{Chum-2003,lebeda2012fixing}.

The main objective of the first step is to obtain a  
camera geometry estimate and the subset of correspondences  consistent with it. 
Small samples 
are preferable since the 
number of RANSAC iterations, and thus the run-time, depends exponentially on the number of correspondences required for model estimation. 
Solvers that estimate the camera geometry using a minimal number of correspondences and using all available polynomial constraints are known as minimal solvers.
The most commonly used minimal solvers for relative pose estimation 
are the well-known 5-point solver~\cite{nister2004efficient} 
for calibrated cameras and the 7-point solver~\cite{hartley2003multiple} for uncalibrated cameras. 
Both 
are highly efficient. 

Minimal solvers produce estimates that perfectly fit the correspondences in the minimal sample. 
In practice, the 2D point correspondences are noisy and the noise in the 2D coordinates propagates to the estimates. 
The goal of the second step, \ie, LO inside RANSAC, is to reduce the impact of measurement noise on pose accuracy~\cite{lebeda2012fixing}. 
Commonly, a non-minimal solver that fits model parameters to a larger-than-minimal sample is used~\cite{Chum-2003}, or a robust cost function that optimizes model parameters on all inliers is minimized~\cite{lebeda2012fixing}.
The most common 
non-minimal solver for relative pose estimation is the linear 8-point solver~\cite{hartley2003multiple}. 

All previously mentioned solvers, \ie, the 5-point, 7-point, and 8-point solvers, are widely used in SfM pipelines and other applications. 
They are based on the pinhole camera model. 
Yet, virtually all cameras exhibit some amount of radial distortion. 
Ignoring the distortion,
even for standard consumer cameras, can lead to 
errors in 3D reconstruction~\cite{fitzgibbon2001simultaneous}, camera calibration accuracy, \etc 

There are several ways to deal with radial distortion: (1) Ignore radial distortion estimation during RANSAC and model 
it only in a post-processing step, \eg, during bundle adjustment in SfM~\cite{snavely2008modeling,schonberger2016structure}. 
(2) Ignore radial distortion in the first RANSAC step but take it into account in the second step 
%
(LO), \eg,  
by using a non-minimal solver that estimates radial distortion parameters or by modeling distortion 
when minimizing a robust cost function. 
(3) Already estimate the radial distortion in the first RANSAC step (and refine it during LO). 
%
Approach (3) is the most principled solution as it enables taking radial distortion into account during inlier counting.  
Ignoring radial distortion inside the solver typically leads to identifying only the subset of the inliers that is less affected by the distortion.\footnote{A point in one image maps to an epipolar \textit{curve} in a radially distorted second image. Ignoring radial distortion thus means approximating this curve by a line. 
The approximation is 
only 
good around the center of 
strongly distorted images.} 
As a result of only containing points that are only mildly affected by distortion, this inlier set often does not contain enough information to accurately estimate the distortion parameters. 
Thus, 
approaches (1) and (2), which operate on the inliers identified beforehand,  
are likely to fall into local minima, without recovering
 correct distortion and camera parameters. 

Radial distortion modeling is a mathematically challenging task, and even the simplest one-parameter radial distortion model leads to complex polynomial equations when incorporated into relative pose solvers~\cite{fitzgibbon2001simultaneous, kukelova2007minimal}. Thus, algorithms for estimating
epipolar geometry for cameras with radial distortion started appearing only after introducing efficient algebraic polynomial solvers into the computer vision community~\cite{fitzgibbon2001simultaneous,barreto2005fundamental,jin2008three,byrod2009minimal,kukelova2010fast}. 
With improvements in methods for generating efficient polynomial solvers, also minimal radial distortion solvers are improving their efficiency and stability. 
However, compared to solvers for the pinhole camera model, most of these solvers are still orders of magnitude slower,  \eg, 
 the fastest 9-point solver for different distortions runs 210$\mu s$~\cite{Oskarsson_2021_CVPR}, and the 6-point solver with unknown common radial distortion for calibrated cameras runs 1.18$ms$. 
This is significantly slower than the 5-point and 7-point pinhole camera solvers that run in less than 6$\mu s$. 
Moreover, since these solvers estimate more unknown parameters, they need to sample more points inside RANSAC.\footnote{Usually one more point for cameras with a common unknown radial distortion and two more points for cameras with different radial distortions.} They also return more potential solutions to the camera model.
Thus, even though radial distortion solvers estimate models that better fit the data, they may require more RANSAC iterations and longer RANSAC run-times.  

Radial distortion solvers are not only slower but also more complex to implement. 
At the same time, many of the existing minimal radial distortion solvers do not have a publicly available implementation. 
Even though the papers that present novel radial distortion solvers show advantages of these solvers on real data, they usually focus on presenting novel parameterizations and solution strategies and their numerical stability on synthetic data. Real experiments are mostly limited to small datasets (\eg, a single scene), simpler variants of RANSAC,  and qualitative instead of quantitative results.   
The above-mentioned facts are most likely 
the reasons why (minimal) radial distortion solvers are not often used in practice. Instead, it is common to use either approach (1) or (2)~\cite{schonberger2016structure,snavely2008modeling}. 
Naturally, this leads to the question whether 
(minimal) radial distortion solvers are actually necessary 
in practical applications. 

The goal of this paper is to answer this question. 
To this end, 
we introduce a new sampling-based strategy that combines the efficient 7-point pinhole camera solver with a sampled undistortion parameter: 
In each RANSAC iteration, we run the solver potentially several times (1-3x) with different (but fixed) undistortion parameters.
The paper makes the following contributions:
(\textbf{i}) We extensively evaluate different approaches for relative pose estimation under radial distortion on multiple datasets, under different scenarios, and using different RANSAC variants with different LO strategies. 
We are not aware of any such practical evaluation of radial distortion solvers in the literature. 
(\textbf{ii}) We show that the simple sampling-based approach, which is easy to implement, performs similar or better than the most accurate minimal radial distortion solvers at significantly faster run-times, and significantly outperforms faster radial distortion solvers. 
We thus show that dedicated minimal radial distortion solvers for the relative pose problem are not necessary in practice.  
(\textbf{iii}) We create a new benchmark, consisting of two scenes, containing images taken with different cameras with multiple different distortions. 
(\textbf{iv}) Code and dataset are available at \url{https://github.com/kocurvik/rd}.

\section{Related Work}
The literature studies three groups of radial distortion relative pose problems: Two cameras with equal and unknown radial distortion; 
two cameras where only one has unknown 
distortion; 
two cameras with different and unknown distortion.

\noindent\textbf{Equal and unknown radial distortion:} Fitzgibbon~\cite{fitzgibbon2001simultaneous} 
introduced an one-parameter division model for modeling distortion and an algorithm for estimating the fundamental matrix with equal and unknown radial distortion using this model.
This algorithm does not use 
the singularity constraint on the fundamental matrix, necessitating 9 point correspondences instead of the minimal 8. 
This approach 
simplifies the problem into a standard quadratic eigenvalue problem with 
up to 10 solutions. 
The first minimal solution for epipolar geometry estimation with 
the one-parameter division model using 8 point correspondences 
was proposed by Kukelova~\etal~\cite{kukelova2007minimal}, 
using the \gb method~\cite{cox2005using} to solve a system of polynomial equations. This solver has been improved by 
using an automatic generator of
\gb solvers~\cite{kukelova2008automatic}, performing Gauss-Jordan (G-J)
elimination of a 32×48 matrix and eigenvalue decomposition of
a 16×16 matrix, which has up to 16 solutions. 
In~\cite{jiang2015minimal}, Jiang~\etal used 7 point correspondences to solve the problem of essential matrix estimation for two cameras with equal and unknown focal length and radial distortion. 
This problem results in a complex system of polynomial equations and a large solver that performs the LU decomposition of an 886×1011 matrix and computes the eigenvalues of a 68×68 matrix. 
Thus, this solver is too 
time-consuming 
for practical 
applications. 

\noindent\textbf{One unknown radial distortion:} In~\cite{kuang2014minimal}, Kuang~\etal studied three minimal cases for relative pose estimation with a
single unknown radial distortion based on the \gb method: 8-point fundamental matrix and radial distortion; 7-point essential matrix, focal length and radial distortion; 6-point essential matrix and radial distortion. However, these solvers assume one of the two cameras has known or no radial distortion, which may not be realistic 
in practice.

\noindent\textbf{Different and unknown radial distortions:} All of the above mentioned algorithms estimate only one radial distortion parameter for one or both cameras. 
In practice, \eg, using images 
downloaded from the Internet, two cameras can 
have different and unknown radial distortions. The problem of fundamental matrix estimation with different and unknown radial distortions, $\M F \lambda_1 \lambda_2$, was first studied by Barreto~\etal~\cite{barreto2005fundamental}, 
proposing 
a non-minimal linear algorithm 
using 15 point correspondences (F15). 
The minimal 9-point case (F9) for this problem was studied in~\cite{kukelova2007two,byrod2008fast,kukelova2008automatic,kukelova2010fast}. 
The solver from~\cite{byrod2008fast} (F9) 
performs LU decomposition of a 393×389 matrix, SVD decomposition of a 69×69 matrix, and 
eigenvalue computation of a 24×24 matrix. 
In~\cite{kukelova2008automatic}, a faster version based on a \gb ($\rm F9_{\rm A}$) was proposed. 
It performs G-J elimination of a 179×203 matrix and eigenvalue decomposition of a 24×24 matrix. 
However, this solver is slightly less stable than F9, and still too slow for real-time applications. 
Kukelova~\etal~\cite{kukelova2010fast} suggested an efficient, non-minimal solver using 12 point correspondences (F12) that 
generates up to four real solutions. 
However, 
this algorithm 
is more sensitive to noise than the minimal 
$\rm F9_{\rm A}$~\cite{kukelova2010fast}. 
Balancing efficiency and  noise sensitivity, Kukelova~\etal~\cite{kukelova2015efficient} proposed a 10-point solver 
that is much faster than the minimal 9-point solver and more robust to image noise than 
the 12-point solver.

Recently, Oskarsson~\cite{Oskarsson_2021_CVPR} presented a unified formulation for relative pose problems involving radial distortion and proposed more efficient minimal solvers for all different configurations. 
While some of the proposed solvers are already quite efficient, \eg, the 8-point solver for uncalibrated cameras with common radial distortion, others, like the 9-point solver for different distortions, are still too slow and/or numerically unstable to be useful in practice.

\noindent\textbf{Parameter sampling:} Instead of jointly estimating the absolute camera pose and the focal length of an uncalibrated camera, Sattler \etal~\cite{Sattler2014ECCV} proposed a RANSAC variant that combines parameter sampling and parameter estimation. 
In each RANSAC iteration, they first randomly sample a focal length value and then estimate the pose of the now-calibrated camera. 
The probability distribution over the focal length values is then updated based on the number of inliers of the estimated pose. 
We propose a simpler sampling-based strategy for relative pose estimation that uses a small fixed set of undistortion parameters.  
In contrast to~\cite{Sattler2014ECCV}, our approach can  easily be applied to 2D sampling problems, \eg, two different and unknown distortion parameters.

\section{Radial Distortion Solvers} \label{sec:rd_solvers}
\noindent\textbf{Background:} 
A pair of corresponding distorted image points $\mathbf{x}_i \leftrightarrow \mathbf{x}_i^\prime$,  detected in two images, is related by the epipolar constraint
\begin{equation}
    u(\mathbf{x}_i, \mathbf{\Lambda})^\top \M F u(\mathbf{x}_i^\prime, \mathbf{\Lambda}^\prime) = 0 \enspace ,\label{eq:01}
\end{equation}
where $\mathbf{x}_i, \mathbf{x}_i^\prime \in \mathbb{P}^2$, $\M F$ is the fundamental matrix encoding the relative pose and the internal calibrations of the two cameras, and 
$u:\mathbb{P}^2 \times \mathbb{R}^{n} \rightarrow \mathbb{P}^2$ denotes an undistortion function, which is a function of the distorted image point $\mathbf{x}_i$ and $n$ undistortion parameters $\mathbf{\Lambda} \in \mathbb{R}^{n}$.

In this paper, we model the undistortion function using the one-parameter division model~\cite{fitzgibbon2001simultaneous}. 
In this model, given an observed radially distorted point with homogeneous coordinates $\mathbf{x} = [x_d, y_d, 1]^\top$, and the undistortion parameter $\lambda \in \mathbb{R}$, the 
undistorted image point is given as 
\begin{equation}
    u(\mathbf{x}, \lambda) = [x_d, y_d, 1 + \lambda (x_d^2 + y_d^2)]^\top \enspace ,
    \label{eq:02}
\end{equation}

\noindent assuming that the distortion center is in the image center. This model is very commonly used in practice due to its simplicity, efficiency, and robustness, since it can adequately capture 
even large distortions of wide-angle lenses. It is incorporated in almost all minimal and non-minimal radial distortion solvers.

\noindent \textbf{Radial distortion solvers:} The goal of this paper is not to introduce novel minimal or non-minimal radial distortion solvers, but to study the performance of the existing solvers under different conditions. 
We study the two most practical scenarios of two uncalibrated cameras with unknown (i) equal and (ii) different radial distortions. 
We denote these problems as (i) the \Fk \ and (ii) the \Fkk \ problems.\footnote{Instead of $\lambda$ and $\lambda^{\prime}$ as used in~\eqref{eq:01}, we use $\lambda_1$ and $\lambda_2$ for better readability.} Next, we briefly describe the radial distortion solvers for these two problems studied 
in this paper, as well as some improvements to these solvers.

\noindent \Fk \ : Assuming equal unknown distortion modeled using the one-parameter division model~\eqref{eq:02}, the relative pose problem for uncalibrated cameras has 8 degrees of freedom (DoF).  For this problem, we consider the following solvers: 
\begin{itemize}
    \item 8pt \Fk \ : Among all minimal 8pt solvers~\cite{kukelova2007minimal,kukelova2008automatic,larsson2017efficient,Oskarsson_2021_CVPR}, the solver from~\cite{Oskarsson_2021_CVPR} is the most efficient. It formulates the elements of $\M{F}$ as functions of the distortion parameter $\lambda$, and obtains an univariate polynomial in $\lambda$ of degree 16, which can be solved using the Sturm sequences, with up to 16 solutions. 

    \item 9pt \Fk \ : By ignoring the $\det(\M{F}) = 0$ constraint, the \Fk \ 
 problem can be solved using nine point correspondences. \cite{fitzgibbon2001simultaneous} solves the nine equations~\eqref{eq:01} by converting them into a polynomial eigenvalue problem.
 While~\cite{fitzgibbon2001simultaneous} was able to remove several spurious solutions by transforming the original eigenvalue problem of size $18 \times 18$ into a problem of size $10 \times 10$,~\cite{fitzgibbon2001simultaneous} also observed that 4 of the 10 solutions of this system are imaginary. In this paper, we propose a modification of the solver proposed in~\cite{fitzgibbon2001simultaneous}, in which we directly remove 4 imaginary solutions, resulting in a more efficient solver that needs to find the eigenvalues of a smaller $6\times 6$ matrix. To remove these 4 imaginary solutions, we use the structure of matrices that appear in the polynomial eigenvalue formulation of this problem and the method proposed in~\cite{kukelova2012polynomial}. For more details, see the Supplementary Material (SM).
\end{itemize}

\noindent \Fkk \ : For the case of 
different unknown radial distoritons, we have 9 DoF. For this problem, we consider the following solvers:
\begin{itemize}
    \item  9pt \Fkk \ : Equations for cameras with different unknown distortions are more complex than for the equal distortion case. 
    In this case the system of equations has 24 solutions and the fastest Gr\"{o}bner basis solver from~\cite{larsson2017efficient}, which returns 24 solutions, performs elimination of a large matrix of size $84 \times 117$ followed by the eigendecomposition of a $24 \times 24$ matrix. The recently published parameterization of this problem in~\cite{Oskarsson_2021_CVPR} performs elimination of a smaller matrix of size $51 \times 99$ followed by the  eigendecomposition of a $48 \times 48$ matrix. The solver returns up to 48 solutions. 
    However, it is still faster than the solver from~\cite{larsson2017efficient}. Thus, in our experiments, we use the solver from~\cite{Oskarsson_2021_CVPR}. 
    \item  10pt \Fkk \ : In~\cite{kukelova2015efficient} it was shown that in many scenarios inside RANSAC it is preferable to sample 10 instead of 9 points and run the more efficient 10pt solver. In~\cite{kukelova2015efficient}, the authors proposed several variants of the 10pt solver. In this paper, we use the variant based on a \gb, made available by the authors. The 10pt solvers cannot be easily modified to work with more than 10 points, and thus we use this solver only in the first step of RANSAC, \ie, instead of the minimal solver, and not in the LO step. 
    \item  12pt \Fkk \ : Similarly to the 9pt \Fk \ solver, 
    we can ignore the $\det(\M{F})=0$ constraint in the case of different radial distortions and formulate the problem as a polynomial eigenvalue problem. Such a solver using 12 point correspondences was proposed in~\cite{kukelova2010fast}. The solver performs the eigendecomposition of a $12 \times 12$  matrix. 
    Yet, as observed in~\cite{kukelova2010fast}, it has eight infinite eigenvalues, and at most four finite real solutions. In this paper, we 
    modify this solver. We remove all infinite eigenvalues in advance. Thus, in our new variant of the solver, we only need to find the eigenvalues of a $4\times 4$ matrix. For more details, see the SM. This solver, as well as the 9pt \Fk \ solver, can be easily extended to more points and used in the LO step of RANSAC.
\end{itemize}

While the non-mininmal 9pt \Fk, 10pt \Fkk, and 12pt \Fkk{} solvers are reasonably efficient, the minimal 8pt \Fk \ and especially the 9pt \Fkk{} solver are significantly slower than the minimal 7pt standard pinhole camera solver. Moreover, the minimal radial distortion solvers return more solutions, \ie, 16, 24, or even 48 compared to the 3 solutions of the 7pt solver. More solutions lead to reduced efficiency, since within a RANSAC framework each solution has to be evaluated. 
This, together with the fact that the radial distortion solvers sample more points and solve significantly more complex equations,  motivates a common strategy in which in the first step of RANSAC, the 7pt perspective solver is applied and the radial distortion is modeled only in the LO step of RANSAC, \eg, by using the 9pt \Fk \ or 12pt \Fkk \ solvers.

However, as mentioned in Sec.~\ref{sec:introduction}, 
for images with larger distortion, the standard perspective camera model without distortion may not properly model the data and may thus not return a large-enough subset of the true inliers 
and/or an accurate-enough initial pose estimate. 
Yet, small changes in the undistortion parameter $\lambda$ in~\eqref{eq:02}, in general, do not result in large changes in the projection of points into the image. 
For an undistortion parameter $\lambda$ that is reasonably close to the true parameter $\lambda_\text{true}$, 
we can thus expect that 
applying the 7pt solver on 2D point positions that were undistorted using $\lambda$ can 
result in sufficiently-large inlier sets and  sufficiently-accurate initial poses 
that will lead to good estimates in the LO step.
This motivates the sampling strategy that we describe next. 

\subsection{Sampling Distortions}
\label{sec:rd_solvers:sampling}
In this paper, in addition to the existing minimal radial distortion solvers that are used in the first step of RANSAC, we propose a simple sampling-based strategy: 
In each iteration of RANSAC, it runs the standard 7-point pinhole camera solver on image points undistorted with a fixed radial undistortion parameter sampled from an interval of feasible undistortion parameters.
In this method, we use the facts that the 7pt solver is significantly faster than the minimal radial distortion solvers, and returns significantly fewer solutions that need to be tested inside RANSAC. Thus, even running the 7pt solver several times with different fixed undistortion parameters in each RANSAC iteration may lead to a higher efficiency compared to running the 8pt \Fk \ or 9pt \Fkk \ radial distortion solvers.

The number $k$ of runs of the 7pt solver in each iteration, and the values $\M U_i = \{\hat{\lambda}_i^1, \hat{\lambda}_i^2, ..., \hat{\lambda}_i^k\}$, which are used to undistort points in the two cameras $i=1,2$, can differ depending on the application and input data.
In our experiments, we test three variants of the sampling solver: (1) $\M U_1 = \M U_2 = \{ 0 \}$,
which represents the above mentioned standard baseline that assumes no distortion in the first step of RANSAC; 
(2) $\M U_i = \{ \hat{\lambda}_i \}$, $i=1,2$
where we run the 7pt solver only once for one fixed value of $\hat{\lambda}_i \neq 0$. This can represent a scenario where we have prior knowledge that our images have visible distortion. In our experiments, 
we test a version with $\hat{\lambda}_i$ that represents medium distortion and
can potentially, after LO, provide good results even for cameras with small or large distortion; 
(3)  $\M U_i = \{\hat{\lambda}_i^1, \hat{\lambda}_i^2, \hat{\lambda}_i^3\}$,  where 
we undistort points in each camera with three different fixed parameters that represent, \eg, small, 
medium, 
and large distortion. This setup is, \eg, useful in scenarios where we have images from the "wild" (\eg, the Internet) that can have a wide variety of different distortions. Note that in this case, if we assume cameras with different distortions, we need to run the 7pt solver nine times. Still, this is more efficient than using the 
9pt \Fkk \ solver.

\section{Experiments}

\noindent\textbf{Datasets.} 
We evaluate the different approaches for radial distortion relative pose estimation on three datasets: 
\Phototourism ~\cite{IMC2020}, 
\ETH \cite{Schops_2017_CVPR}, 
and our new benchmark, 
each covering different scenarios. 
\Phototourism 
contains images of popular landmarks downloaded from Internet photo collections, 
taken with 
a wide variety of cameras. 
\Phototourism provides ground truth camera poses obtained via SfM and undistorted images with known intrinsics. 
The dataset is 
widely used for evaluating local features and feature matching~\cite{IMC2020}. 
For our experiments, we used 13 scenes with 5,000 image pairs per scene. 
It covers indoor and outdoor scenes captured with a DSLR
camera and a multi-camera rig containing synchronized cameras with varying field-of-views. 
Unlike 
\Phototourism, 
which contains images taken at different points in time and under varying lighting conditions, the images were taken in a short period of time for each scene. 
Ground truth poses were obtained by aligning the images to high-precision laser scans. 
\ETH provides undistorted images together with their intrinsic calibration. 
We use 2,037 image pairs from 12 \ETH scenes. 

Existing datasets~\cite{schubert2018vidataset,Schops_2017_CVPR} containing radially distorted images 
mostly involve only one or two different camera lenses with little 
variation in the distortion parameters. 
Testing the sampling-based strategy on such images could be biased, as it would have been as good as the distance of the used sampled value from the one/two ground truth values. 
We thus created a new benchmark with a higher variation in the distortion parameters,  
consisting of two scenes:  \ROTUNDA and {\fontfamily{cmtt}\selectfont CATHERAL}. 
For both scenes, we build upon previously recorded images~\cite{kukelova2015efficient,Sattler_2019_CVPR}, taken by GoPro cameras and kindly provided by the authors. 
We added 
images 
from phone cameras with very small lens distortion, images taken by action cameras (GoPro, Insta360 Ace Pro) with different distortions, 
and a few images downloaded from Flickr. 
We created ground truth for the benchmark (poses and the parameters of the one-parameter division model) using the RealityCapture SfM software~\cite{RealityCapture}. 
We use 3,424 image pairs with $\lambda_1 \neq \lambda_2$ and 1,795 image pairs with $\lambda_1 = \lambda_2$ for \ROTUNDA and 10k 
sampled image pairs for both $\lambda_1 \neq \lambda_2$ and $\lambda_1 = \lambda_2$ for {\fontfamily{cmtt}\selectfont CATHERAL}. 
The scenes are visualized in the SM. 

\noindent \textbf{Evaluation measures.} Following~\cite{IMC2020}, given the ground truth and the estimated relative pose, we measure the rotation error and the translation error. 
The rotation error is defined as the angle of the rotation matrix aligning the estimated with the ground truth rotation matrix. 
The translation error is defined as the angle between the estimated and the ground truth translation vector. 
Finally, the pose error is defined as the maximum of the rotation error and the translation error. To obtain poses from the estimated fundamental matrices we use ground truth intrinsics~\cite{IMC2020}. 
We report the average (AVG) and median (MED) pose errors, as well as the area under the recall curve (AUC) for different thresholds on the pose error. 
In addition, we measure the distortion error $\epsilon(\lambda)$ as the absolute distance between ground truth and estimated distortion parameters. For the \Fkk \ problem, we measure the distortion error as  $0.5 \cdot (\epsilon(\lambda_1)+\epsilon(\lambda_2))$. 

\noindent \textbf{Experimental setup.} 
We obtain point correspondences between images by matching Superpoint~\cite{detone2018superpoint} 
features with the LightGlue (LG)~\cite{lindenberger2023lightglue} 
matcher.  
Experiments with SIFT~\cite{lowe2004distinctive} and LG can be found in the SM. 

In the first step of RANSAC, we use the minimal 7pt \F, 8pt \Fk, 9pt \Fkk, and the non-minmal 9pt \Fk, and  10pt \Fkk{} solvers (\cf Sec.~\ref{sec:rd_solvers}), as well as the sampling strategies that combine the 7pt solver with a set of pre-defined undistortion parameters (\cf Sec.~\ref{sec:rd_solvers:sampling}). 
We denote the latter by appending the list of parameters, \eg, $\{0, -0.6, -1.2\}$, to the solver configuration. 
We integrate the solvers and strategies 
into two modern RANSAC implementations: 
Graph-Cut RANSAC (GC-RANSAC)~\cite{barath2017graph}, and Poselib~\cite{PoseLib}. 
We chose these two variants as they use different approaches for Local Optimization (LO). 
GC-RANSAC uses a non-minimal solver for fitting the relative pose and distortion parameter(s) to a larger-than-minimal sample when doing final pose polishing on all inliers and in the local optimization step. 
We use the refined 9-point solver (9pt $\M F \lambda$) for the equal and unknown radial distortion case ($\lambda_1\,=\,\lambda_2$), and the refined 12-point solver (12pt $\M F \lambda_1\lambda_2$) for the different and unknown radial distortion case ($\lambda_1\,\not=\,\lambda_2$). 
In contrast, the LO step in PoseLib relies on Levenberg-Marquardt (LM) optimization of the truncated Tangent Sampson Error~\cite{Terekhov_2023_ICCV}, starting from the estimate provided by the minimal solver / sampling strategy. 
The pose returned by RANSAC is further polished 
by LM optimization over all inliers. 
For Poselib we also include experiments with a version
corresponding to Approach (1) mentioned in Sec.~\ref{sec:introduction}, \ie, modelling the radial distortion only in the final optimization. We denote this approach as 7pt \F \ with \F + \Fk~and~\F + \Fkk \ for 
refinement.
To determine which points are inliers we use the Tangent Sampson Error~\cite{Terekhov_2023_ICCV} with a fixed threshold (1 px in GC-RANSAC, 
3 px in Poselib). 
Using the Tangent Sampson Error is important since the standard Sampson Error in undistorted images leads to a radial bias in the optimization~\cite{Terekhov_2023_ICCV}.

We use normalized image coordinates from the range 
$[-0.5, 0.5]^2$, 
obtained by subtracting the image center 
and dividing by the length of the longer image side. 
For this normalization, the undistortion parameter should be greater than $-2$, as otherwise the distortion would mirror the image. 
In RANSAC, we discard models with radial distortion 
outside of the plausible range $\left[-2.0, 0.5\right]$.

\begin{figure*}[t]
    \centering
    \begin{subfigure}[t]{0.45\textwidth}
	    \includegraphics[width=1.0\textwidth]{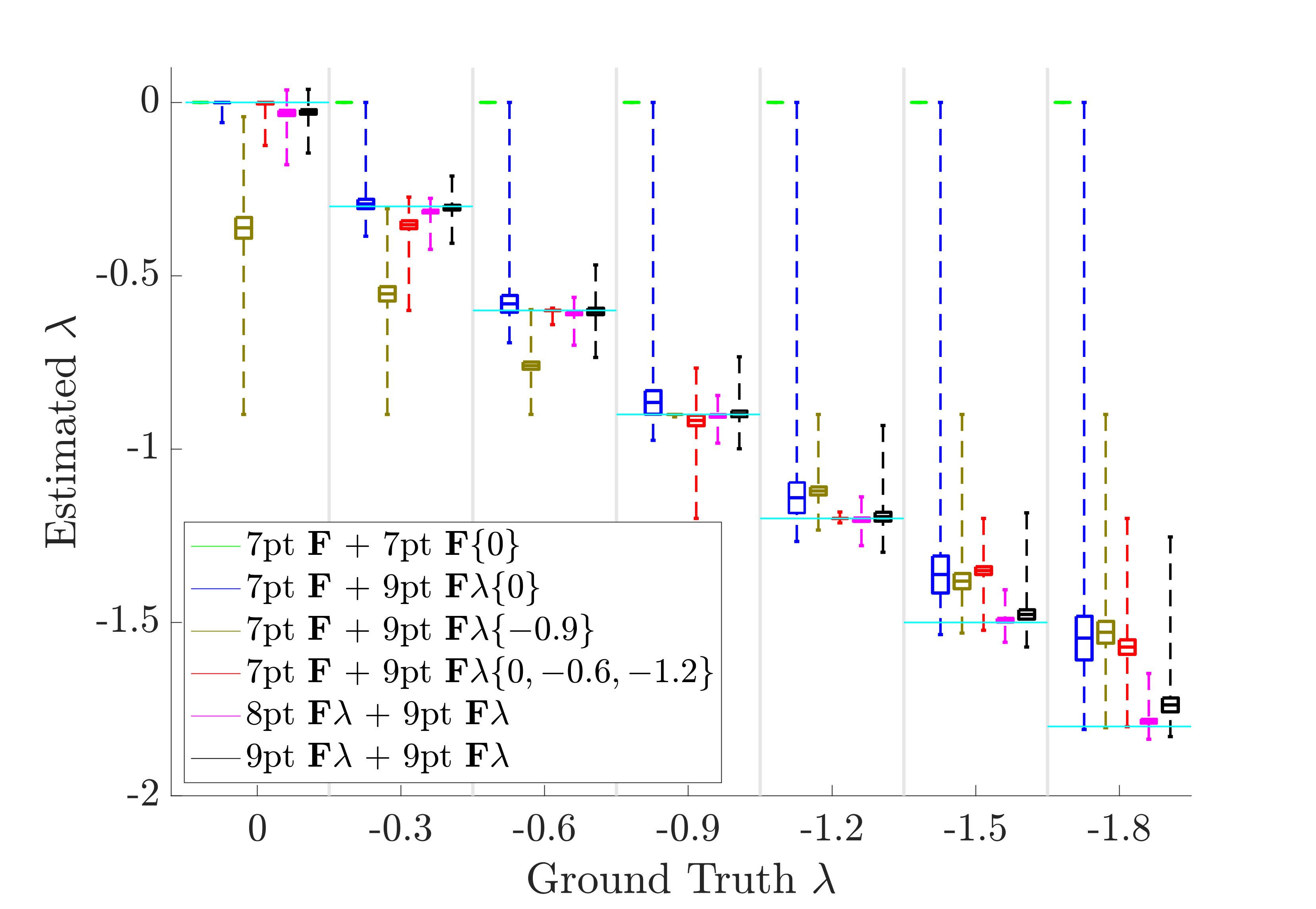}
	\end{subfigure}
    \begin{subfigure}[t]{0.45\textwidth}
	    \includegraphics[width=1.0\textwidth]{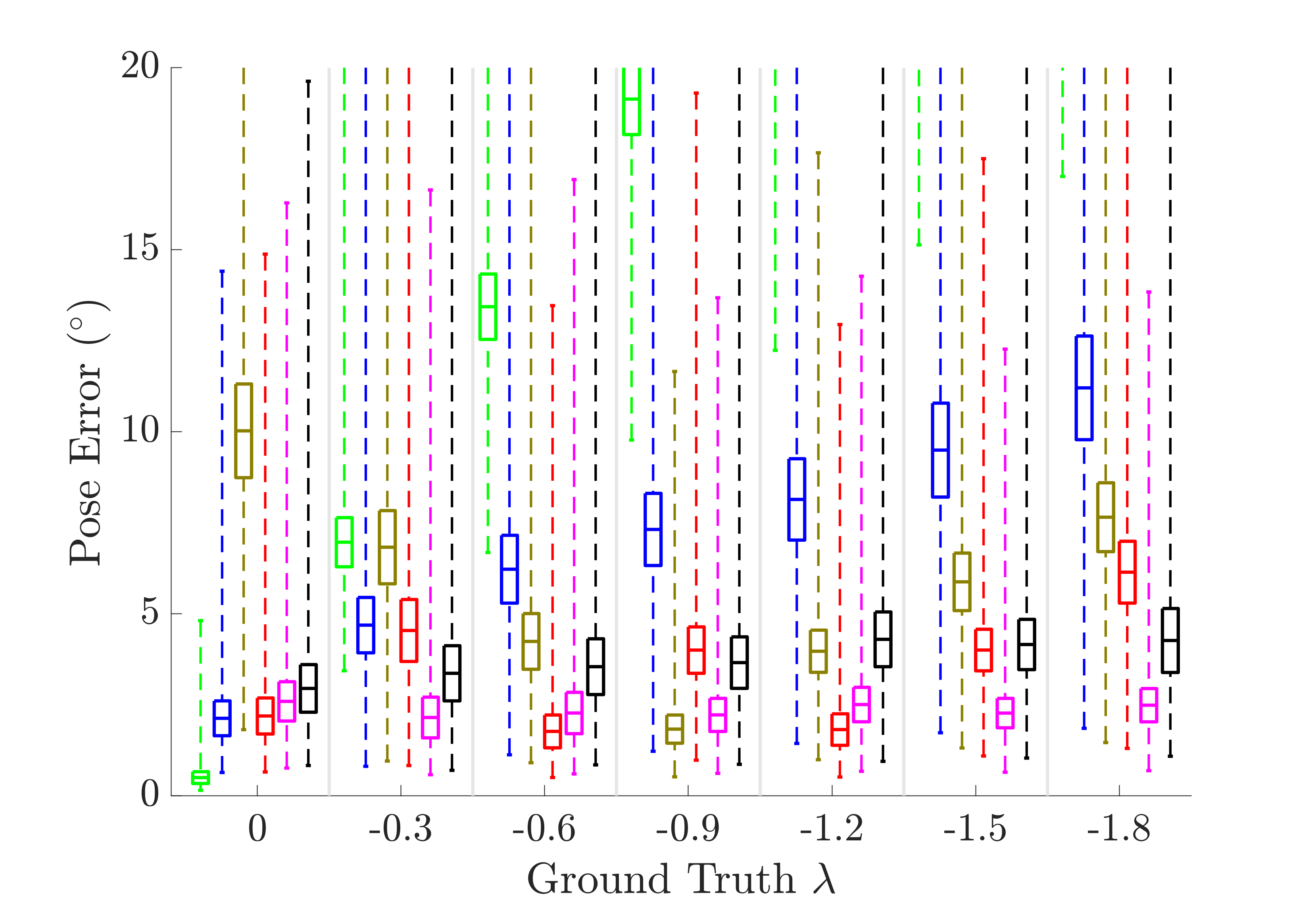}
	\end{subfigure}
    \caption{\textbf{Robustness to varying levels of equal distortion:} (Left) The estimated undistortion parameter and  (Right) the pose error. 
    An x-axis value of $-0.3$ means that we apply a distortion that can be undistorted with the undistortion parameter $-0.3$. The cyan line indicates the ground truth $\lambda$ value on the y-axis.}
    \label{fig:eth3d_box}
\end{figure*}

\noindent \textbf{Robustness to varying levels of distortion.} 
In the first experiment, we study the robustness of different approaches \wrt increasing lens distortion. 
To this end, we distort the positions of the local features extracted from the undistorted images provided by the \ETH dataset.\footnote{
In contrast to extracting feaures from images distorted with varying levels of distortion, this approach 
isolates the effect of distortion on the different strategies since 
the same feature matches are used for each distortion level.} 
We apply the same amount of distortion to both images in each image pair using the inverse of ~\eqref{eq:02} (see the SM for the closed-form formula for the inverse). 
The added distortion ranges from 0 to the distortion that corresponds to the undistortion parameter $\lambda = -1.8$. 
This represents scenarios from no/small to significant distortion. For each 
distortion, we run GC-RANSAC on all image pairs with different combinations of minimal and non-minimal solvers, including the three versions of the 7pt solver with sampled 
parameters: (1) $\M U = \{0\}$, (2) $\M U = \{-0.9\}$, and (3) $\M U = \{0, -0.6, -1.2\}$. 
Fig.~\ref{fig:eth3d_box} shows 
the estimated undistortion parameter $\lambda$ (left) and the pose error (right). 
The rotation and translation error, as well as results for one scene from the \Phototourism dataset are in the SM. 
Naturally, estimating the distortion already in the first step of RANSAC (8pt \Fk + 9pt \Fk \ and 9pt \Fk + 9pt \Fk) leads to the best results. Still, sampling-based strategies followed by LO with 9pt \Fk \ 
can perform similarly well if 
they use samples 
close enough to the true level of distortion. 
For three samples, 
slightly larger errors were observed only for larger distortion values 
($\lambda \leq -1.5$). 
However, this can be easily eliminated by adding one more sample 
and running the 7pt solver four times.

\begin{table*}[ht!]
    \centering
    \caption{\textbf{Prior knowledge about cameras}: results on all scenes of the \Phototourism dataset, using Poselib RANSAC. 
    The table shows the average and median pose error 
    in degrees; the Area Under Recall Curve (AUC) at 5$^\circ$, 10$^\circ$, and 20$^\circ$ thresholds; the average and median absolute error of the distortion parameter; and the average runtime of RANSAC. We highlight the \textbf{best} and \underline{second-best} results.}
    \setlength{\tabcolsep}{4.8pt}
    \resizebox{1.0\linewidth}{!}{
\begin{tabular}{ c | r r c | c c c c c | c c | c}
    \toprule
    & & & & \multicolumn{8}{c}{Poselib RANSAC - \Phototourism{}  - \textit{Scenario A} \emph{Wild}} \\
    \midrule
    & Minimal & Refinement & Sample & AVG $(^\circ)$ $\downarrow$ & MED $(^\circ)$ $\downarrow$ & AUC@5 $\uparrow$ & @10 & @20 & AVG $\epsilon(\lambda)$ $\downarrow$ & MED $\epsilon(\lambda)$ $\downarrow$ & Time (ms) $\downarrow$ \\
    \midrule
    \multirow{10}{*}{\rotatebox[origin=c]{90}{$\lambda_1 = \lambda_2$}}     & 7pt \F & \F & 0 & 29.45 & 15.46 & 0.14 & 0.24 & 0.37 & 0.87 & 0.86 & \phantom{1}\textbf{11.85} \\
    & 7pt \F & \Fk & $0$ & 11.43 & \phantom{1}3.32 & \underline{0.44} & \underline{0.57} & 0.69 & \underline{0.14} & \textbf{0.05} & \phantom{1}43.07 \\
    & 7pt \F & \Fkk & $0$ & 14.48 & \phantom{1}4.04 & 0.40 & 0.53 & 0.65 & 0.57 & \underline{0.11} & \phantom{1}45.54 \\
    & 7pt \F & \Fk & $\{0.0, -0.6, -1.2\}$ & \textbf{10.78} & \phantom{1}\textbf{3.16} & \textbf{0.45} & \textbf{0.58} & \textbf{0.71} & \textbf{0.13} & \textbf{0.05} & \phantom{1}66.33 \\ 
    & 7pt \F & \Fkk & $\{0.0, -0.6, -1.2\}$ & 13.23 & \phantom{1}3.66 & 0.42 & 0.55 & 0.68 & 0.51 & \underline{0.11} & 142.00 \\
    \cmidrule{2-12}
    & 8pt \Fk & \Fk & \ding{55} & \underline{10.86} & \phantom{1}\underline{3.18} & \textbf{0.45} & \textbf{0.58} & \underline{0.70} & \textbf{0.13} & \textbf{0.05} & \phantom{1}85.66 \\
    & 9pt \Fk & \Fk & \ding{55} & 11.98 & \phantom{1}3.39 & \underline{0.44} & 0.56 & 0.68 & 0.15 & \textbf{0.05} & \phantom{1}42.66 \\
    \cmidrule{2-12}
    & 9pt \Fkk & \Fkk & \ding{55} & 15.77 & \phantom{1}4.41 & 0.38 & 0.50 & 0.63 & 0.50 & 0.12 & 161.83 \\
    & 10pt \Fkk & \Fkk & \ding{55} & 14.94 & \phantom{1}4.19 & 0.39 & 0.51 & 0.64 & 0.44 & \underline{0.11} & \phantom{1}\underline{31.38} \\
    \midrule
    \midrule
    \multirow{5}{*}{\rotatebox[origin=c]{90}{$\lambda_1 \neq \lambda_2$}}     & 7pt \F & \F & 0 & 33.51 & 20.14 & 0.09 & 0.17 & 0.29 & 0.87 & 0.87 & \phantom{1}\textbf{12.61} \\
    & 7pt \F & \Fkk & $0$ & 15.16 & \phantom{1}\underline{4.17} & \underline{0.39} & \underline{0.52} & \underline{0.64} & 0.60 & \textbf{0.11} & \phantom{1}47.99 \\
    & 7pt \F & \Fkk & $\{0.0, -0.6, -1.2\}$ & \textbf{13.35} & \phantom{1}\textbf{3.69} & \textbf{0.42} & \textbf{0.55} & \textbf{0.67} & \underline{0.49} & \textbf{0.11} & 148.43 \\
    \cmidrule{2-12}
    & 9pt \Fkk & \Fkk & \ding{55} & 15.81 & \phantom{1}4.43 & 0.38 & 0.50 & 0.63 & \underline{0.49} & \underline{0.12} & 170.47 \\
    & 10pt \Fkk & \Fkk & \ding{55} & \underline{14.74} & \phantom{1}4.18 & \underline{0.39} & \underline{0.52} & \underline{0.64} & \textbf{0.43} & \textbf{0.11} & \phantom{1}\underline{32.43} \\
    \bottomrule
        \toprule
    & & & & \multicolumn{8}{c}{Poselib RANSAC - \Phototourism{} - \textit{Scenario B} \emph{Small distortion}} \\
    \midrule
    & Minimal & Refinement & Sample & AVG $(^\circ)$ $\downarrow$ & MED $(^\circ)$ $\downarrow$ & AUC@5 $\uparrow$ & @10 & @20 & AVG $\epsilon(\lambda)$ $\downarrow$ & MED $\epsilon(\lambda)$ $\downarrow$ & Time (ms) $\downarrow$ \\
    \midrule
    \multirow{8}{*}{\rotatebox[origin=c]{90}{$\lambda_1 = \lambda_2$}}     & 7pt \F & \F & 0 & 12.91 & 4.47 & 0.37 & 0.51 & 0.65 & 0.15 & 0.15 & \phantom{1}\textbf{11.22} \\
    & 7pt \F & \Fk & $0$ & \textbf{10.74} & \textbf{3.15} & \textbf{0.45} & \textbf{0.58} & \textbf{0.71} & \textbf{0.12} & \textbf{0.04} & \phantom{1}43.62 \\
    & 7pt \F & \Fkk & $0$ & 13.45 & 3.67 & 0.42 & 0.55 & 0.68 & 0.50 &\underline{0.09} & \phantom{1}46.01 \\
    \cmidrule{2-12}
    & 8pt \Fk & \Fk & \ding{55} & \underline{11.00} & \underline{3.18} & \textbf{0.45} & \textbf{0.58} & \underline{0.70} & \underline{0.13} & \textbf{0.04} & \phantom{1}91.70 \\
    & 9pt \Fk & \Fk & \ding{55} & 11.66 & 3.31 & \underline{0.44} & \underline{0.57} & 0.69 & 0.14 & \textbf{0.04} & \phantom{1}46.96 \\
    \cmidrule{2-12}
    & 9pt \Fkk & \Fkk & \ding{55} & 14.48 & 4.03 & 0.40 & 0.52 & 0.65 & 0.47 & 0.10 & 169.90 \\
    & 10pt \Fkk & \Fkk & \ding{55} & 15.34 & 4.27 & 0.39 & 0.51 & 0.64 & 0.47 & 0.11 & \phantom{1}\underline{33.06} \\
    \midrule
    \midrule
    \multirow{4}{*}{\rotatebox[origin=c]{90}{$\lambda_1 \neq \lambda_2$}}     & 7pt \F & \F & 0 & \underline{13.95} & 5.14 & 0.33 & 0.47 & 0.62 & \textbf{0.15} & 0.15 & \phantom{1}\textbf{12.56} \\
    & 7pt \F & \Fkk & $0$ & \textbf{13.58} & \textbf{3.71} & \textbf{0.41} & \textbf{0.55} & \textbf{0.67} & 0.51 & \textbf{0.10} & \phantom{1}50.26 \\
    \cmidrule{2-12}
    & 9pt \Fkk & \Fkk & \ding{55} & 14.44 & \underline{4.02} & \underline{0.40} & \underline{0.53} & \underline{0.65} & \underline{0.47} & \textbf{0.10} & 184.45 \\
    & 10pt \Fkk & \Fkk & \ding{55} & 15.31 & 4.26 & 0.39 & 0.51 & 0.64 & \underline{0.47} & \underline{0.11} & \phantom{1}\underline{36.07} \\
    \bottomrule
        \toprule
    & & & & \multicolumn{8}{c}{Poselib RANSAC - \Phototourism{} - \textit{Scenario C} \emph{Visible distortion}} \\
    \midrule
    & Minimal & Refinement & Sample & AVG $(^\circ)$ $\downarrow$ & MED $(^\circ)$ $\downarrow$ & AUC@5 $\uparrow$ & @10 & @20 & AVG $\epsilon(\lambda)$ $\downarrow$ & MED $\epsilon(\lambda)$ $\downarrow$ & Time (ms) $\downarrow$ \\
    \midrule
    \multirow{10}{*}{\rotatebox[origin=c]{90}{$\lambda_1 = \lambda_2$}}     & 7pt \F & \F & 0 & 34.85 & 21.10 & 0.08 & 0.16 & 0.28 & 1.10 & 1.10 & \phantom{1}\textbf{12.23} \\
    & 7pt \F & \Fk & $-0.9$ & 10.75 & \phantom{1}\underline{3.18} & \textbf{0.45} & \textbf{0.58} & \underline{0.70} & \textbf{0.13} & \textbf{0.05} & \phantom{1}42.14 \\
    & 7pt \F & \Fkk & $-0.9$ & 13.43 & \phantom{1}3.78 & 0.41 & 0.54 & 0.67 & 0.50 & 0.11 & \phantom{1}44.73 \\
    & 7pt \F & \Fk & $\{-0.6, -0.9, -1.2\}$ & \textbf{10.54} & \phantom{1}\textbf{3.15} & \textbf{0.45} & \textbf{0.58} & \textbf{0.71} & \textbf{0.13} & \textbf{0.05} & \phantom{1}65.36 \\
    & 7pt \F & \Fkk & $\{-0.6, -0.9, -1.2\}$ & 13.10 & \phantom{1}3.64 & 0.42 & 0.55 & 0.68 & 0.49 & 0.11 & 141.56 \\
    \cmidrule{2-12}
    & 8pt \Fk & \Fk & \ding{55} & \underline{10.74} & \phantom{1}3.18 & \textbf{0.45} & \textbf{0.58} & \underline{0.70} & \textbf{0.13} & \textbf{0.05} & \phantom{1}84.60 \\
    & 9pt \Fk & \Fk & \ding{55} & 12.04 & \phantom{1}3.41 & \underline{0.43} & \underline{0.56} & 0.68 & \underline{0.15} & \underline{0.06} & \phantom{1}41.72 \\
    \cmidrule{2-12}
    & 9pt \Fkk & \Fkk & \ding{55} & 16.28 & \phantom{1}4.52 & 0.38 & 0.50 & 0.62 & 0.51 & 0.12 & 162.72 \\
    & 10pt \Fkk & \Fkk & \ding{55} & 14.79 & \phantom{1}4.21 & 0.39 & 0.51 & 0.64 & 0.45 & 0.12 & \phantom{1}\underline{31.27} \\
    \midrule
    \midrule
    \multirow{5}{*}{\rotatebox[origin=c]{90}{$\lambda_1 \neq \lambda_2$}}     & 7pt \F & \F & 0 & 36.05 & 22.65 & 0.07 & 0.14 & 0.26 & 1.10 & 1.10 & \phantom{1}\textbf{12.69} \\
    & 7pt \F & \Fkk & $-0.9$ & \underline{14.14} & \phantom{1}\underline{3.94} & \underline{0.40} & \underline{0.53} & \underline{0.66} & 0.56 & \textbf{0.11} & \phantom{1}47.27 \\
    & 7pt \F & \Fkk & $\{-0.6, -0.9, -1.2\}$ & \textbf{13.19} & \phantom{1}\textbf{3.68} & \textbf{0.42} & \textbf{0.55} & \textbf{0.68} & \underline{0.49} & \textbf{0.11} & 146.23 \\
    \cmidrule{2-12}
    & 9pt \Fkk & \Fkk & \ding{55} & 16.22 & \phantom{1}4.52 & 0.38 & 0.50 & 0.62 & 0.50 & \underline{0.12} & 168.76 \\
    & 10pt \Fkk & \Fkk & \ding{55} & 14.73 & \phantom{1}4.20 & 0.39 & 0.52 & 0.64 & \textbf{0.43} & \underline{0.12} & \phantom{1}\underline{32.41} \\
    \bottomrule
\end{tabular}
    }
    
    \label{tab:poselib_phototourism_scenaraB}
\end{table*}

\noindent \textbf{Prior knowledge about cameras.}
For the sampling-based strategy, we can adjust the number $k$ of samples and the sampled values $\M U_i = \{\hat{\lambda}_i^1, \hat{\lambda}_i^2, ..., \hat{\lambda}_i^k\}$ based on 
prior knowledge about the cameras.
We study three different scenarios:

\noindent \textit{Scenario A} -  \textit{Wild}: In the first scenario, 
we assume no knowledge about the 
cameras. The cameras can have distortions ranging from small to very high. This scenario represents, \eg, images downloaded from the Internet. To simulate this scenario,  we distorted the positions of the local features extracted from the undistorted images provided by the \ETH and \Phototourism datasets.\
For each pair of images, we sample undistortion parameters from a distribution $\mathcal{U}$. 
Given an undistortion parameter $\lambda$, we apply the corresponding distortion to an image 
using the inverse function of~\eqref{eq:02}. 
We apply either the 
same or different distortions based on the studied setup (\Fk \ or \Fkk). 
We define $\mathcal{U}$ as a piecewise distribution, which is uniform between $-1.5$ and $0$, while its density decreases linearly from $-1.5$ to $-1.8$, reaching half the density of the uniform range. 
This is done to simulate that in practice, distortion parameters in 
the range $[-1.5, 0]$ are more common than in the range $[-1.8, -1.5]$. 
Thus, 
it is natural to sample undistortion parameters from a wide rage of parameters for the sampling-based approach. 
Based on the 
results of the solver with $\M U_1 = \M U_2 = \{0,-0.6,-1.2\}$ in Fig.~\ref{fig:eth3d_box}, 
we evaluated this solver for this scenario.
Tab.~\ref{tab:poselib_phototourism_scenaraB} (top) shows the results for the \Phototourism datasets and Poselib RANSAC. 
Results for this scenario for \ETH and for GC-RANSAC are shown in the SM. 
Poselib RANSAC does not use non-minimal solvers in LO but refines poses using LM. 
Thus, the Refinement column 
indicates which parameters are optimized inside LO. 
We show results for cameras with the same distortion ($\lambda_1 = \lambda_2)$ 
and for cameras with different distortions ($\lambda_1 \not= \lambda_2)$.
In this scenario, we also tested variants in which we optimize different distortions even for cameras with the same distortion. 
As can be seen in Tab.~\ref{tab:poselib_phototourism_scenaraB}, the 7pt solver with the sampling-based strategy with $\M U_1 = \M U_2 = \{0,-0.6,-1.2\}$ even outperforms the dedicated minimal radial distortion solvers that are applied in the first step of RANSAC. The non-minimal 9pt \Fk \ and 10pt \Fkk \ solvers are 
faster than the sampling-based strategy. 
Yet, they are 
significantly less accurate. 

\begin{figure*}[t]
    \centering

    \begin{tikzpicture} 
        \begin{axis}[%
        hide axis, xmin=0,xmax=0,ymin=0,ymax=0,
        legend style={draw=white!15!white, 
        line width = 1pt,
        legend  columns =5, 
        /tikz/every even column/.append style={column sep=0.5cm},
        }
        ]
        
        \addlegendimage{Seaborn1}
        \addlegendentry{\tiny{7pt \Fk\{0\} }};
        \addlegendimage{Seaborn2}
        \addlegendentry{\tiny{7pt \Fk\{0, -0.6, -1.2\} }};
        \addlegendimage{Seaborn3}
        \addlegendentry{\tiny{8pt \Fk}};
        \addlegendimage{Seaborn4}
        \addlegendentry{\tiny{9pt \Fk}};
        \addlegendimage{Seaborn9}
        \addlegendentry{\tiny{7pt \F}};

        \addlegendimage{Seaborn7}
        \addlegendentry{\tiny{7pt \Fkk\{0\}}};
        \addlegendimage{Seaborn10}
        \addlegendentry{\tiny{7pt \Fkk\{0, -0.6, -1.2\}}};
        
        \addlegendimage{Seaborn5}
        \addlegendentry{\tiny{9pt \Fkk}};
        \addlegendimage{Seaborn6}
        \addlegendentry{\tiny{10pt \Fkk}};
        
        \addlegendimage{white}
        \addlegendentry{~};
        \end{axis}
    \end{tikzpicture}
             
    \begin{subfigure}{0.49\textwidth}
    \includegraphics[width=0.49\linewidth]
    {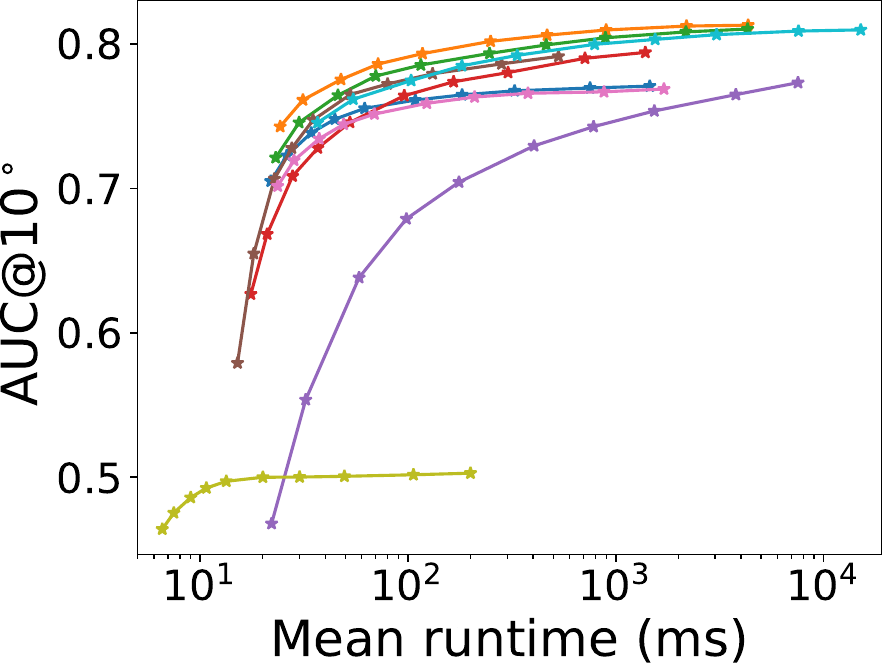}  \hfill \includegraphics[width=0.49\linewidth]{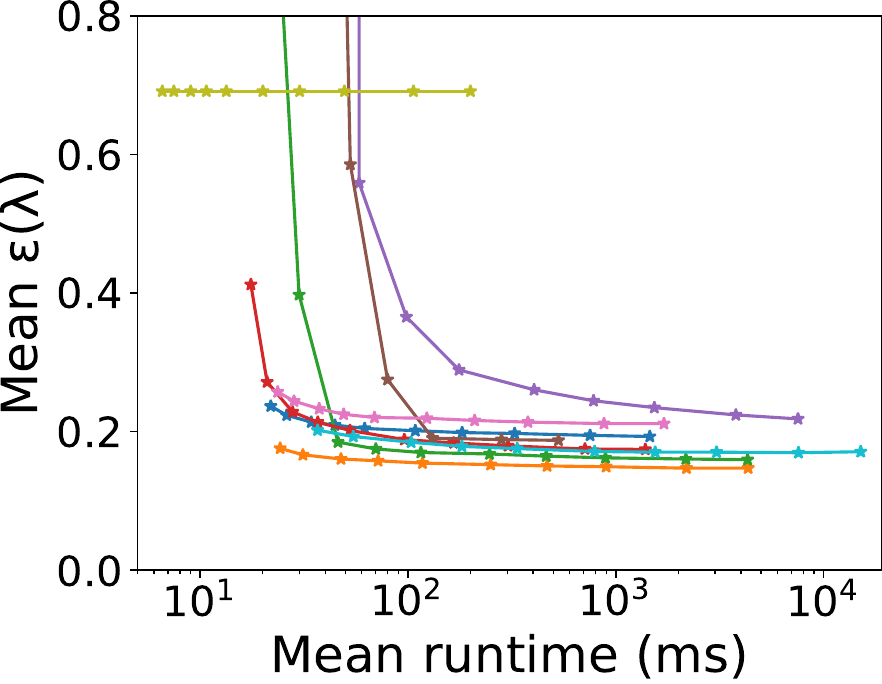}   
    \caption{}
    \label{fig:graph_vitus_neq}
    \end{subfigure}
    \hfill
    \begin{subfigure}{0.49\textwidth}       
    \includegraphics[width=0.49\linewidth]{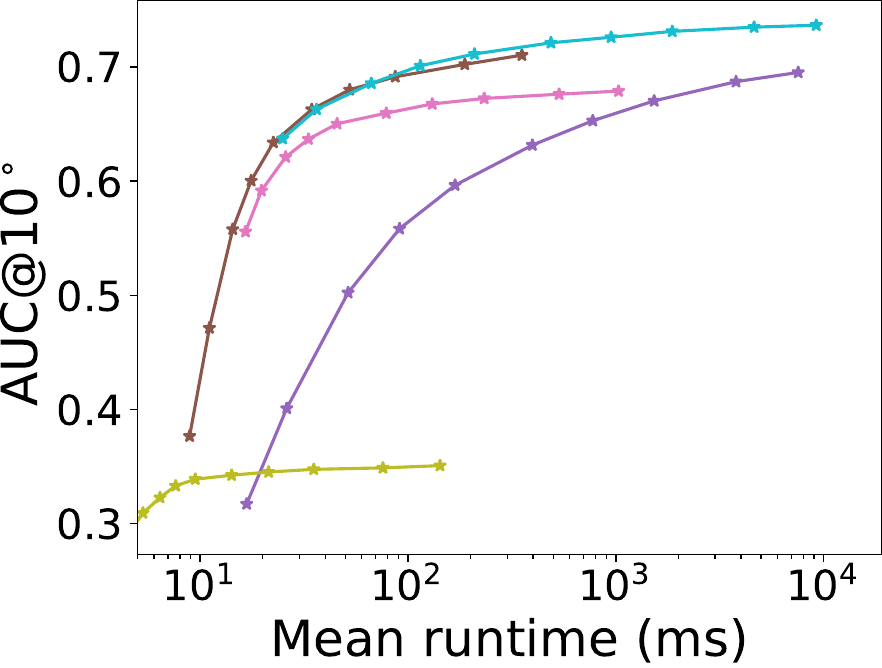} \hfill     \includegraphics[width=0.49\linewidth]{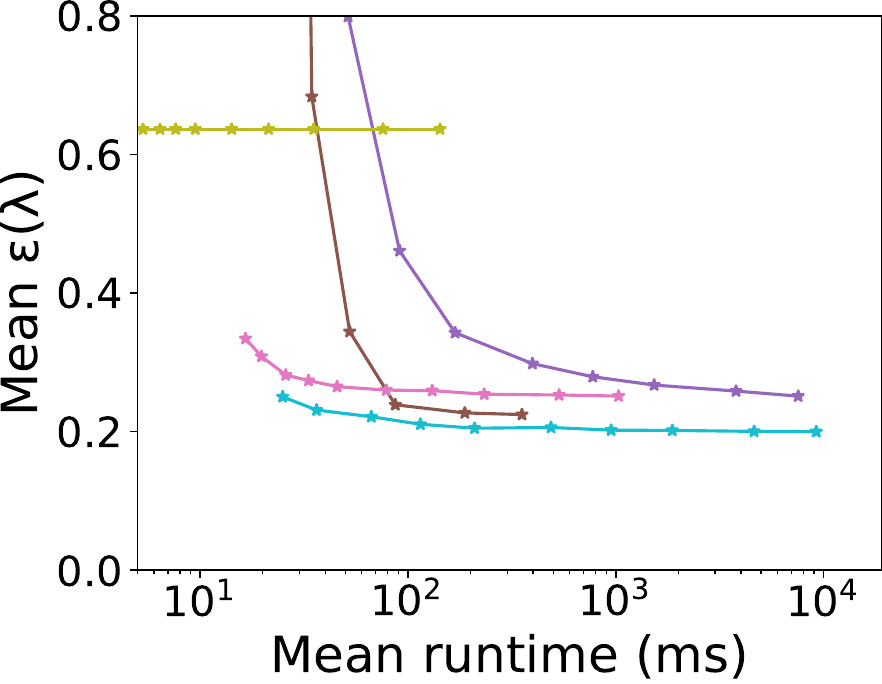}           
    \caption{}
    \label{fig:graph_vitus_eq}
    \end{subfigure}
    
    \begin{subfigure}{0.49\textwidth}
    \includegraphics[width=0.49\linewidth]{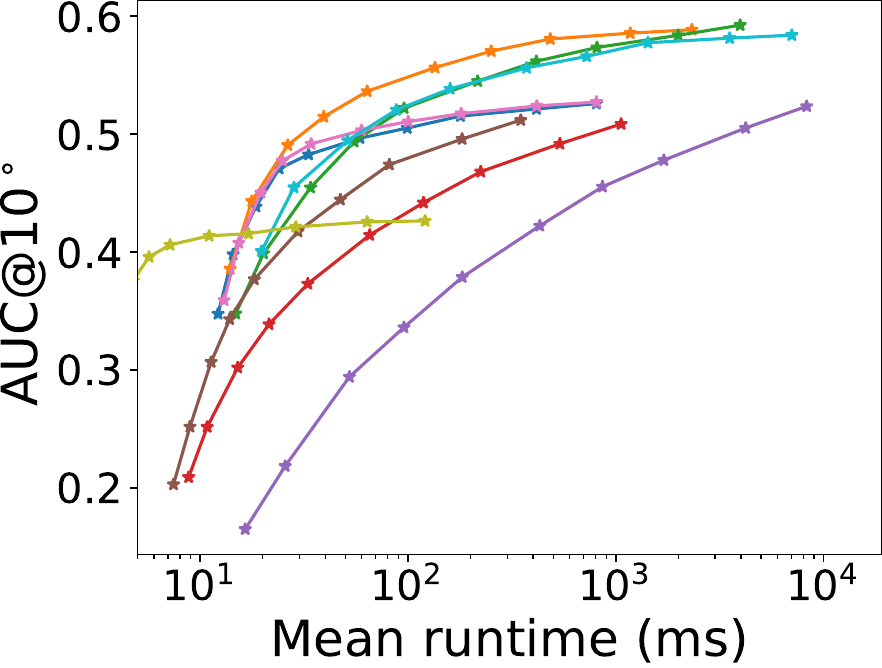}  \hfill \includegraphics[width=0.49\linewidth]{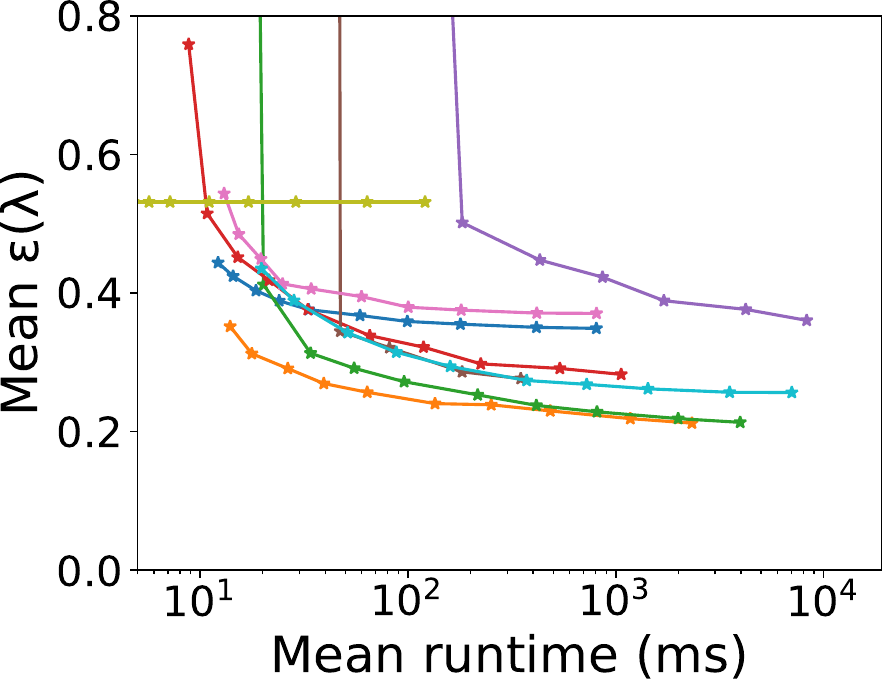}  
    \caption{}
    \label{fig:graph_rotunda_neq}
    \end{subfigure}
    \hfill
    \begin{subfigure}{0.49\textwidth}       
    \includegraphics[width=0.49\linewidth]        {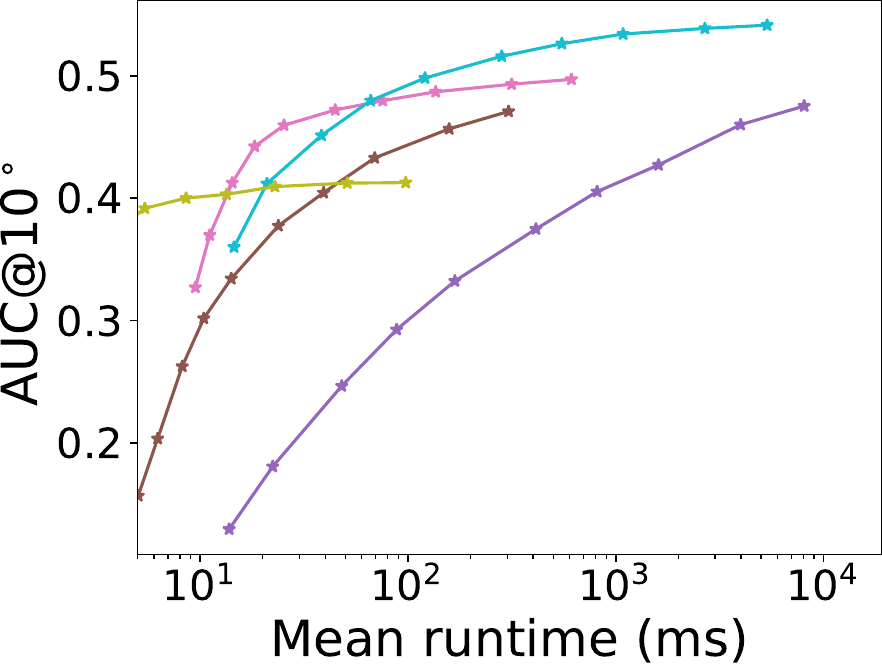} \hfill     \includegraphics[width=0.49\linewidth]{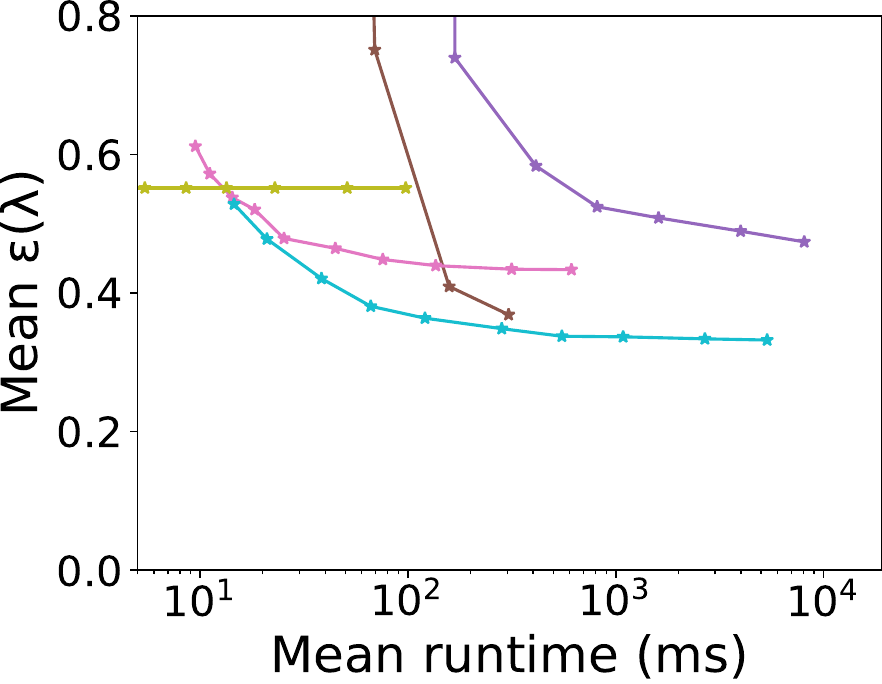}   
    \caption{}
    \label{fig:graph_rotunda_eq}
    \end{subfigure}
    
    \caption{Pose AUC@10$^\circ$ and mean absolute $\lambda$ errors plotted for different number of RANSAC iterations on the \VITUS (a, b) and \ROTUNDA (c,d) scenes when considering pairs of images taken by the same camera (a,c) and two different cameras (b,d). 
    } 
    \label{fig:poselib_vitus_time}
\end{figure*}

\noindent \textit{Scenario B}  - \textit{Small distortion}: In the second scenario, we simulate prior knowledge that our cameras have small distortion, \eg, we are processing images taken by mobile phone or DSLR cameras. 
To simulate this scenario, we distort the feature points with distortions corresponding to  undistortion parameters uniformly sampled from the interval $[-0.3, 0]$. 
In this case, it makes sense to run the 7pt solver in the sampling-based strategy only once with a small undistortion parameter. We decided to use $\M U_1 = \M U_2 = \{0\}$ to simulate the standard baseline.
Tab.~\ref{tab:poselib_phototourism_scenaraB} (middle) shows the results for this scenario. 
In this case, it can be seen that the baseline 7pt solver followed by LO of the radial distortion parameters outperforms the minimal radial distortion solvers at competitive run-times.

\begin{table*}[ht!]
    \centering
    \caption{Results on 10,000 image pairs from the \VITUS scene, and on 3,424 $\lambda_1 \neq \lambda_2$ image pairs and 1,795 $\lambda_1 = \lambda_2$ image pairs from the \ROTUNDA scene, using both GC-RANSAC and Poselib RANSAC. The reported statistics are the same as in Tab.~\ref{tab:poselib_phototourism_scenaraB}.}
    \setlength{\tabcolsep}{4.8pt}
    \resizebox{1.0\linewidth}{!}{
    \begin{tabular}{ c | r r c | c c c c c | c c | c}
    \toprule
        & & & & \multicolumn{8}{c}{GC-RANSAC - \ROTUNDA} \\
        \midrule
        & Minimal & Non-Minimal & Sample & AVG $(^\circ)$ $\downarrow$ & MED $(^\circ)$ $\downarrow$ & AUC@5 $\uparrow$ & @10 & @20 & AVG $\epsilon(\lambda)$ $\downarrow$ & MED $\epsilon(\lambda)$ $\downarrow$ & Time (ms) $\downarrow$ \\
        \midrule
        \multirow{12}{*}{\rotatebox[origin=c]{90}{$\lambda_1 = \lambda_2$}} & 7pt \F & 7pt \F & 0 & \underline{17.17} & \phantom{1}6.62 & \underline{0.21} & 0.36 & \underline{0.52} & 0.53 & \underline{0.09} & \phantom{1}420.61 \\
        & 7pt \F & 8pt \F & 0 & 20.48 & 10.55 & 0.12 & 0.24 & 0.42 & 0.53 & \underline{0.09} & \phantom{1}\underline{370.18}\\
        & 7pt \F & 9pt \Fk & 0 & 19.01 & \phantom{1}7.58 & 0.18 & 0.32 & 0.48 & 0.43 & 0.10 & \bf \phantom{1}280.01 \\
        & 7pt \F & 12pt \Fkk & 0 & 18.84 & \phantom{1}8.72 & 0.13 & 0.28 & 0.46 & 0.48 & \bf 0.05 & \phantom{1}380.92 \\
        & 7pt \F & 9pt \Fk & $\{-1.2, -0.6, 0\}$ & 17.77 & \phantom{1}6.06 & \underline{0.21} & \underline{0.37} & \underline{0.52} & 0.32 & 0.15 & \phantom{1}570.47 \\
        & 7pt \F & 12pt \Fkk & $\{-1.2, -0.6, 0\}$ & \bf 16.63 & \phantom{1}\bf 5.44 & \underline{0.21} & \textbf{0.38} & \bf 0.54 & \bf 0.25 & \underline{0.09} & \phantom{1}600.09 \\
        
        \cmidrule{2-12}
        & 8pt \Fk & 9pt \Fk & \ding{55} & 17.62 & \phantom{1}6.36 & \bf 0.23 & \underline{0.37} & \underline{0.52} & 0.33 & 0.17 & 2250.00 \\
        & 8pt \Fk & 12pt \Fkk & \ding{55} & 17.37 & \phantom{1}\underline{5.67} & \textbf{0.23} & \bf 0.38 & \textbf{0.54} & \underline{0.28} & 0.14 & 2260.93 \\
        \cmidrule{2-12}
        & 9pt \Fk & 9pt \Fk & \ding{55} & 20.38 & \phantom{1}9.98 & 0.16 & 0.28 & 0.43 & 0.40 & 0.22 & \phantom{1}720.22\\
        & 9pt \Fk & 12pt \Fkk & \ding{55} & 20.27 & 10.08 & 0.15 & 0.27 & 0.43 & 0.34 & 0.17 & \phantom{1}750.08\\
        \midrule
        \midrule
        \multirow{6}{*}{\rotatebox[origin=c]{90}{$\lambda_1 \neq \lambda_2$}} & 7pt \F & 7pt \F & 0 & \underline{28.70} & \phantom{1}\underline{8.33} & \bf 0.24 & \underline{0.35} & \underline{0.48} & 0.55 & \bf 0.09 & \phantom{1}350.41 \\
        & 7pt \F & 8pt \F & 0 & 30.39 & 11.17 & 0.16 & 0.27 & 0.41 & 0.55 & \bf 0.09 & \phantom{1}\underline{310.68}\\
        & 7pt \F & 12pt \Fkk & $\{-1.2, -0.6, 0\}$ & \bf 28.07 & \phantom{1}\bf 6.36 & \underline{0.21} & \bf 0.36 & \bf 0.49 & \bf 0.36 & \underline{0.26} & 1480.41\\
        \cmidrule{2-12}
        & 9pt \Fkk & 12pt \Fkk & \ding{55} & 29.63 & \phantom{1}9.32 & 0.17 & 0.30 & 0.44 & \underline{0.43} & \underline{0.26} & 7340.00 \\
        & 10pt \Fkk & 12pt \Fkk & \ding{55} & 29.94 & 10.63 & 0.17 & 0.29 & 0.42 & 0.45 & 0.28 & \phantom{1}\bf 310.55\\
        \bottomrule
     \toprule
    & & & & \multicolumn{8}{c}{Poselib RANSAC - \ROTUNDA} \\
    \midrule
    & Minimal & Refinement & Sample & AVG $(^\circ)$ $\downarrow$ & MED $(^\circ)$ $\downarrow$ & AUC@5 $\uparrow$ & @10 & @20 & AVG $\epsilon(\lambda)$ $\downarrow$ & MED $\epsilon(\lambda)$ $\downarrow$ & Time (ms) $\downarrow$ \\
    \midrule
    \multirow{12}{*}{\rotatebox[origin=c]{90}{$\lambda_1 = \lambda_2$}}     & 7pt \F & \F & 0 & 21.34 & 6.58 & 0.29 & 0.42 & 0.56 & 0.53 & \underline{0.09} & \phantom{11}\textbf{25.96} \\
    & 7pt \F & \F + \Fk & 0 & 19.96 & 5.39 & 0.34 & 0.47 & 0.60 & 0.44 & \underline{0.09} & \phantom{11}\underline{38.34} \\
    & 7pt \F & \F + \Fkk & 0 & 19.97 & 5.21 & 0.35 & 0.47 & 0.60 & 0.46 & 0.11 & \phantom{11}38.51 \\
    & 7pt \F & \Fk & $0$ & 18.45 & 4.36 & 0.38 & 0.51 & 0.63 & 0.35 & 0.10 & \phantom{1}124.65 \\
    & 7pt \F & \Fkk & $0$ & 18.66 & 4.22 & 0.39 & 0.52 & 0.64 & 0.37 & 0.11 & \phantom{1}124.28 \\
    & 7pt \F & \Fk & $\{0.0, -0.6, -1.2\}$ & \underline{17.24} & \underline{3.13} & \underline{0.45} & \underline{0.57} & \underline{0.68} & \textbf{0.22} & \textbf{0.08} & \phantom{1}259.52 \\ 
    & 7pt \F & \Fkk & $\{0.0, -0.6, -1.2\}$ & 17.32 & \underline{3.13} & \underline{0.45} & \underline{0.57} & \underline{0.68} & \underline{0.26} & \underline{0.09} & \phantom{1}723.32 \\
    \cmidrule{2-12}
    & 8pt \Fk & \Fk & \ding{55} & \textbf{17.05} & \textbf{3.01} & \textbf{0.46} & \textbf{0.58} & \textbf{0.68} & \textbf{0.22} & \textbf{0.08} & 1165.29 \\
    & 9pt \Fk & \Fk & \ding{55} & 19.89 & 5.44 & 0.35 & 0.46 & 0.58 & 0.29 & 0.10 & \phantom{1}328.17 \\ 
    \cmidrule{2-12}
    & 9pt \Fkk & \Fkk & \ding{55} & 20.06 & 4.87 & 0.36 & 0.48 & 0.60 & 0.37 & 0.12 & 2837.52 \\
    
    & 10pt \Fkk & \Fkk & \ding{55} & 21.63 & 4.84 & 0.37 & 0.48 & 0.59 & 0.29 & 0.12 & \phantom{1}120.21 \\
    \midrule
    \midrule
    \multirow{6}{*}{\rotatebox[origin=c]{90}{$\lambda_1 \neq \lambda_2$}}     & 7pt \F & \F & 0 & 35.31 & 8.00 & 0.31 & 0.41 & 0.51 & 0.55 & \textbf{0.09} & \phantom{11}\textbf{25.09} \\
    & 7pt \F & \F + \Fkk & 0 & 34.06 & 6.34 & 0.36 & 0.45 & 0.54 & 0.51 & 0.20 & \phantom{11}\underline{37.52} \\
    & 7pt \F & \Fkk & $0$ & \underline{32.90} & \underline{4.76} & \underline{0.39} & \underline{0.49} & \underline{0.57} & 0.44 & 0.18 & \phantom{1}107.31 \\
    & 7pt \F & \Fkk & $\{0.0, -0.6, -1.2\}$ & \textbf{31.92} & \textbf{3.41} & \textbf{0.44} & \textbf{0.53} & \textbf{0.60} & \textbf{0.34} & \underline{0.14} & \phantom{1}668.34 \\
    \cmidrule{2-12}
    & 9pt \Fkk & \Fkk & \ding{55} & 35.12 & 6.82 & 0.35 & 0.44 & 0.53 & 0.49 & 0.19 & 3607.10 \\
    & 10pt \Fkk & \Fkk & \ding{55} & 35.26 & 6.74 & 0.36 & 0.44 & 0.53 & \underline{0.38} & 0.19 & \phantom{1}135.31 \\
    \bottomrule
        \toprule
        & & & & \multicolumn{8}{c}{GC-RANSAC - \VITUS} \\
        \midrule
        & Minimal & Non-Minimal & Sample & AVG $(^\circ)$ $\downarrow$ & MED $(^\circ)$ $\downarrow$ & AUC@5 $\uparrow$ & @10 & @20 & AVG $\epsilon(\lambda)$ $\downarrow$ & MED $\epsilon(\lambda)$ $\downarrow$ & Time (ms) $\downarrow$ \\
        \midrule
        \multirow{11}{*}{\rotatebox[origin=c]{90}{$\lambda_1 = \lambda_2$}} & 7pt \F & 7pt \F & 0 & 11.16 & 5.48 & 0.25 & 0.40 & 0.58 & 0.69 & 0.72 & \phantom{1}620.91 \\
        & 7pt \F & 8pt \F & 0 & 10.84 & 5.41 & 0.25 & 0.41 & 0.59 & 0.69 & 0.72 & \phantom{1}510.49 \\
        & 7pt \F & 9pt \Fk & 0 & \phantom{1}7.48 & 2.72 & 0.35 & 0.55 & 0.71 & 0.30 & 0.16 & \bf 270.36 \\
        & 7pt \F & 12pt \Fkk & 0 & 10.36 & 4.71 & 0.23 & 0.42 & 0.61 & 0.51 & 0.25 & \phantom{1}500.98 \\
        & 7pt \F & 9pt \Fk & $\{-1.2, -0.6, 0\}$ & \phantom{1}\bf 6.25 & \bf 2.18 & \bf 0.39 & \bf 0.60 & \bf 0.75 & \underline{0.23} & 0.14 & \phantom{1}480.04 \\
        & 7pt \F & 12pt \Fkk & $\{-1.2, -0.6, 0\}$ & \phantom{1}6.65 & 2.52 & 0.36 & 0.57 & \underline{0.73} & \bf 0.20 & \bf 0.12 & \phantom{1}540.93 \\
        \cmidrule{2-12}
        & 8pt \Fk & 9pt \Fk & \ding{55} & \phantom{1}\underline{6.31} & \underline{2.23} & \textbf{0.39} & \underline{0.59} & \textbf{0.75} & 0.24 & \underline{0.13} & 1050.71 \\
        & 8pt \Fk & 12pt \Fkk & \ding{55} & \phantom{1}6.81 & 2.65 & 0.35 & 0.56 & 0.72 & 0.24 & \bf 0.12 & 1120.83 \\
        \cmidrule{2-12}
        & 9pt \Fk & 9pt \Fk & \ding{55} & \phantom{1}6.83 & 2.48 & \underline{0.37} & 0.57 & \underline{0.73} & 0.27 & 0.14 & \phantom{1}\underline{420.32} \\
        & 9pt \Fk & 12pt \Fkk & \ding{55} & \phantom{1}7.29 & 2.90 & 0.32 & 0.54 & 0.71 & 0.28 & 0.15 & \phantom{1}460.46\\
        \midrule
        \midrule
        \multirow{6}{*}{\rotatebox[origin=c]{90}{$\lambda_1 \neq \lambda_2$}} & 7pt \F & 7pt \F & 0 & 16.71 & 9.51 & 0.14 & 0.27 & 0.45 & 0.64 & 0.68 & 1240.02 \\
        & 7pt \F & 8pt \F & 0 & 16.76 & 9.70 & 0.14 & 0.26 & 0.44 & 0.64 & 0.68 & \underline{1140.21} \\
        & 7pt \F & 12pt \Fkk & $\{-1.2, -0.6, 0\}$ & \phantom{1}\bf 8.23 & \bf 3.37 & \bf 0.29 & \bf 0.50 & \bf 0.68 & \bf 0.21 & \bf 0.13 & 3600.91 \\
        \cmidrule{2-12}
        & 9pt \Fkk & 12pt \Fkk & \ding{55} & \phantom{1}8.60 & \underline{3.63} & 0.27 & \underline{0.48} & 0.66 & \underline{0.31} & \underline{0.21} & 7050.20\\
        & 10pt \Fkk & 12pt \Fkk & \ding{55} & \phantom{1}\underline{8.55} & 3.65 & \underline{0.28} & \underline{0.48} & \underline{0.67} & 0.32 & 0.22 & \phantom{1}\bf 340.40 \\
        \bottomrule
    \toprule
    & & & & \multicolumn{8}{c}{Poselib RANSAC - \VITUS} \\
    \midrule
    & Minimal & Refinement & Sample & AVG $(^\circ)$ $\downarrow$ & MED $(^\circ)$ $\downarrow$ & AUC@5 $\uparrow$ & @10 & @20 & AVG $\epsilon(\lambda)$ $\downarrow$ & MED $\epsilon(\lambda)$ $\downarrow$ & Time (ms) $\downarrow$ \\
    \midrule
    \multirow{12}{*}{\rotatebox[origin=c]{90}{$\lambda_1 = \lambda_2$}}     & 7pt \F & \F & 0 & 13.12 & 4.83 & 0.38 & 0.50 & 0.65 & 0.69 & 0.72 & \phantom{11}\textbf{28.75} \\
    & 7pt \F & \F + \Fk & 0 & \phantom{1}9.58 & 1.79 & 0.57 & 0.68 & 0.78 & 0.32 & 0.13 & \phantom{11}\underline{40.54} \\
    & 7pt \F & \F + \Fkk & 0 & \phantom{1}9.50 & 1.85 & 0.57 & 0.68 & 0.78 & 0.34 & 0.14 & \phantom{11}40.99 \\
    & 7pt \F & \Fk & $0$ & \phantom{1}7.40 & 1.11 & 0.68 & 0.77 & 0.84 & 0.19 & \underline{0.05} & \phantom{1}112.58 \\
    & 7pt \F & \Fkk & $0$ & \phantom{1}7.38 & 1.19 & 0.67 & 0.76 & 0.84 & 0.21 & 0.06 & \phantom{1}110.99 \\
    & 7pt \F & \Fk & $\{0.0, -0.6, -1.2\}$ & \phantom{1}\textbf{6.13} & \textbf{0.98} & \textbf{0.73} & \textbf{0.81} & \textbf{0.87} & \textbf{0.15} & \textbf{0.04} & \phantom{1}221.91 \\ 
    & 7pt \F & \Fkk & $\{0.0, -0.6, -1.2\}$ & \phantom{1}\underline{6.19} & 1.04 & \underline{0.72} & \textbf{0.81} & \textbf{0.87} & 0.17 & \underline{0.05} & \phantom{1}576.07 \\
    \cmidrule{2-12}
    & 8pt \Fk & \Fk & \ding{55} & \phantom{1}6.37 & \underline{1.00} & \underline{0.72} & \textbf{0.81} & \textbf{0.87} & \underline{0.16} & \textbf{0.04} & \phantom{1}495.53 \\
    & 9pt \Fk & \Fk & \ding{55} & \phantom{1}6.91 & 1.07 & 0.70 & \underline{0.78} & \underline{0.85} & 0.18 & \underline{0.05} & \phantom{1}175.25 \\
    \cmidrule{2-12}
    & 9pt \Fkk & \Fkk & \ding{55} & \phantom{1}7.72 & 1.21 & 0.67 & 0.76 & 0.83 & 0.23 & 0.07 & 1161.15 \\
    & 10pt \Fkk & \Fkk & \ding{55} & \phantom{1}6.78 & 1.13 & 0.69 & \underline{0.78} & \underline{0.85} & 0.19 & 0.06 & \phantom{11}75.65 \\
    \midrule
    \midrule
    \multirow{6}{*}{\rotatebox[origin=c]{90}{$\lambda_1 \neq \lambda_2$}}     & 7pt \F & \F & 0 & 18.13 & 8.72 & 0.23 & 0.35 & 0.52 & 0.64 & 0.68 & \phantom{11}\textbf{45.52} \\
    & 7pt \F & \F + \Fkk & 0 & 14.16 & 4.83 & 0.37 & 0.50 & 0.64 & 0.45 & 0.30 & \phantom{11}\underline{62.19} \\
    & 7pt \F & \Fkk & $0$ & \phantom{1}9.47 & 1.84 & 0.57 & 0.67 & 0.77 & 0.25 & 0.11 & \phantom{1}177.30 \\
    & 7pt \F & \Fkk & $\{0.0, -0.6, -1.2\}$ & \phantom{1}\textbf{7.49} & \textbf{1.49} & \textbf{0.63} & \textbf{0.73} & \textbf{0.82} & \textbf{0.20} & \textbf{0.09} & 1014.09 \\ 
    \cmidrule{2-12}
    & 9pt \Fkk & \Fkk & \ding{55} & \phantom{1}8.92 & 1.73 & 0.58 & 0.69 & 0.78 & 0.26 & 0.11 & 2182.09 \\
    & 10pt \Fkk & \Fkk & \ding{55} & \phantom{1}\underline{8.10} & \underline{1.62} & \underline{0.60} & \underline{0.70} & \underline{0.80} & \underline{0.23} & \underline{0.10} & \phantom{1}108.71 \\
    \bottomrule
    \end{tabular}
    }
    \label{tab:vitus_gcr}
\end{table*}

\noindent \textit{Scenario C} - \textit{Visible distortion}:
In the last scenario, we assume that we know that our images have visible (but unknown) distortion. 
To simulate this, we distort the feature points with distortions corresponding to  undistortion parameters uniformly sampled from the interval $[-1.8, -0.5]$.
For the sampling-based strategy, we tested two different variants: (1)  $\M U_1 = \M U_2 = \{-0.9\}$ and  (2) $\M U_1 = \M U_2 = \{-0.6,-0.9,-1.2\}$. The results are shown in Tab.~\ref{tab:poselib_phototourism_scenaraB} (bottom). 
Again, the 7pt solver with 
$\M U_1 = \M U_2 = \{-0.6,-0.9,-1.2\}$ outperforms the minimal radial distortion solvers. 
Even sampling one distortion -0.9 leads to similar results to the 8pt \Fk \ solver, 
 while being much faster. 
For the case $\lambda_1 \not= \lambda_2$, 
both sampling strategies significantly outperform the radial distortion solvers.

This experiment shows two important observations: 
(1) The sampling-based strategy performs similar to or even better than the dedicated minimal radial distortion solvers, \ie, dedicated radial distortion solvers do not seem necessary in practice. 
(2) Having additional knowledge about the cameras (even vague knowledge, \eg, that the cameras have visible distortion) can improve the performance of the sampling-based strategy. 
The same observations can be made for 
GC-RANSAC, 
where the sampling-based strategy moreover is significantly faster than running minimal radial distortion solvers inside RANSAC (see SM). 

\noindent \textbf{Real-World Scenario.} 
In the previous experiments, we synthesized distortions to be able to precisely measure the behavior of the different approaches under varying levels of distortion. 
To evaluate the performance of the tested methods under real-world conditions and for cameras with different distortions, 
we use our own dataset consisting of the \ROTUNDA and \VITUS scenes.
Tab.~\ref{tab:vitus_gcr} shows results for both GC-RANSAC and Poselib RANSAC. 
In addition, using Poselib RANSAC, we conducted an experiment on both scenes to assess each solver's performance, in terms of the AUC@10$^\circ$ of pose errors, and the median absolute error of the estimated distortion parameter(s), for different numbers of RANSAC iterations. The plots of the measured metrics vs. the average run-time are reported in Fig.~\ref{fig:poselib_vitus_time}.
The upper parts of the tables and  Fig.~\ref{fig:poselib_vitus_time} (a,c) show results for pairs of images taken by the same camera. 
The lower parts of the tables and Fig.~\ref{fig:poselib_vitus_time} (b,d)  show the results for two different cameras. We observe a similar behavior as in our previous experiments for both RANSAC variants. In both cases and for both scenes, the sampling-based strategy with $\M U_1 = \M U_2 = \{0,-0.6,-1.2\}$ is, in general, the best performing method, slightly outperforming minimal radial distortion solvers. 
Since the \ROTUNDA scene contains many images taken with a standard mobile phone camera, for this scene the 7pt solver followed by local optimization of the distortion parameters performs well. 
Approach (1) from Sec.~\ref{sec:introduction}, modelling distortion only during the final refinement step (7pt \F + \F + \Fk), does not always perform significantly better than not even modelling distortion.

\clearpage

\section{Conclusion}
Modelling radial distortion during relative pose estimation 
is important. 
Yet, (minimal) radial distortion solvers 
are significantly more complex than 
solvers for pinhole cameras, 
in terms of both 
run-time and 
implementation efforts. 
This paper thus asks the question whether minimal radial distortion solvers are actually necessary in practice. 
Extensive experiments 
show that a simple strategy combining a minimal non-distortion solver with sampling radial distortion parameters outperforms existing 
distortion solvers. 
This 
approach is simpler to implement, faster than the best-performing minimal distortion solvers at a similar accuracy, and significantly more accurate than faster non-minimal 
distortion solvers. 
We conclude that 
minimal distortion solvers are not truly necessary. 

\section*{Acknowledgements}

C. T., Y.D. and Z.K. were supported by the Czech Science Foundation (GAČR) JUNIOR STAR Grant (No. 22-23183M). 
V. K. was supported by the project no. 1/0373/23. and the TERAIS project, a Horizon-Widera-2021 program of the European Union under the Grant agreement number 101079338. 
T. S. was supported by the EU Horizon 2020 project RICAIP (grant agreement No. 857306).
Part of the research results was obtained using the computational resources procured in the national project National competence centre for high performance computing (project code: 311070AKF2) funded by European Regional Development Fund, EU Structural Funds Informatization of society, Operational Program Integrated Infrastructure. 

%
%

%% file: supplementary.tex
\title{\texorpdfstring{Are Minimal Radial Distortion Solvers Necessary for Relative Pose Estimation? \\ 
- \\
Supplementary Material}{Are Minimal Radial Distortion Solvers Necessary for Relative Pose Estimation? - Supplementary Material}}

\titlerunning{Are Minimal Radial Distortion Solvers Necessary? - supp. mat.}


\authorrunning{C.~Tzamos et al.}


\author{}
\institute{}

\maketitle

This supplementary material provides additional details, tables, and figures to support the results presented in the main paper. 
More precisely, Sec.~\ref{sec:refined_solvers} presents details on the modified solvers discussed in Sec. 3 of the main paper. 
Sec.~\ref{sec:inverse_one_div_model} provides the closed-form inverse of the one-parameter division undistortion model mentioned in Sec.~4 of the main paper. 
Sec.~\ref{sec:dataset_details} presents additional details on our new benchmark (see Sec.~4 of the main paper). 
Finally, Sec.~\ref{sec:experiments_details} provides additional details on the experimental setup and the experimental results mentioned in Sec.~4 of the main paper. 

\section{Modified Solvers for Non-minimal Fitting}
\label{sec:refined_solvers}

\subsection{ \Fk \ solver for equal and unknown distortion}

Based on~\cite{fitzgibbon2001simultaneous}, the epipolar constraint with equal and unknown radial distortion can be written as
\begin{equation}
\begin{matrix}
& & [ & x'_d x_d & x'_d y_d & x'_d & y'_d x_d & y'_d y_d & y'_d & x_d & y_d & 1 & ]& \cdot &  \M f & + &  \\
+& \lambda & [ & 0 & 0 & x_d r^2 & 0 & 0 & y'_d r^2 & x_d r'^2 & y_d r' & r^2+r'^2 & ]&\cdot &  \M f & + &  \\
+ &\lambda^2 & [ & 0 & 0 &0 & 0 & 0 & 0 & 0 & 0 & r^2 r'^2 &]& \cdot &  \M f & = & 0 \enspace ,
\end{matrix}\label{eq:03}
\end{equation}
where $\M f$ is a $9\times 1$ vector that contains the entries of the fundamental matrix \F\ in row-wise order and $r, r^\prime$ denote the Euclidean distances of the distorted points $\mathbf{x_i}, \mathbf{x_i^\prime}$, respectively, to the center of distortion. It is common to assume that the center of distortion is in the center of the image, \ie, $r = \sqrt{x_d^2 + y_d^2}$. 
By stacking the rows of $n$ correspondences,~\eqref{eq:03} can be written as
\begin{equation}
(\M A_0 + \lambda \M A_1 + \lambda^2 \M A_2)\M f = \M 0 \enspace .\label{eq:04}
\end{equation}
For 9 point correspondences in the 9pt \Fk\ solver, 
equation \eqref{eq:04} defines a polynomial eigenvalue problem that can be solved by computing the eigenvalues of a $18 \times 18$ matrix. In~\cite{fitzgibbon2001simultaneous}, it was shown how the number of solutions of~\eqref{eq:04} can be reduced from 18 to 10 by transforming the problem to an eigenvalue problem of size $10 \times 10$.
However, in~\cite{fitzgibbon2001simultaneous} it was also noted that 4 of these 10 solutions have been found to be imaginary. In our case, we show that the 4 imaginary solutions can be directly removed and we only need to find the eigenvalues of a $6\times 6$ matrix. Since matrix $\M A_2$ is singular while $\M A_0$ is full-rank, we first let $\sigma = 1/\lambda$. Then~\eqref{eq:04} can be written as 
\begin{equation}
(\M A_2 + \sigma \M A_1 + \sigma^2 \M A_0)\M f = \M 0 \enspace .\label{eq:05}
\end{equation}
Solving for $\sigma$ is equivalent to finding the eigenvalues of the following $18 \times 18$ matrix
\begin{equation}
\M B = \begin{bmatrix}
\M 0 & {\M I}\\
-{\M A}_0^{-1}{\M A}_2 & -{\M A}_0^{-1}{\M A}_1 
\end{bmatrix} \enspace .\label{eq:06}
\end{equation}
There are 8 zero columns in $\M A_2$, and 4 zero columns in $\M A_1$. To solve this problem efficiently, we use the technique from~\cite{kukelova2012polynomial}: the zero columns in $-{\M A}_0^{-1}{\M A}_2$ and $-{\M A}_0^{-1}{\M A}_1$ can be removed together with the corresponding rows. In this case, the size of the matrix $\M B$ is reduced to $6\times 6$, and we only need to find the eigenvalues of a $6\times 6$ matrix. Note that in the solver, we directly construct the reduced $6 \times 6$ matrix and avoid computations on the matrix~\eqref{eq:06}

For the non-minimal case, \ie, the case where the number of point correspondences is larger than 9, $-{\M A}_0^{-1}{\M A}_2$ and $-{\M A}_0^{-1}{\M A}_1$ are solved using linear least squares (which can be efficiently solved using the \texttt{ColPivHouseholderQR} function in the Eigen library~\cite{eigenweb}).

\subsection{\Fkk \ solver for different unknown distortions}
Similarly to the equal distortion case, here we propose minor modifications to the 12pt solver for two cameras with different distortions. The original 12pt solver was proposed in~\cite{kukelova2010fast}.
Note that in the following equations we use $\lambda$ and $\lambda^{\prime}$, similarly to Eq. (1) in the main paper, 
and not $\lambda_1$ and $\lambda_2$ as used in the names of the solvers.

As suggested in\cite{kukelova2010fast}, we consider $\lambda'$ as a hidden variable, \ie, we ''hide'' it into coefficient matrices. Then the epipolar constraint can be written as
\begin{equation}
\begin{matrix}
& & [ & x'_d x_d & x'_d y_d & x'_d & y'_d x_d & y'_d y_d & y'_d & x_d & y_d & 1 & x'_d r^2 & y'_d r^2 & r^2 & ]& \cdot &  \tilde{\M f} & + &  \\
+ &\lambda' & [ & 0 & 0 &0 & 0 & 0 & 0 & x_d r'^2 & y_d r'^2 & r'^2 & 0 & 0 & r^2 r'^2 &]& \cdot &  \tilde{\M f} & = & 0 \enspace ,
\end{matrix}\label{eq:07}
\end{equation}
where $\tilde{\M f}$ is a $12\times 1$ vector of the form $\tilde{\M f} = [\M f; \lambda f_3; \lambda f_6; \lambda f_9]$, $\M f$ is a vector containing the entries of the fundamental matrix, and $f_i$ is the $i$th entry of the vector $\M f$ . By stacking the rows of $n$ correspondences,~\eqref{eq:07} can be written as
\begin{equation}
(\M C_0 + \lambda' \M C_1 )\tilde{\M f} = \M 0 \enspace .\label{eq:08}
\end{equation}
Since $\M C_0$ is full-rank, we can let $\sigma' =1/ \lambda'$ and rewrite~\eqref{eq:08} as 
\begin{equation}
(\M C_1 + \sigma' \M C_0 )\tilde{\M f} = \M 0 \enspace .\label{eq:09}
\end{equation}
For 12 point correspondences, equation~\eqref{eq:09} defines a generalized eigenvalue problem. 
The solutions to $\sigma'$ can thus be computed as the eigenvalues of the $12\times 12$ matrix 
\begin{equation}
\M D = -{\M C}_0^{-1}{\M C}_1 \enspace .\label{eq:10}
\end{equation}
As noted in~\cite{kukelova2010fast}, there are eight infinite eigenvalues, and at most four finite real solutions to this problem. Here we show that these infinite eigenvalues can be removed in advance. 

Since there are 8 zero columns in $\M C_1$, based on the method proposed in~\cite{kukelova2012polynomial}, these zero columns can be removed together with the corresponding rows. In this case, we only need to find the eigenvalues of a $4\times 4$ matrix. 

Similarly to the equal distortion case, for more than 12 point correspondences, $-{\M C}_0^{-1}{\M C}_1$ is solved using linear least squares.

\begin{figure}[t!]
    \centering
    \includegraphics[width=0.95\textwidth]{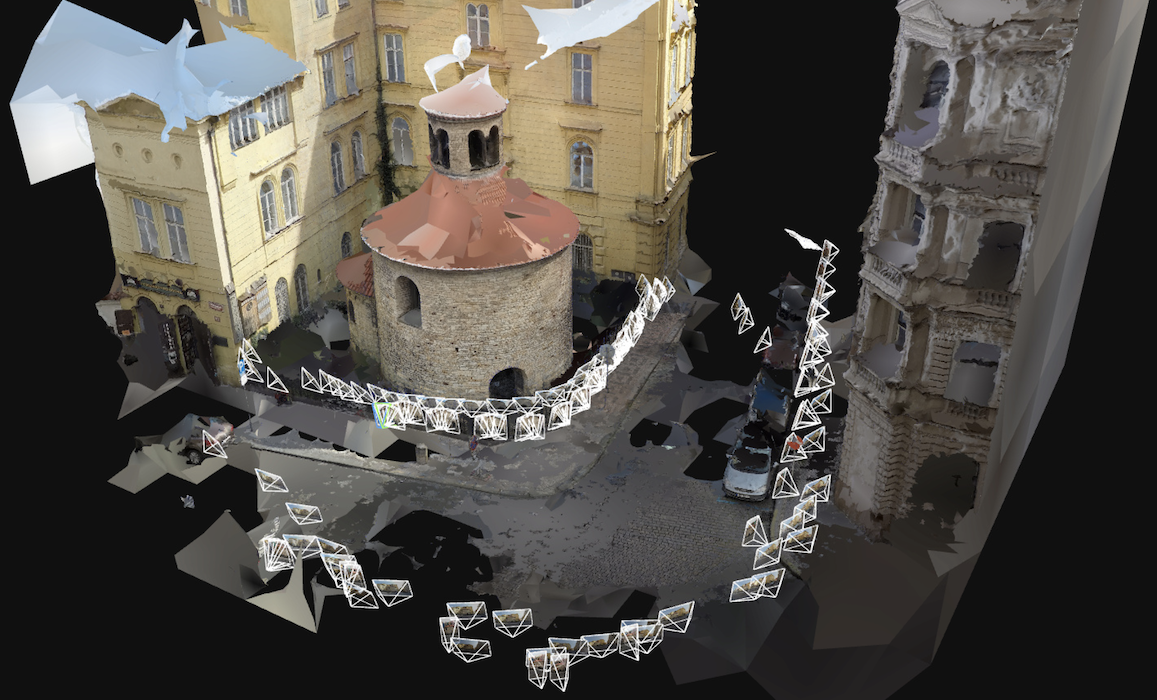}%
    \caption{Visualization of the \ROTUNDA scene. We show a textured mesh of the scene to provide a clearer visualization. We also show the poses of the 157 images of the dataset.}
    \label{fig:datasets_rotunda_3D}
\end{figure}

\section{Inverse of the One-parameter Division Model}
\label{sec:inverse_one_div_model}
The one-parameter division undistortion model is given as
\begin{equation}
    u(\mathbf{x}, \lambda) = [x_d, y_d, 1 + \lambda (x_d^2 + y_d^2)]^\top \enspace .
\end{equation}
We have 
\begin{equation}
     x_u = \frac{x_d}{1 + \lambda (x_d^2 + y_d^2)} \enspace ,\ 
     y_u = \frac{y_d}{1 + \lambda (x_d^2 + y_d^2)} \enspace .
     \label{eq:xud}
\end{equation}
The sum of the squares of the two equations in~\eqref{eq:xud} gives
\begin{equation}
     x_u^2 + y_u^2 = \frac{x_d^2 + y_d^2}{(1 + \lambda (x_d^2 + y_d^2))^2} \enspace. \label{eq:xud2}
\end{equation}
Take the square root of both sides of~\eqref{eq:xud2}, we obtain
\begin{equation}
    r_u = \frac{r_d}{1+\lambda r_d^2} \enspace ,\label{eq:rud}
\end{equation}
where $r_u$ is the distance of the undistorted point from the center of distortion and $r_d$ is the distance of the distorted point from the center of distortion: 
\begin{equation}
    r_u = \sqrt{x_u^2+y_u^2} \text{ and } \ r_d = \sqrt{x_d^2+y_d^2} \enspace .
\end{equation}
Eq.~\eqref{eq:rud} can be rewritten as 
\begin{equation}
    \lambda r_u r_d^2  - r_d + r_u = 0 \enspace .
\end{equation}
There are two solutions to $r_d$. Since $\lambda$ is negative and $r_d$ should be positive, we have
\begin{equation}
    r_d = \frac{1-\sqrt{1-4\lambda r_u^2}}{2\lambda r_u} \enspace . \label{eq:rd_for_inverse}
\end{equation}
Then the coordinates of the distorted point are given by
\begin{equation}
\begin{split}
    x_d &= x_u(1+\lambda r_d^2)\enspace ,\\
    y_d &= y_u(1+\lambda r_d^2)\enspace ,
\end{split} \label{eq:distortion_model}
\end{equation}
with $r_d$ computed as~\eqref{eq:rd_for_inverse}.

\begin{figure}[t!]
    \centering
    \begin{tabular}{c@{\hskip 5mm}c@{\hskip 5mm}c@{\hskip 5mm}c}
    \includegraphics[height=20mm]{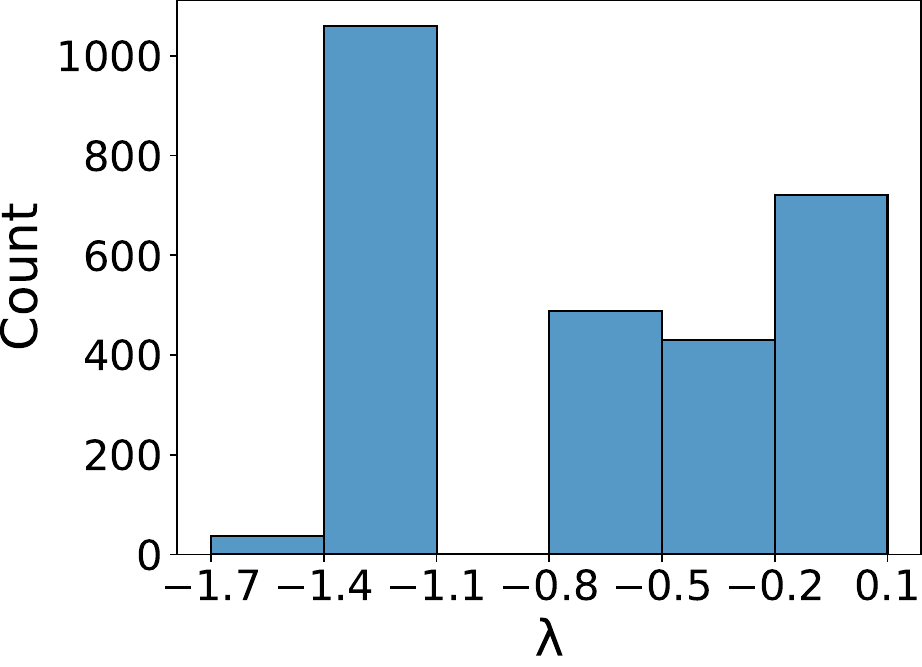} &
    \includegraphics[height=20mm]{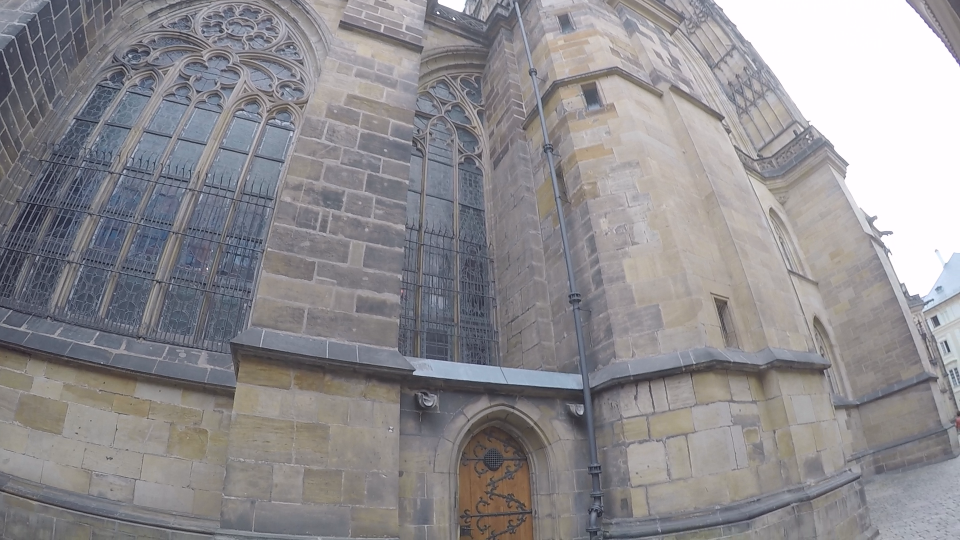} &
    \includegraphics[height=20mm]{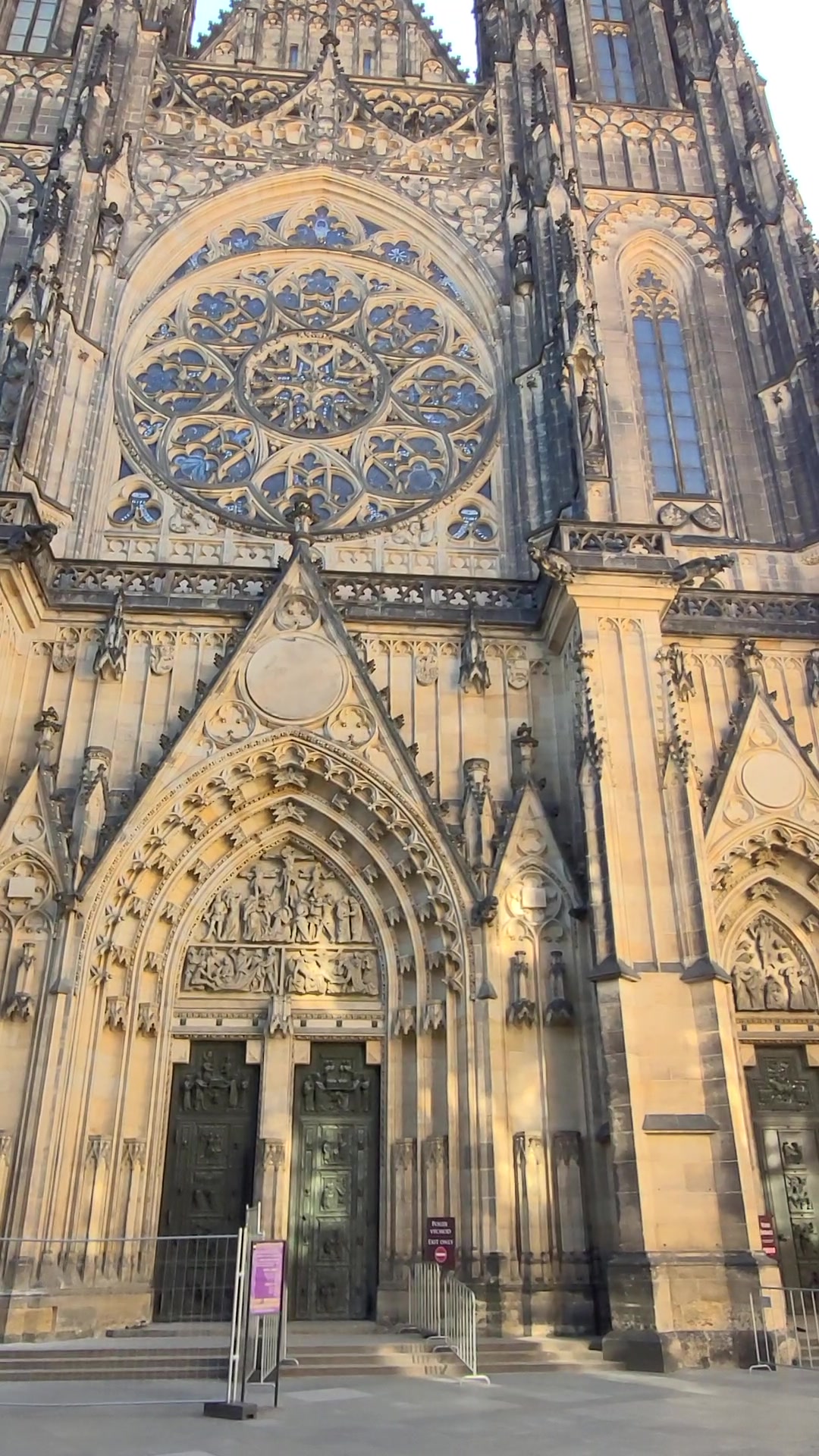} &
    \includegraphics[height=20mm]{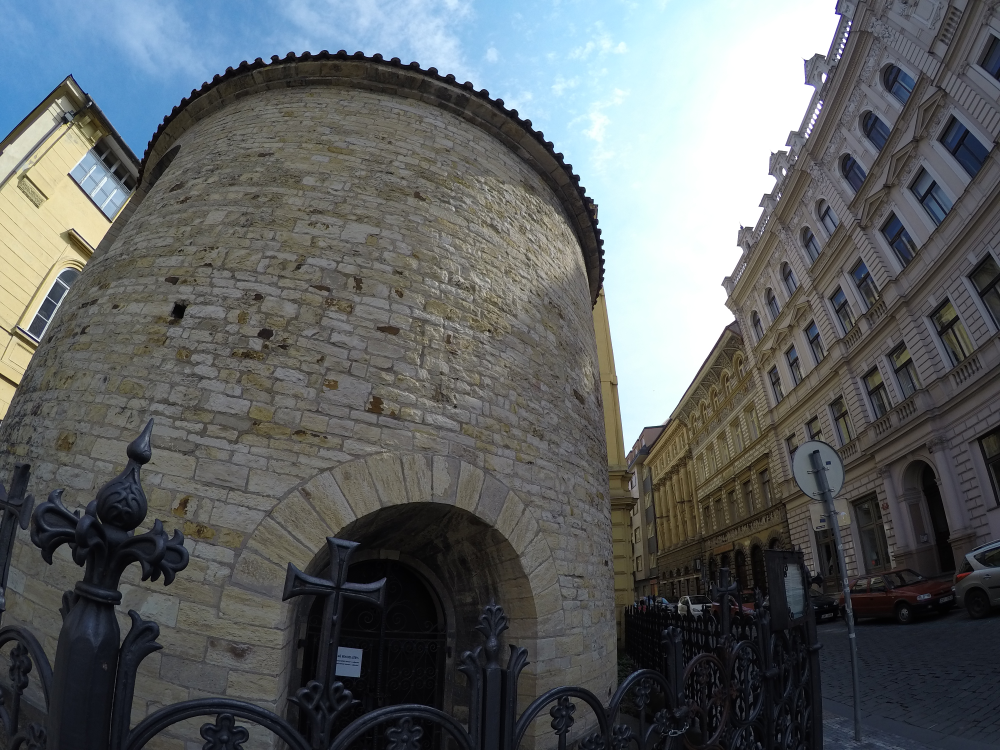}\\    
    a & b & c & d
    \end{tabular}
    \caption{Distribution of $\lambda$ radial undistortion parameters for the  \VITUS scene (a). The parameters were obtained by normalizing the ground truth parameters estimated by RealityCapture. Example images from \VITUS (b, c) and \ROTUNDA (d). }
    \label{fig:datasets1}
\end{figure}

\begin{figure}[t!]
    \centering
    \begin{tabular}{c@{\hskip 5mm}c@{\hskip 5mm}c@{\hskip 5mm}c}
    \includegraphics[width=27mm]{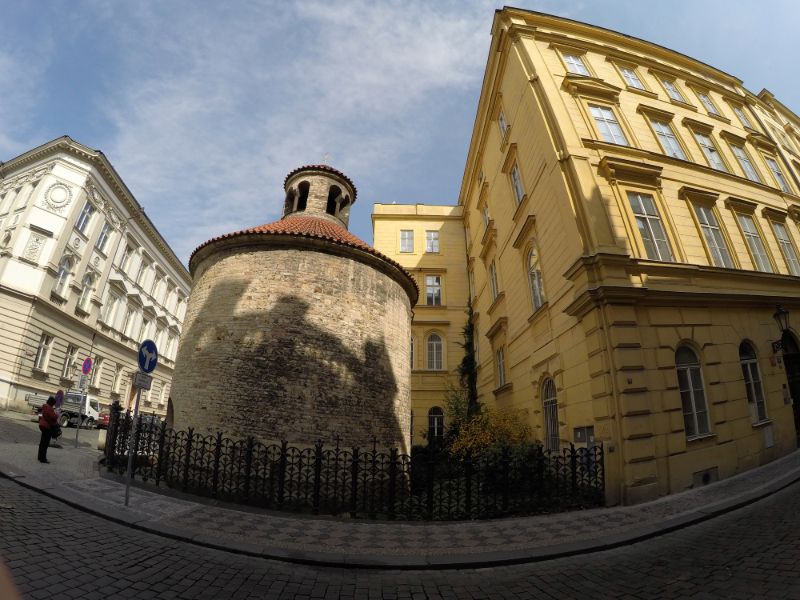} &
    \includegraphics[width=27mm]{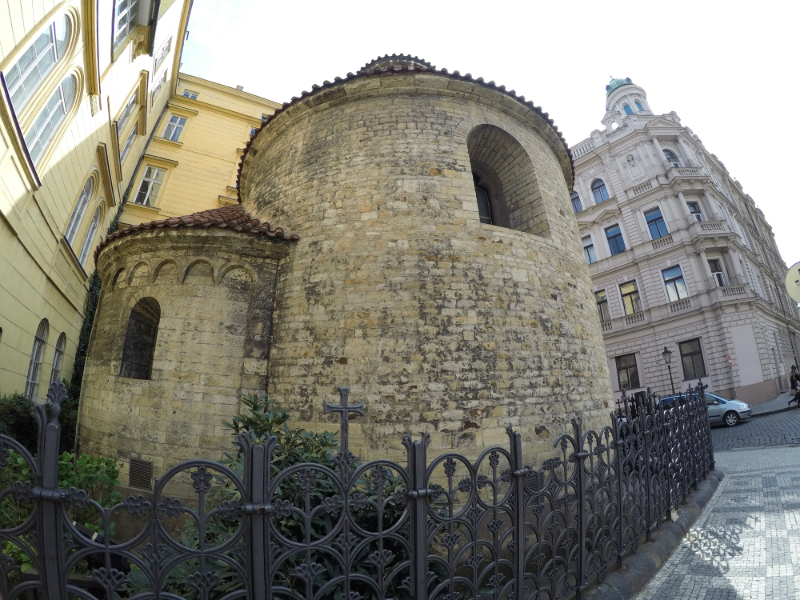} &
    \includegraphics[width=27mm]{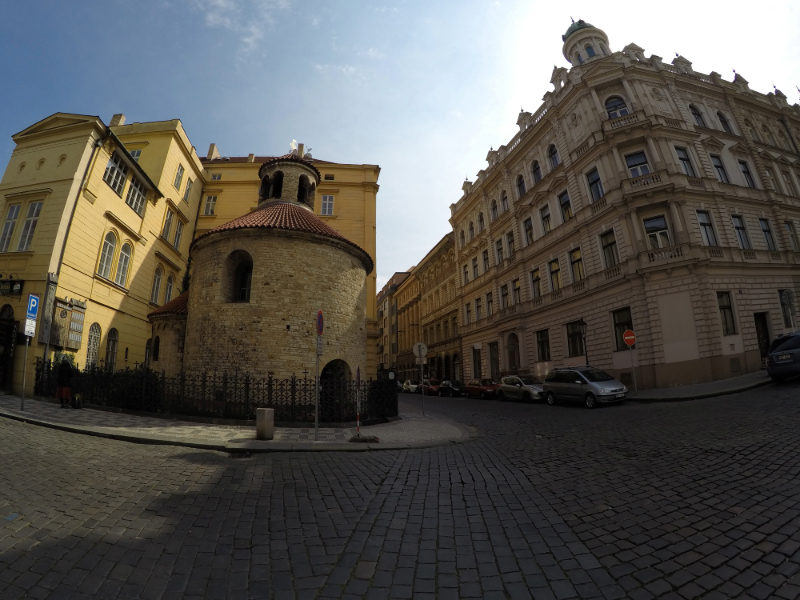} &
    \includegraphics[width=27mm]{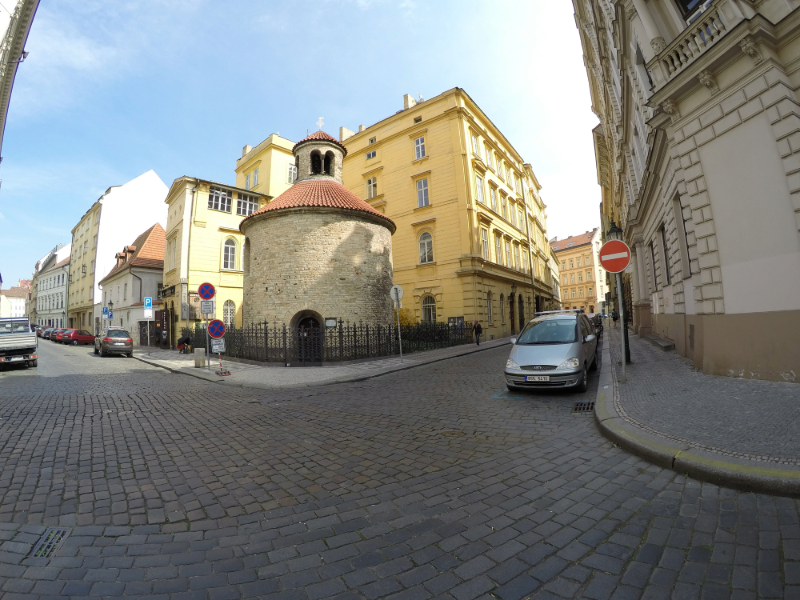}\\ 
    \end{tabular}
    \caption{Example images from the \ROTUNDA scene. }
    \label{fig:datasets2}
\end{figure}

\begin{figure}[t!]
    \centering
    \begin{tabular}{c@{\hskip 5mm}c@{\hskip 5mm}c@{\hskip 5mm}c}
    \includegraphics[width=27mm]{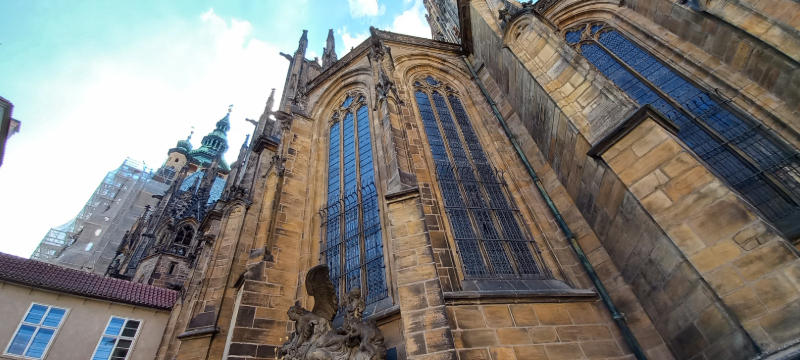} &
    \includegraphics[width=27mm]{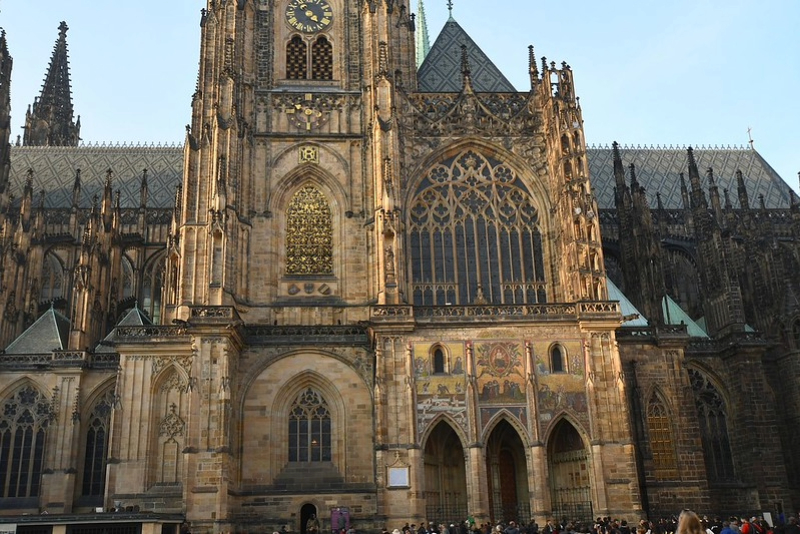} &
    \includegraphics[width=27mm]{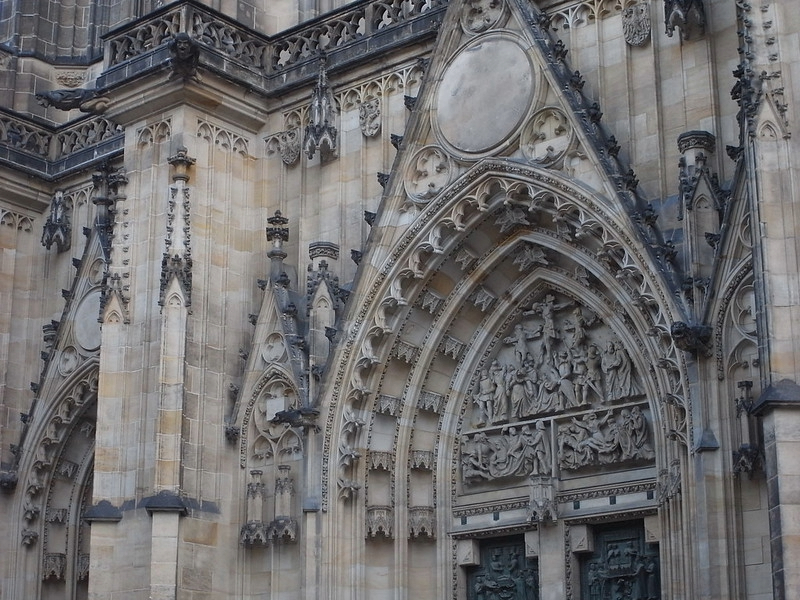} &
    \includegraphics[width=27mm]{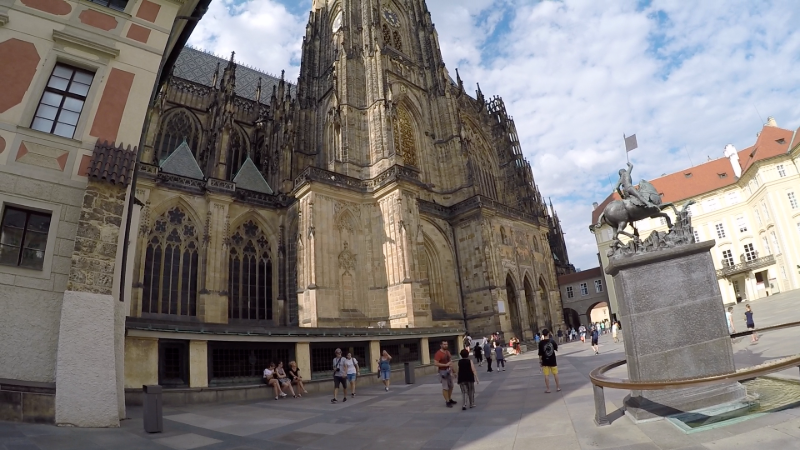}\\
    \end{tabular}
    \caption{Example images from the \VITUS scene.}
    \label{fig:datasets3}
\end{figure}

\begin{figure}[t!h]
    \centering
    \includegraphics[width=0.95\textwidth]{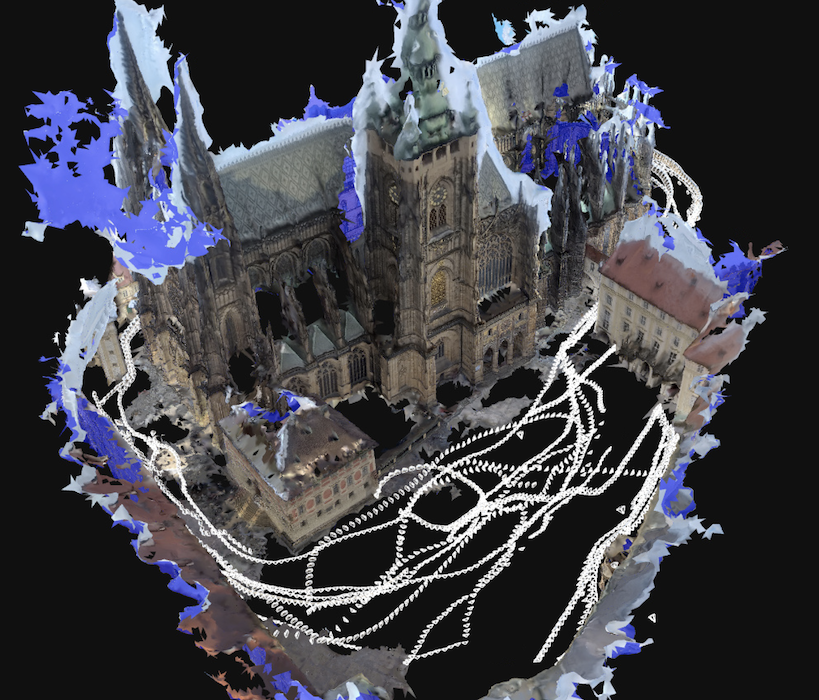}%
    \caption{Visualization of the \VITUS scene. We show a colored mesh of the scene to provide a clearer visualization. We also show the poses of the 2,734 images of the dataset.}
    \label{fig:datasets_cathedral_3D}
\end{figure}

\section{Benchmark Details}
\label{sec:dataset_details}
We created a new benchmark consisting two scenes, \ROTUNDA and \VITUS, for measuring relative pose accuracy under radial distortion. 
For both scenes, we build upon existing images: 
For the \ROTUNDA scene, we use 62 GoPro images, kindly provided by the authors of~\cite{kukelova2015efficient}. 
For the \VITUS scene, we use 655 GoPro images, kindly provided by the authors of~\cite{Sattler_2019_CVPR}. 
For both scenes, we recorded additional images with different cameras. 
To obtain ground truth poses and camera intrinsics including radial distortion for both the original and the newly recorded images, we used the RealityCapture software~\cite{RealityCapture}. 
We configured RealityCapture to estimate the undistortion parameters for the images using the one-parameter division model that we use in all of our experiments, \ie, RealityCapture directly provides ground truth estimates for the undistortion parameters. 
We enforced that images taken by the same camera (with the same field of view) share the same intrinsic camera parameters. 
In the following, we briefly describe both scenes. 

The \ROTUNDA scene contains 157 outdoor images of a historical building captured by two mobile phone cameras (95 images in total) and one GoPro camera (62 images, provided by the authors of~\cite{kukelova2015efficient}). 
The GoPro images were captured using two different settings that affect the field-of-view and image distortion. 
Overall, images were taken with four different $\lambda$ values: $\left\{-1.50, -0.81,  0.02,  0.09\right\}$ (ground truth values provided by RealityCapture after normalization). 
Fig.~\ref{fig:datasets_rotunda_3D} visualizes the \ROTUNDA scene by showing a textured mesh of the scene together with the camera poses of the recorded images. 
Figures~\ref{fig:datasets1} and~\ref{fig:datasets2} show example images from the \ROTUNDA scene.

The \VITUS scene contains 2,734 outdoor images of a historical cathedral, captured by two mobile phone cameras (708 images in total), one GoPro camera (655 images, provided by the authors of~\cite{Sattler_2019_CVPR}), and an Insta360 Ace Pro camera (1,358). Most of the images were extracted from videos captured while walking around the building. In addition, we are using 13 images from Flickr. 
The dataset contains images from cameras with 42 different $\lambda$ parameters. Their distribution is shown in Fig.~\ref{fig:datasets1} along with example images for both scenes. 
Fig.~\ref{fig:datasets3} shows more example images from the 
\VITUS scene. 
Fig.~\ref{fig:datasets_cathedral_3D} visualizes the \VITUS scene by showing a colored mesh of the scene together with the camera poses of the recorded images. 

\begin{figure*}[t]
    \centering
    \begin{subfigure}[t]{0.3\textwidth}
        \centering
        \adjustbox{valign=c}{\includegraphics[width=1.0\textwidth]{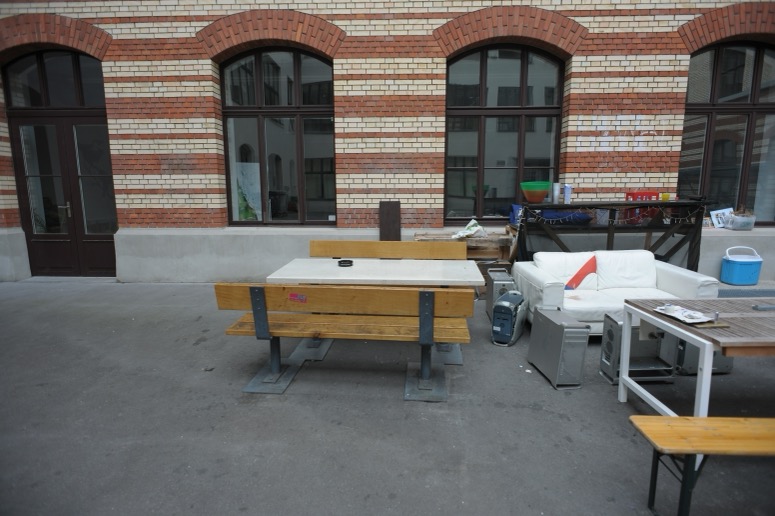}}
    \end{subfigure}
    \begin{subfigure}[t]{0.3\textwidth}
        \centering
        \adjustbox{valign=c}{\includegraphics[width=1.0\textwidth, trim={2.5cm 2cm 2.5cm 2cm}, clip]{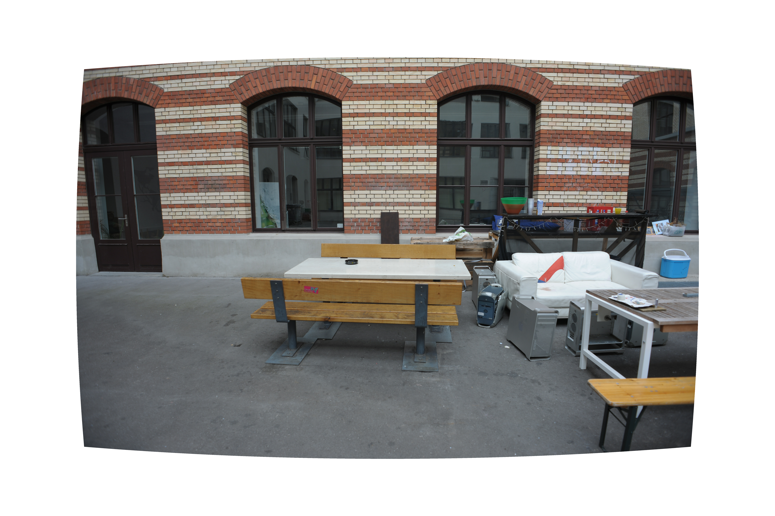}}
    \end{subfigure}
    \begin{subfigure}[t]{0.3\textwidth}
        \centering
        \adjustbox{valign=c}{\includegraphics[width=1.0\textwidth, trim={3.1cm 2.2cm 3.1cm 2.2cm}, clip]{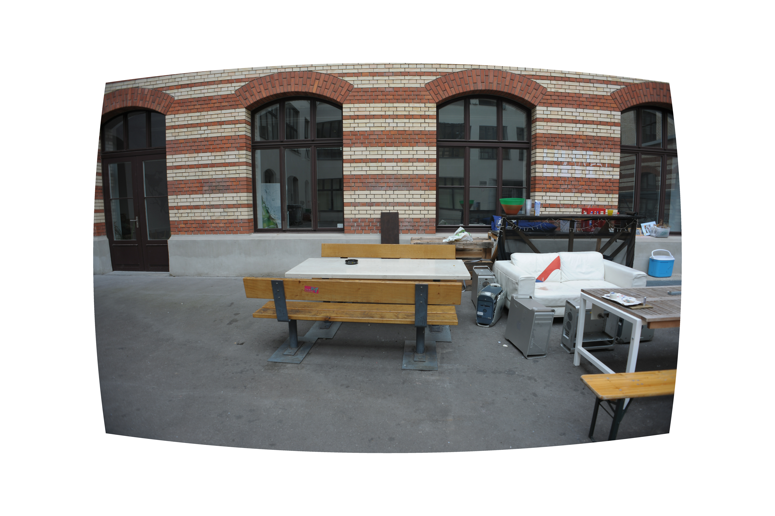}}
    \end{subfigure}
    \\
    \begin{subfigure}[t]{0.3\textwidth}
        \centering
        \adjustbox{valign=c}{\includegraphics[width=1.0\textwidth, trim={3.7cm 2.4cm 3.7cm 2.4cm}, clip]{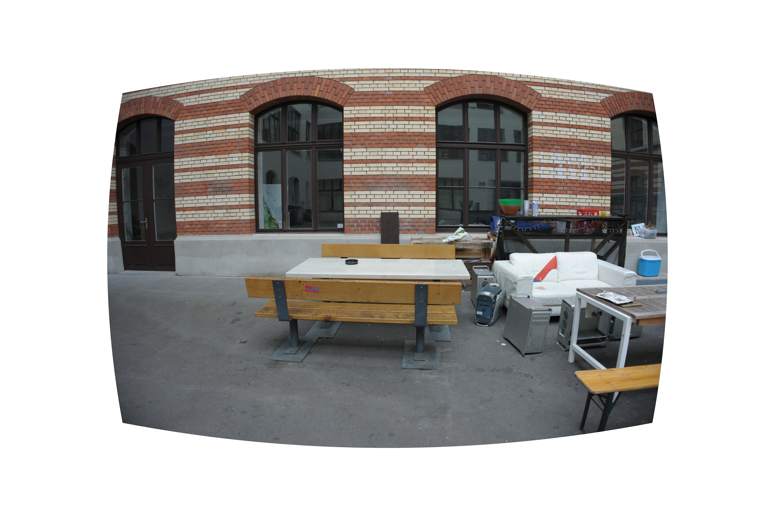}}
    \end{subfigure}
    \begin{subfigure}[t]{0.3\textwidth}
        \centering
        \adjustbox{valign=c}{\includegraphics[width=1.0\textwidth, trim={4.1cm 2.6cm 4.1cm 2.6cm}, clip]{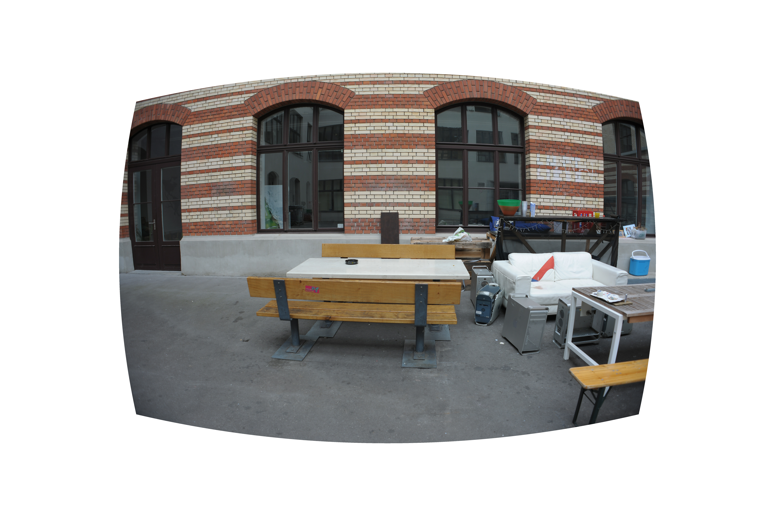}}
    \end{subfigure}
    \begin{subfigure}[t]{0.3\textwidth}
        \centering
        \adjustbox{valign=c}{\includegraphics[width=1.0\textwidth, trim={4.8cm 2.8cm 4.8cm 2.8cm}, clip]{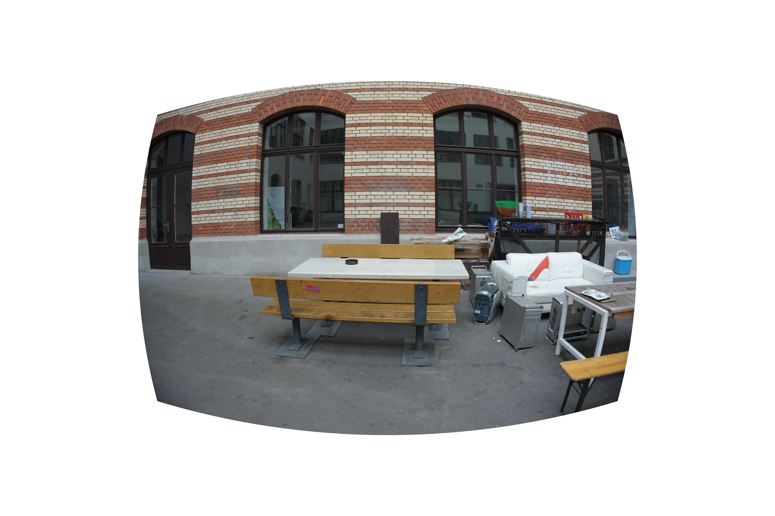}}
    \end{subfigure}
    \caption{Example image of the Courtyard scene from the \ETH dataset, w.r.t. synthetically added distortions. We use undistortion parameters $\{0, -0.3, -0.6, -0.9, -1.2, -1.8\}$ and compute the distorted images using the inverse of the one-parameter division undistortion model. The corresponding images are shown in left-to-right and top-to-bottom order. For the parameters, we consider the one-parameter division model and $[-0.5, 0.5]$ coordinate normalization.}
    \label{fig:courtyard_box}
\end{figure*}

\section{Experiments}
\label{sec:experiments_details}
\subsubsection{Example visualization of varying levels of distortion.} In order to provide a more intuitive understanding of the range of distortion used in this work, Fig.~\ref{fig:courtyard_box} visualizes the impact of various undistortion parameters on an image. 
As in the main paper, the parameters are defined for normalized coordinates in the range $[-0.5, 0.5]^2$. 

\subsubsection{Run-time measurements.} We report run-times for Poselib using a 2 GHz Intel Xeon Gold 6338 CPU, and run-times for GC-RANSAC using a 2.2 GHz Intel Xeon Silver 4214 CPU.

\begin{table*}[t!]
    \centering
    \caption{Results on image pairs from the \VITUS scene, using both the GC-RANSAC and Poselib frameworks and SIFT+LG matches.}
    \setlength{\tabcolsep}{4.8pt}
    \resizebox{1.0\linewidth}{!}{
    \begin{tabular}{ c | r r c | c c c c c | c c | c}
    \toprule
        & & & & \multicolumn{7}{c}{GC-RANSAC - \VITUS} \\
        \midrule
        & Minimal & Non-Minimal & Sample & AVG $(^\circ)$ $\downarrow$ & MED $(^\circ)$ $\downarrow$ & AUC@5 $\uparrow$ & @10 & @20 & AVG $\epsilon(\lambda)$ $\downarrow$ & MED $\epsilon(\lambda)$ $\downarrow$ & Time (ms) $\downarrow$ \\
        \midrule
        \multirow{11}{*}{\rotatebox[origin=c]{90}{$\lambda_1 = \lambda_2$}} & 7pt \F & 7pt \F & 0 & 15.77 & 7.67 & 0.19 & 0.33 & 0.50 & 0.69 & 0.72 & \phantom{1}160.02\\
        & 7pt \F & 8pt \F & 0 & 16.39 & 8.13 & 0.19 & 0.32 & 0.49 & 0.69 & 0.72 & \phantom{1}\underline{120.49}\\
        & 7pt \F & 9pt \Fk & 0 & 14.10 & 5.68 & 0.21 & 0.38 & 0.55 & 0.44 & 0.25 & \phantom{1}\textbf{101.09}\\
        & 7pt \F & 12pt \Fkk & 0 & 17.86 & 9.96 & 0.09 & 0.23 & 0.42 & 0.64 & 0.62 & \phantom{1}140.33\\
        & 7pt \F & 9pt \Fk & $\{-1.2, -0.6, 0\}$ & \textbf{12.81} & \textbf{4.46} & \textbf{0.25} & \textbf{0.43} & \textbf{0.59} & \underline{0.36} & \underline{0.20} & \phantom{1}161.83 \\
        & 7pt \F & 12pt \Fkk & $\{-1.2, -0.6, 0\}$ & 14.10 & 6.15 & 0.18 & 0.35 & 0.54 & \textbf{0.29} & \textbf{0.19} & \phantom{1}170.88\\
        \cmidrule{2-12}
        & 8pt \Fk & 9pt \Fk & \ding{55} & \underline{13.02} & \underline{4.78} & \underline{0.24} & \underline{0.42} & \underline{0.58} & 0.40 & 0.24 & \phantom{1}603.67\\
        & 8pt \Fk & 12pt \Fkk & \ding{55} & 14.43 & 6.59 & 0.17 & 0.34 & 0.52 & 0.39 & 0.25 & \phantom{1}650.98\\
        \cmidrule{2-12}
        & 9pt \Fk & 9pt \Fk & \ding{55} & 13.21 & 5.19 & 0.22 & 0.40 & 0.57 & 0.43 & 0.27 & \phantom{1}251.00\\
        & 9pt \Fk & 12pt \Fkk & \ding{55} & 14.88 & 7.16 & 0.15 & 0.31 & 0.51 & 0.46 & 0.30 & \phantom{1}270.76\\
        \midrule
        \midrule
        \multirow{6}{*}{\rotatebox[origin=c]{90}{$\lambda_1 \neq \lambda_2$}} & 7pt \F & 7pt \F & 0 & 27.79 & 18.45 & 0.08 & 0.16 & 0.29 & 0.65 & 0.62 & \phantom{1}\underline{173.91}\\
        & 7pt \F & 8pt \F & 0 & 29.61 & 21.58 & \underline{0.07} & 0.14 & 0.26 & 0.65 & 0.62 & \phantom{1}\textbf{140.55}\\
        & 7pt \F & 12pt \Fkk & $\{-1.2, -0.6, 0\}$ & \textbf{23.27} & \textbf{12.72} & \textbf{0.09} & \textbf{0.21} & \textbf{0.37} & \textbf{0.36} & \textbf{0.20} & \phantom{1}520.49\\
        \cmidrule{2-12}
        & 9pt \Fkk & 12pt \Fkk & \ding{55} & \underline{24.90} & \underline{14.68} & \underline{0.07} & \underline{0.19} & \underline{0.34} & \underline{0.51} & \underline{0.38} & 3417.09\\
        & 10pt \Fkk & 12pt \Fkk & \ding{55} & 25.44 & 14.63 & \underline{0.07} & \underline{0.19} & \underline{0.34} & 0.54 & 0.41 & \phantom{1}180.72\\
        \bottomrule
        \toprule
    & & & & \multicolumn{7}{c}{Poselib - \VITUS} \\
    \midrule
    & Minimal & Refinement & Sample & AVG $(^\circ)$ $\downarrow$ & MED $(^\circ)$ $\downarrow$ & AUC@5 $\uparrow$ & @10 & @20 & AVG $\epsilon(\lambda)$ $\downarrow$ & MED $\epsilon(\lambda)$ $\downarrow$ & Time (ms) $\downarrow$ \\
    \midrule
    \multirow{12}{*}{\rotatebox[origin=c]{90}{$\lambda_1 = \lambda_2$}}     & 7pt \F & \F & 0 & 20.30 & 6.83 & 0.30 & 0.42 & 0.57 & 0.69 & 0.72 & \phantom{11}\textbf{12.99} \\
    & 7pt \F & \F + \Fk & 0 & 18.24 & 4.97 & 0.37 & 0.49 & 0.62 & 0.49 & 0.27 & \phantom{11}\underline{22.89} \\
    & 7pt \F & \F + \Fkk & 0 & 18.12 & 4.92 & 0.37 & 0.49 & 0.62 & 0.50 & 0.29 & \phantom{11}23.71 \\
    & 7pt \F & \Fk & $0$ & 16.16 & 3.67 & 0.42 & 0.54 & 0.66 & 0.34 & 0.15 & \phantom{11}64.82 \\
    & 7pt \F & \Fkk & $0$ & 16.27 & 3.74 & 0.42 & 0.54 & 0.67 & 0.35 & 0.17 & \phantom{11}64.85 \\
    & 7pt \F & \Fk & $\{-1.2, -0.6, 0\}$ & \textbf{14.36} & \textbf{2.35} & \textbf{0.52} & \textbf{0.63} & \textbf{0.73} & \textbf{0.25} & \textbf{0.09} & \phantom{1}108.10 \\ 
    & 7pt \F & \Fkk & $\{-1.2, -0.6, 0\}$ & \underline{14.69} & \underline{2.39} & \underline{0.51} & \underline{0.62} & \underline{0.72} & \underline{0.26} & 0.11 & \phantom{1}258.28 \\
    \cmidrule{2-12}
    & 8pt \Fk & \Fk & \ding{55} & 14.71 & 2.63 & 0.49 & 0.61 & 0.71 & 0.27 & \underline{0.10} & \phantom{1}401.45 \\
    & 9pt \Fk & \Fk & \ding{55} & 17.00 & 4.31 & 0.38 & 0.52 & 0.65 & 0.37 & 0.19 & \phantom{1}143.71 \\
     \cmidrule{2-12}
    & 9pt \Fkk & \Fkk & \ding{55} & 19.67 & 5.87 & 0.33 & 0.45 & 0.59 & 0.47 & 0.27 & 1182.12 \\
    & 10pt \Fkk & \Fkk & \ding{55} & 17.82 & 4.74 & 0.36 & 0.49 & 0.63 & 0.37 & 0.23 & \phantom{11}60.39 \\
    \midrule
    \midrule
    \multirow{6}{*}{\rotatebox[origin=c]{90}{$\lambda_1 \neq \lambda_2$}} & 7pt \F & \F & 0 & 35.34 & 17.58 & 0.13 & 0.21 & 0.34 & 0.65 & 0.68 & \phantom{11}\textbf{12.10} \\
    & 7pt \F & \F + \Fkk & 0 & 33.38 & 16.03 & 0.16 & 0.24 & 0.37 & 0.61 & 0.58 & \phantom{11}\underline{19.70} \\
    & 7pt \F & \Fkk & $0$ & \underline{30.33} & \underline{13.30} & \underline{0.19} & \underline{0.28} & \underline{0.41} & 0.52 & \underline{0.42} & \phantom{11}66.39 \\
    & 7pt \F & \Fkk & $\{-1.2, -0.6, 0\}$ & \textbf{24.45} & \phantom{1}\textbf{7.90} & \textbf{0.29} & \textbf{0.39} & \textbf{0.51} & \textbf{0.41} & \textbf{0.28} & \phantom{1}291.60 \\ \cmidrule{2-12}
    & 9pt \Fkk & \Fkk & \ding{55} & 32.13 & 15.44 & 0.16 & 0.25 & 0.37 & 0.57 & 0.46 & 1953.51 \\
    & 10pt \Fkk & \Fkk & \ding{55} & 30.39 & 13.41 & 0.18 & \underline{0.28} & 0.40 & \underline{0.51} & 0.42 & \phantom{11}85.19 \\
    \bottomrule
    \end{tabular}
    }
    \label{tab:rotunda}
\end{table*}

\subsubsection{Feature Matching.}
For experiments in the main paper, we use SuperPoint~\cite{detone2018superpoint} features extracted on images without resizing. We only kept at most 2048 features and matched them with LightGlue (LG)~\cite{lindenberger2023lightglue}. We only considered images with sufficient overlap quantified by the co-visibility constraint proposed in~\cite{IMC2020}. We also only kept image pairs, which resulted in at least 20 matches.

We also used the same strategy with SIFT~\cite{Lowe_sift} local features. Note that due to the criterion of at least 20 matches for each image pair, the sets of image pairs for all datasets differ slightly. 
We report results for SIFT+LG for both GC-RANSAC and Poselib frameworks on the 
\VITUS scene (see 
Tab.~\ref{tab:rotunda}). 
We used 10,000 randomly sampled image pairs for both $\lambda_1 = \lambda_2$ and $\lambda_1 \neq \lambda_2$. 

In the tables, the reported statistics are the following: the average and median pose error, \ie, the maximum of rotation and translation error, in degrees, the Area Under Recall Curve (AUC) at 5$^\circ$, 10$^\circ$, and 20$^\circ$ thresholds of the pose error, the average and median absolute error of the undistortion parameter, and the average run-time of RANSAC. Bold font denotes the best values, while underline indicates the second best. 

Compared to Tab.~1 in the main paper, which shows the results of the same experiment with SuperPoint+LG matches, the main conclusions drawn in the main paper remain valid: 
The sampling-based approach that combines the 7pt \F \ solver with three sampled undistortion parameters performs comparably or better than the best-performing approaches based on dedicated minimal solvers that estimate radial undistortion parameters. 
At the same time, the sampling-based approach is significantly faster. 
Compared to non-minimal radial distortion solvers that are more efficient than the sampling strategy, the sampling-based approach is (significantly) more accurate.

\begin{table*}[ht]
    \centering
    \caption{Results for $\{($\textit{Scenario A}) \textit{Wild}, (\textit{Scenario B}) \textit{Small distortion}, (\textit{Scenario C}) \textit{Visible distortion}$\}$ on 12 scenes from the  \ETH dataset, using the GC-RANSAC framework and SP+LG matches.}
    \setlength{\tabcolsep}{4.8pt}
    \resizebox{1.0\linewidth}{!}{
    \begin{tabular}{ c | r r c | c c c c c | c c | c}
        \toprule
        & & & & \multicolumn{8}{c}{GC-RANSAC - \ETH - \textit{Scenario A} \textit{Wild}} \\
        \midrule
        & Minimal & Non-Minimal & Sample & AVG $(^\circ)$ $\downarrow$ & MED $(^\circ)$ $\downarrow$ & AUC@5 $\uparrow$ & @10 & @20 & AVG $\epsilon(\lambda)$ $\downarrow$ & MED $\epsilon(\lambda)$ $\downarrow$ & Time (ms) $\downarrow$ \\
        \midrule
        \multirow{11}{*}{\rotatebox[origin=c]{90}{$\lambda_1 = \lambda_2$}} & 7pt \F & 7pt \F & 0 & 28.38 & 16.58 & 0.09 & 0.18 & 0.31 & 0.87 & 0.85 & \phantom{1}520.07\\
        & 7pt \F & 8pt \F & 0 & 27.48 & 16.05 & 0.09 & 0.18 & 0.32 & 0.87 & 0.85 & \phantom{1}\underline{460.82}\\
        & 7pt \F & 9pt \Fk & 0 & 21.21 & \phantom{1}6.41 & 0.27 & 0.39 & 0.51 & 0.44 & 0.19 & \bf \phantom{1}302.73 \\
        & 7pt \F & 12pt \Fkk & 0 & 27.39 & 15.93 & 0.10 & 0.20 & 0.33 & 0.81 & 0.78 & \phantom{1}530.99 \\
        & 7pt \F & 9pt \Fk & $\{-1.2, -0.6, 0\}$ & 17.55 & \phantom{1}3.21 & 0.34 & 0.48 & 0.60 & 0.25 & 0.12 & \phantom{1}540.11\\
        & 7pt \F & 12pt \Fkk & $\{-1.2, -0.6, 0\}$ & 17.94 & \phantom{1}4.73 & 0.25 & 0.41 & 0.56 & 0.27 & 0.20 & \phantom{1}631.43\\
        \cmidrule{2-12}
        & 8pt \Fk & 9pt \Fk & \ding{55} & \bf 16.28 & \phantom{1}\bf 2.29 & \bf 0.39 & \bf 0.52 & \bf 0.63 & \underline{0.23} & \bf 0.07 & 1930.03 \\
        & 8pt \Fk & 12pt \Fkk & \ding{55} & \underline{16.53} & \phantom{1}\underline{3.02} & \underline{0.34} & \underline{0.49} & \underline{0.61} & \bf 0.21 & \underline{0.08} & 1972.48 \\
        \cmidrule{2-12}
        & 9pt \Fk & 9pt \Fk & \ding{55} & 18.82 & \phantom{1}3.79 & 0.32 & 0.46 & 0.57 & 0.32 & 0.12 & \phantom{1}610.00\\
        & 9pt \Fk & 12pt \Fkk & \ding{55} & 19.37 & \phantom{1}5.51 & 0.25 & 0.39 & 0.53 & 0.37 & 0.20 & \phantom{1}650.30\\
        \midrule
        \midrule
        \multirow{5}{*}{\rotatebox[origin=c]{90}{$\lambda_1 \neq \lambda_2$}} & 7pt \F & 7pt \F & 0 & 34.11 & 23.24 & 0.05 & 0.10 & 0.22 & 0.86 & 0.86 & \phantom{1}870.98 \\
        & 7pt \F & 8pt \F & 0 & 34.14 & 23.67 & 0.05 & 0.10 & 0.22 & 0.86 & 0.86 & \phantom{1}\underline{772.06} \\
        & 7pt \F & 12pt \Fkk & $\{-1.2, -0.6, 0\}$ & 20.29 & \phantom{1}6.88 & 0.20 & 0.34 & 0.50 & 0.35 & 0.29 & 2413.01 \\
        \cmidrule{2-12}
        & 9pt \Fkk & 12pt \Fkk & \ding{55} & \bf 18.49 & \phantom{1}\bf 4.77 & \bf 0.26 & \bf 0.41 & \bf 0.56 & \bf 0.31 & \bf 0.19 & 5388.56\\
        & 10pt \Fkk & 12pt \Fkk & \ding{55} & \underline{19.29} & \phantom{1}\underline{5.11} & \underline{0.26} & \underline{0.41} & \underline{0.55} & \underline{0.33} & \underline{0.22} & \phantom{1}\bf 250.11\\
        \bottomrule
        \toprule
        & & & & \multicolumn{8}{c}{GC-RANSAC - \ETH - \textit{Scenario B} \textit{Small distortion}} \\
        \midrule
        & Minimal & Non-Minimal & Sample & AVG $(^\circ)$ $\downarrow$ & MED $(^\circ)$ $\downarrow$ & AUC@5 $\uparrow$ & @10 & @20 & AVG $\epsilon(\lambda)$ $\downarrow$ & MED $\epsilon(\lambda)$ $\downarrow$ & Time (ms) $\downarrow$ \\
        \midrule
        \multirow{9}{*}{\rotatebox[origin=c]{90}{$\lambda_1 = \lambda_2$}} & 7pt \F & 7pt \F & 0 & 16.89 & 3.66 & 0.30 & 0.47 & \underline{0.61} & \textbf{0.15} & 0.15 & \phantom{1}\underline{280.81} \\
        & 7pt \F & 8pt \F & 0 & \underline{16.62} & 3.42 & 0.31 & 0.48 & \underline{0.61} & \textbf{0.15} & 0.15 & \phantom{1}\textbf{241.37} \\
        & 7pt \F & 9pt \Fk & 0 & 17.81 & \underline{3.01} & \underline{0.35} & 0.49 & 0.60 & \underline{0.22} & 0.10 & \phantom{1}\textbf{247.41} \\
        & 7pt \F & 12pt \Fkk & 0 & 17.44 & 4.41 & 0.25 & 0.43 & 0.58 & \textbf{0.15} & 0.15 & \phantom{1}293.01 \\
        \cmidrule{2-12}
        & 8pt \Fk & 9pt \Fk & \ding{55} & 16.80 & 2.38 & \textbf{0.38} & \textbf{0.52} & \textbf{0.62} & 0.26 & \textbf{0.06} & 2012.09 \\
        & 8pt \Fk & 12pt \Fkk & \ding{55} & \textbf{16.05} & 3.12 & 0.34 & \underline{0.50} & \textbf{0.62} & \underline{0.22} & \textbf{0.06} & 2070.99\\
        \cmidrule{2-12}
        & 9pt \Fk & 9pt \Fk & \ding{55} & 18.24 & \textbf{2.93} & \underline{0.35} & 0.49 & 0.59 & 0.25 & \underline{0.08} & \phantom{1}610.32\\
        & 9pt \Fk & 12pt \Fkk & \ding{55} & 17.69 & 3.76 & 0.30 & 0.46 & 0.59 & 0.23 & 0.10 & \phantom{1}640.67\\
        \midrule
        \midrule
        \multirow{4}{*}{\rotatebox[origin=c]{90}{$\lambda_1 \neq \lambda_2$}} & 7pt \F & 7pt \F & 0 & 20.49 & 5.68 & 0.21 & 0.37 & 0.52 & \textbf{0.15} & \underline{0.15} & \phantom{1}343.89\\
        & 7pt \F & 8pt \F & 0 & 20.56 & 5.60 & 0.22 & 0.38 & 0.52 & \textbf{0.15} & \underline{0.15} & \phantom{1}\underline{303.70}\\        
        \cmidrule{2-12}
        & 9pt \Fkk & 12pt \Fkk & \ding{55} & \textbf{18.31} & \textbf{4.55} & \textbf{0.27} & \textbf{0.42} & \textbf{0.56} & 0.30 & \textbf{0.13} & 5760.08\\
        & 10pt \Fkk & 12pt \Fkk & \ding{55} & \underline{19.05} & \underline{5.33} & \underline{0.26} & \underline{0.40} & \underline{0.54} & 0.36 & 0.17 & \phantom{1}\textbf{270.11}\\
        \bottomrule
        \toprule
        & & & & \multicolumn{8}{c}{GC-RANSAC - \ETH - \textit{Scenario C} \textit{Visible distortion}} \\
        \midrule
        & Minimal & Non-Minimal & Sample & AVG $(^\circ)$ $\downarrow$ & MED $(^\circ)$ $\downarrow$ & AUC@5 $\uparrow$ & @10 & @20 & AVG $\epsilon(\lambda)$ $\downarrow$ & MED $\epsilon(\lambda)$ $\downarrow$ & Time (ms) $\downarrow$ \\
        \midrule
        \multirow{9}{*}{\rotatebox[origin=c]{90}{$\lambda_1 = \lambda_2$}} & 7pt \F & 9pt \Fk & $-0.9$ & \phantom{1}\textbf{7.67} & \phantom{1}3.96 & 0.31 & 0.46 & 0.58 & 0.25 & 0.14 & \phantom{1}\textbf{230.37}\\
        & 7pt \F & 12pt \Fkk & $-0.9$ & 19.48 & \phantom{1}6.30 & 0.20 & 0.36 & 0.52 & 0.30 & 0.27 & \phantom{1}\underline{291.12}\\
        & 7pt \F & 9pt \Fk & $\{-1.2, -0.9, -0.6\}$ & 16.36 & \phantom{1}\underline{2.73} & \underline{0.36} & \underline{0.50} & \underline{0.62} & \textbf{0.20} & 0.10 & \phantom{1}481.64\\
        & 7pt \F & 12pt \Fkk & $\{-1.2, -0.9, -0.6\}$ & 17.32 & \phantom{1}3.88 & 0.29 & 0.45 & 0.59 & \textbf{0.20} & 0.13 & \phantom{1}560.19\\
        \cmidrule{2-12}
        & 8pt \Fk & 9pt \Fk & \ding{55} &\underline{15.51} & \phantom{1}\textbf{2.41} & \textbf{0.38} & \textbf{0.52} & \textbf{0.63} & 0.22 & \textbf{0.07} & 1883.03\\
        & 8pt \Fk & 12pt \Fkk & \ding{55} & 15.73 & \phantom{1}2.88 & 0.35 & \underline{0.50} & \underline{0.62} & \underline{0.21} & \underline{0.08} & 1840.12\\
        \cmidrule{2-12}
        & 9pt \Fk & 9pt \Fk & \ding{55} & 19.14 & \phantom{1}3.97 & 0.31 & 0.45 & 0.56 & 0.34 & 0.14 & \phantom{1}580.93\\
        & 9pt \Fk & 12pt \Fkk & \ding{55} & 19.81 & \phantom{1}6.17 & 0.24 & 0.38 & 0.52 & 0.40 & 0.22 & \phantom{1}630.49\\
        \midrule
        \midrule
        \multirow{5}{*}{\rotatebox[origin=c]{90}{$\lambda_1 \neq \lambda_2$}} & 7pt \F & 7pt \F & $-0.9$ & 26.80 & 12.35 & 0.11 & 0.23 & 0.38 & 0.33 & 0.31 & \phantom{1}560.78\\
        & 7pt \F & 8pt \F & $-0.9$ & 26.96 & 12.10 & 0.11 & 0.22 & 0.38 & 0.33 & 0.31 & \phantom{1}\underline{470.90}\\
        & 7pt \F & 12pt \Fkk & $\{-1.2, -0.9, -0.6\}$ & \textbf{17.91} & \phantom{1}5.23 & 0.23 & 0.40 & \underline{0.55} & \textbf{0.25} & 0.20 & 1939.30\\
        \cmidrule{2-12}
        & 9pt \Fkk & 12pt \Fkk & \ding{55} & \underline{18.62} & \phantom{1}\textbf{4.53} & \textbf{0.28} & \textbf{0.43} & \textbf{0.56} & \underline{0.32} & 0\textbf{0.15} & 5210.00\\
        & 10pt \Fkk & 12pt \Fkk & \ding{55} & 19.23 & \phantom{1}\underline{4.93} & \underline{0.27} & \underline{0.41} & \underline{0.55} & 0.35 & \underline{0.18} & \phantom{1}\textbf{250.02}\\
        \bottomrule
    \end{tabular}
     }
    \label{tab:eth3d_all_scenaria_gcr}
\end{table*}

\subsubsection{Prior knowledge about cameras.}
Considering the three scenarios with regard to prior knowledge of cameras that are discussed in the main paper, \ie, (\textit{Scenario A}) \textit{Wild}, (\textit{Scenario B}) \textit{Small distortion}, and (\textit{Scenario C}) \textit{Visible distortion}, we present results for the \ETH dataset. The results for GC-RANSAC are shown in Tab.~\ref{tab:eth3d_all_scenaria_gcr}. 
Result for Poselib are shown in Tab.~\ref{tab:poselib_eth3d_scenaria}. 
For GC-RANSAC, we also show results of (\textit{Scenario A}) \textit{Wild}, on three scenes from \Phototourism, namely Sacre Coeur, St. Peter's Square, and Temple Nara Japan. The results are reported in Tab.~\ref{tab:phototourism_st_peter_square_main_splg}  and~\ref{tab:phototourism_sacre_coeur_main_siftlg}, for SP+LG and SIFT+LG matches, respectively. 

Again, we observe that sampling-based strategies offer a very good trade-off between pose accuracy and run-time efficiency.

\subsubsection{Biased prior knowledge about cameras.} 
In the special case where we know that the images considered are captured using a set of cameras that have a similar lens, \ie, undistortion parameters lie in a small interval, solvers that sample undistortion parameters to estimate the pose can get a noticeable performance boost. 
We conducted two synthetic experiments on images from 12 scenes from the \ETH dataset~\cite{Schops_2017_CVPR}, in which we applied uniformly random distortion from the ranges (1) $[-1.7, -1.3]$, and (2) $[-0.9, -0.5]$. 
We refer to the former as \textbf{Wide-angle lens}, and to the latter as \textbf{Medium-angle lens}. 
The tested sampling solvers consider only one undistortion parameter, in particular the mean value of the undistortion interval in the corresponding setup. The results are reported in Tab.~\ref{tab:eth_bias_wide}  and~\ref{tab:eth3d_bias_medium}, for the \textbf{Wide-angle lens}, and \textbf{Medium-angle lens} setups, respectively. 
In both scenarios, we consider the case of equal unknown radial distortion (\ie, $\lambda_1 = \lambda_2$). 
As can be seen, combining the 7pt \F \ solver with a sampled radial undistortion parameter leads to  a comparable accuracy as employing the minimal 8pt \Fk \ solver at much faster run-times. 
At the same time, the combination of the sample and the 7pt \F \ solver is more accurate and faster compared to using the non-minimal 9pt \Fk \ solver.

\begin{figure*}[t]
    \centering
    \begin{subfigure}[t]{0.48\textwidth}
	    \includegraphics[width=1.0\textwidth]{figures/distortion/eth3d_k.png}
	\end{subfigure}
 \begin{subfigure}[t]{0.48\textwidth}
	    \includegraphics[width=1.0\textwidth]{figures/pose/eth3d_p.png}
	\end{subfigure}
    \begin{subfigure}[t]{0.48\textwidth}
	    \includegraphics[width=1.0\textwidth]{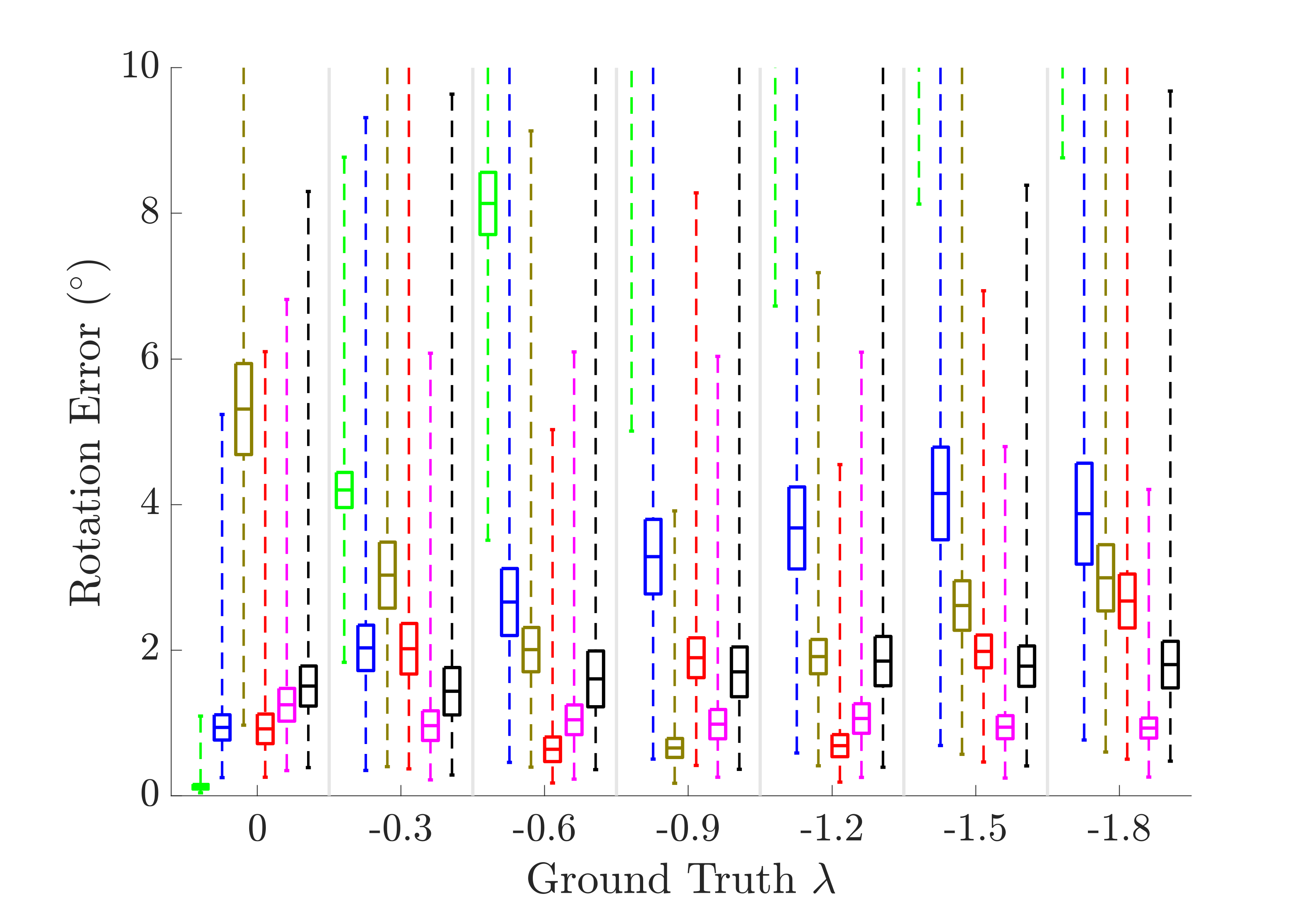}
	\end{subfigure}
    \begin{subfigure}[t]{0.48\textwidth}
	    \includegraphics[width=1.0\textwidth]{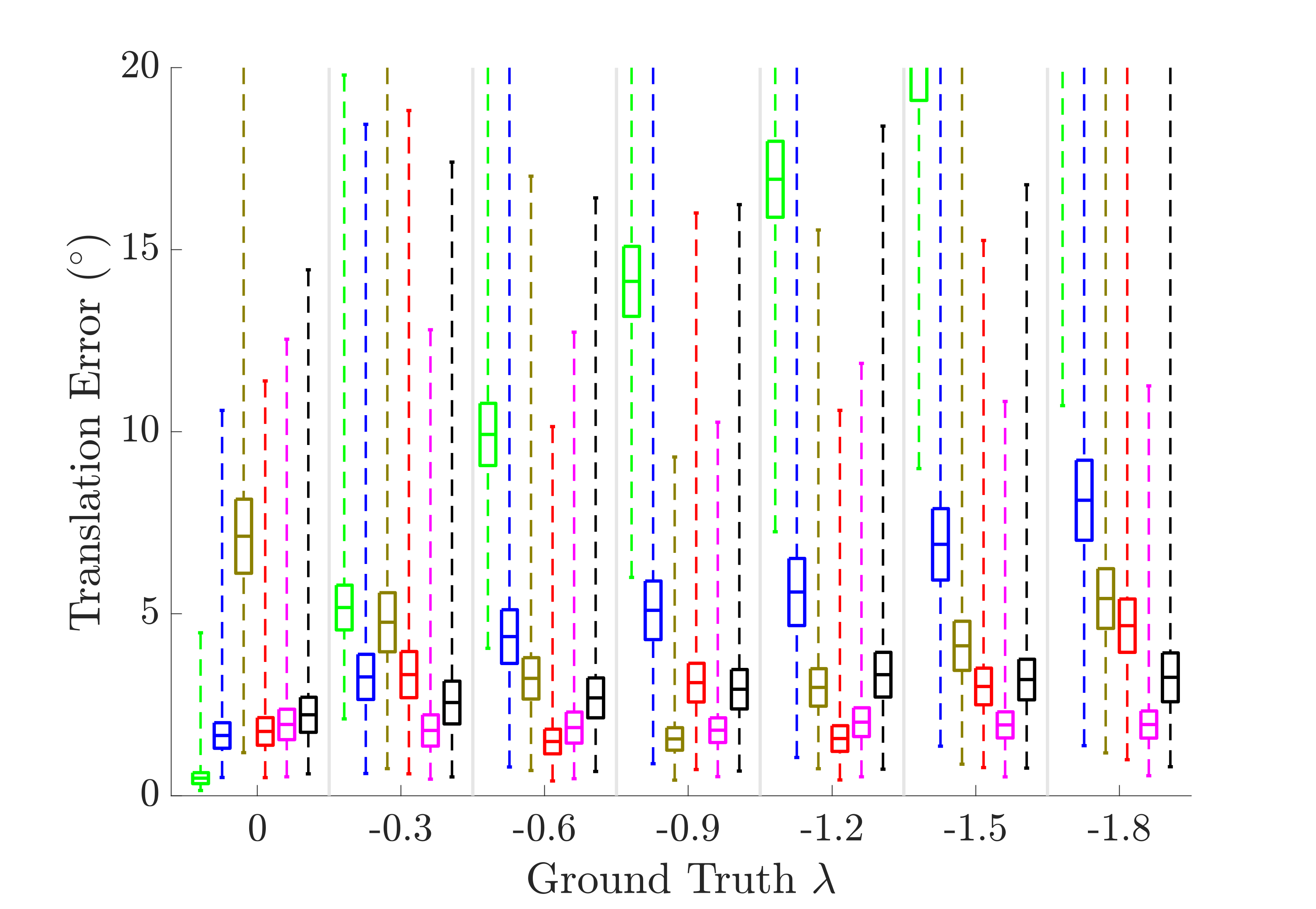}
	\end{subfigure}
     \caption{\textbf{Robustness to varying levels of equal distortion:} We show the estimated radial undistortion parameter, the pose error, and the rotation and translation errors for the \ETH dataset. 
    An x-axis value of $-0.3$ means that we apply a distortion that can be undistorted with the undistortion parameter $-0.3$. The cyan line indicates the ground truth $\lambda$ value on the y-axis.}
    \label{fig:eth_box}
\end{figure*}

\begin{figure*}[t]
    \centering
    \begin{subfigure}[t]{0.48\textwidth}
	    \includegraphics[width=1.0\textwidth]{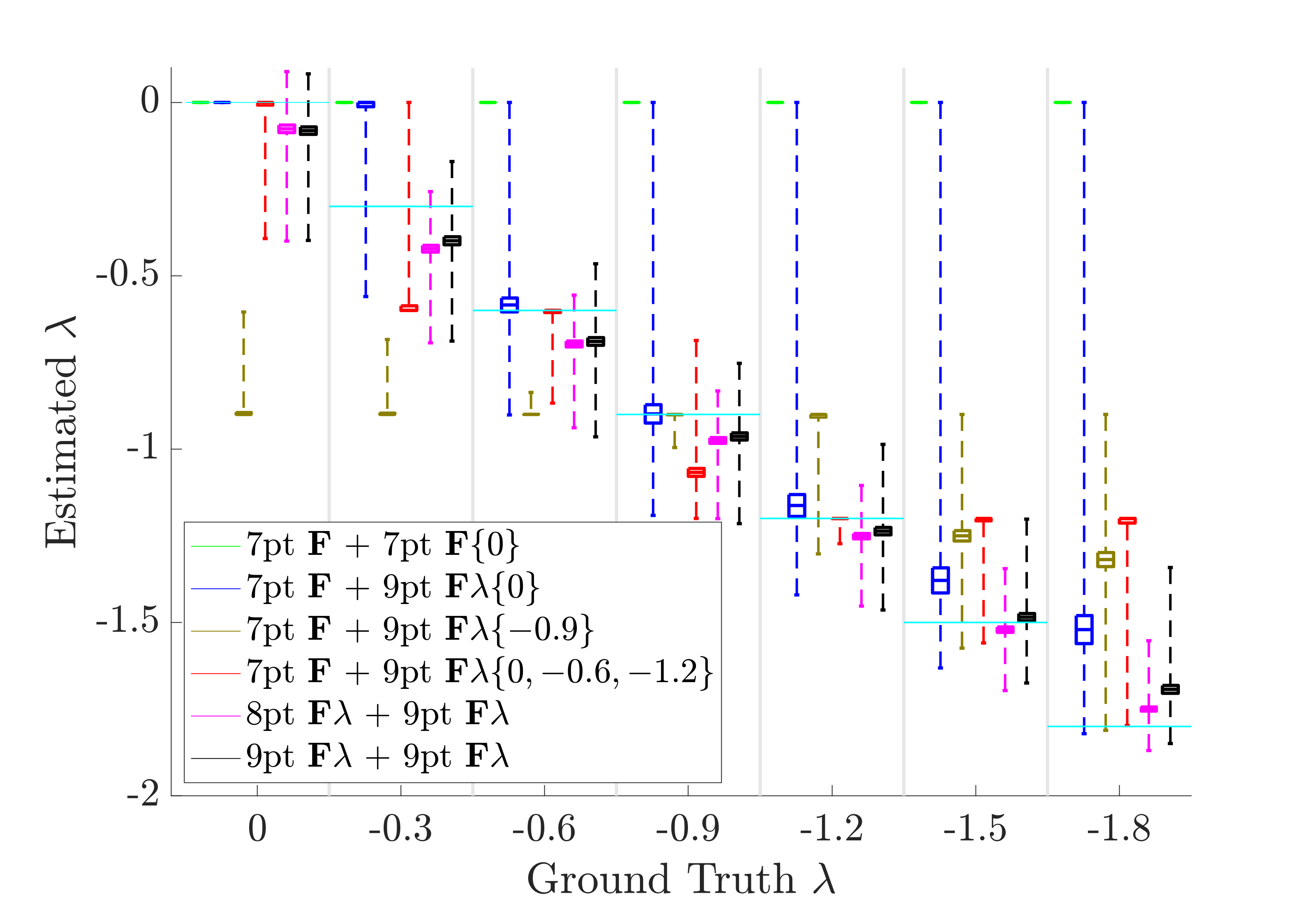}
	\end{subfigure}
    \begin{subfigure}[t]{0.48\textwidth}
	    \includegraphics[width=1.0\textwidth]{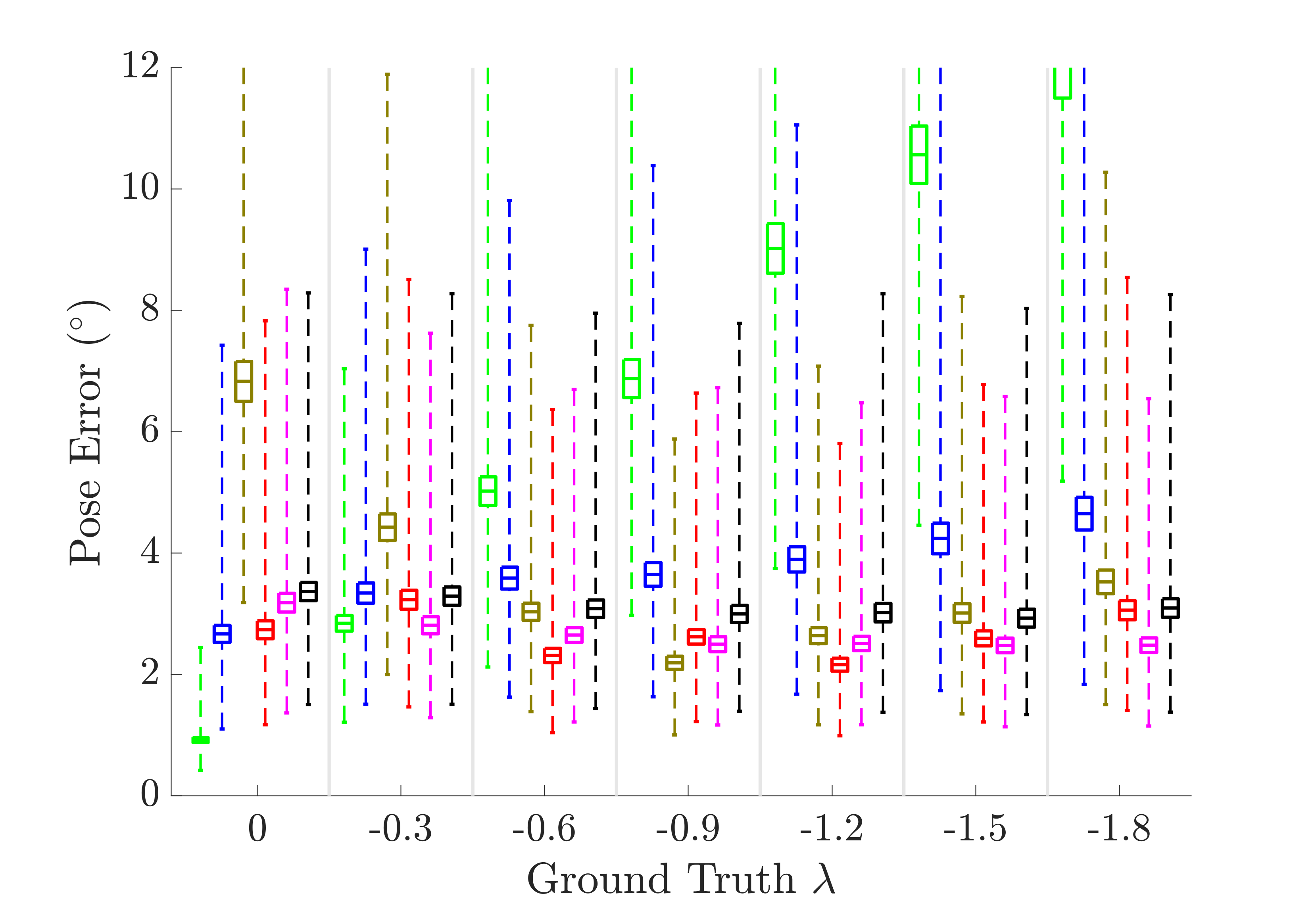}
	\end{subfigure}
	\begin{subfigure}[t]{0.48\textwidth}
	    \includegraphics[width=1.0\textwidth]{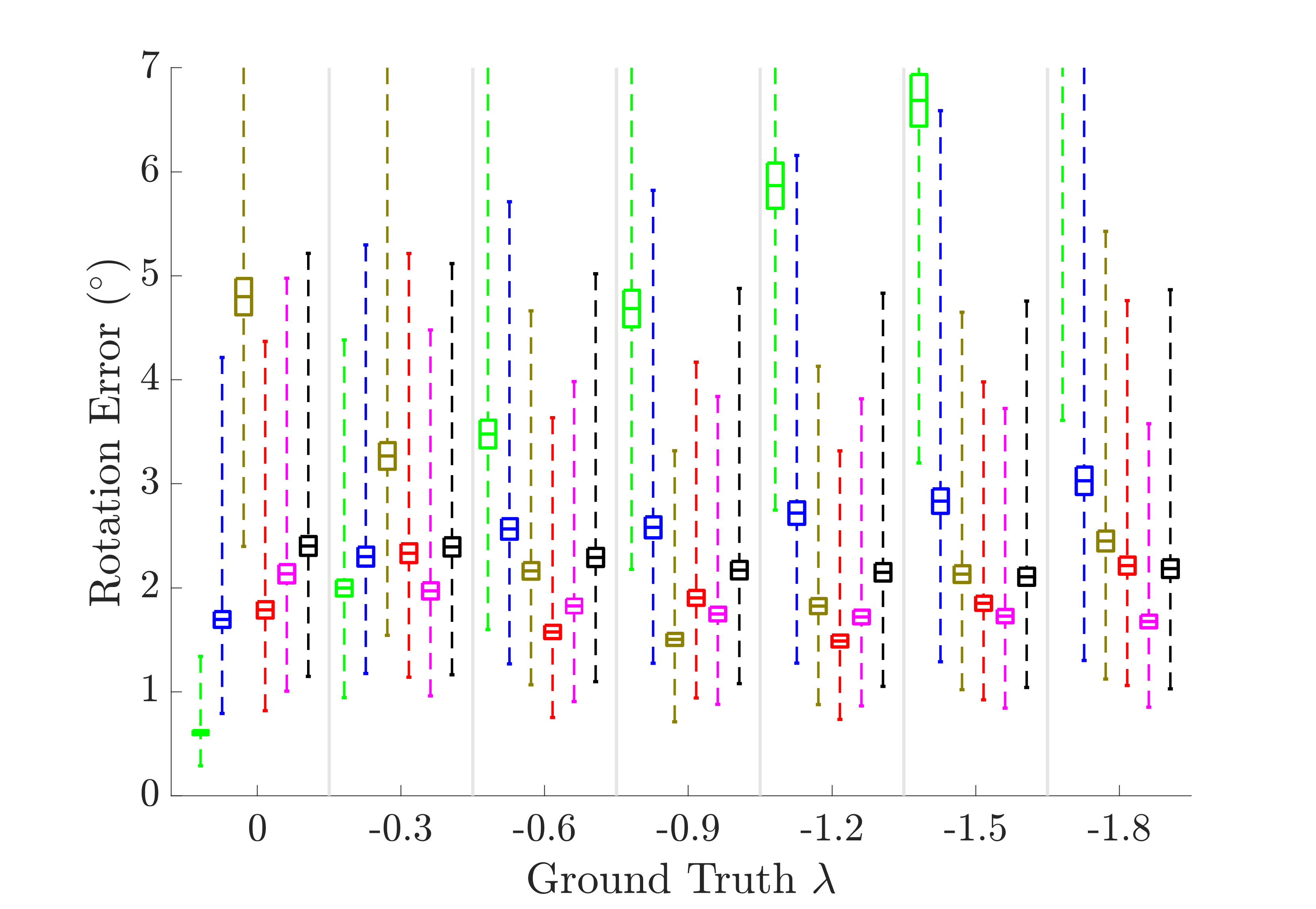}
	\end{subfigure}
	\begin{subfigure}[t]{0.48\textwidth}
	    \includegraphics[width=1.0\textwidth]{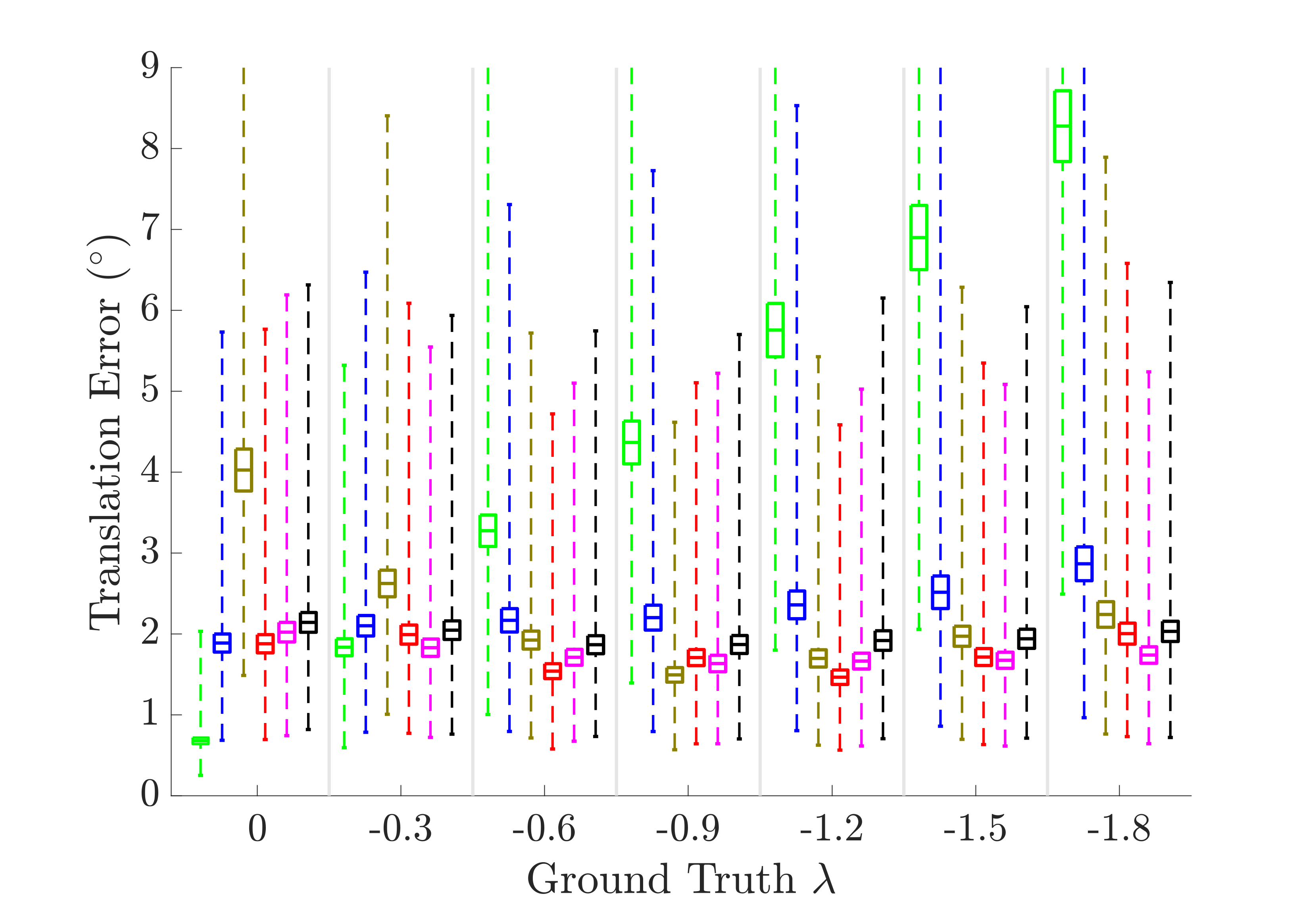}
	\end{subfigure}
    \caption{\textbf{Robustness to varying levels of equal distortion:} We show the estimated radial undistortion parameter, the pose error, and the rotation and translation errors for the Sacre Coeur scene from the \Phototourism dataset. 
    An x-axis value of $-0.3$ means that we apply a distortion that can be undistorted with the undistortion parameter $-0.3$. The cyan line indicates the ground truth $\lambda$ value on the y-axis.}
    \label{fig:sacre_box3}
\end{figure*}

\subsubsection{Robustness to varying levels of distortion.} 
In this section, we provide results similar to the ones of Fig.~1 in the main paper. 
In particular, the solvers discussed in Sec.~3 of the main paper are compared w.r.t. synthetically added distortion. 
Given a pair of images, equal distortion is applied to both of them, and the distortion parameters range from values that correspond to the undistortion parameters $\lambda=0$ to $\lambda=-1.8$. 
We provide results for the \ETH and \Phototourism datasets. 
We evaluate the solvers by measuring the rotation and translation errors, the pose error, which is the maximum of rotation and translation error, and the estimated undistortion parameter $\lambda$, w.r.t. the ground truth. 

Fig.~\ref{fig:eth_box} extends the results shown in Fig.~1 of the main paper 
for 12 scenes from \ETH dataset~\cite{Schops_2017_CVPR}. 
Fig.~1 in the main paper showed the estimated undistortion parameter and the pose error. 
In addition to these two error measures, Fig.~\ref{fig:eth_box} also shows the rotation and translation errors. 

Fig.~\ref{fig:sacre_box3} shows results for the Sacre Coeur scene from the \Phototourism dataset. 
Interestingly, we observe that for the largest part of the range of undistortion parameters, the radial distortion solvers typically predict smaller undistortion values than the "ground truth" (cyan lines in the top-left of Fig.~\ref{fig:sacre_box3}). 
Note that since undistortion values are negative, smaller values mean that in the absolute value these values are larger, and thus correspond to a larger distortion.

Our explanation is that the "undistorted" images provided by the dataset are not fully undistorted, \ie, COLMAP~\cite{Schoenberger2016CVPR} (which was used to estimate ground truth poses and intrinsic parameters) did not fully predict the distortion of each image. 
Rather, a small distortion remains. 
COLMAP uses Approach (1) discussed in the introduction of the main paper, \ie, COLMAP handles radial distortion only during the bundle adjustment stage of the reconstruction process. 
Our results seem to indicate that modeling distortions in earlier stages could be beneficial. 
Due to their simplicity, efficiency, and effectiveness, the sampling-based approaches proposed in our work are promising candidates for this task. 

\begin{table*}[ht]
    \centering
    \caption{Results for $\{($\textit{Scenario A}) \textit{Wild}, (\textit{Scenario B}) \textit{Small distortion}, (\textit{Scenario C}) \textit{Visible distortion}$\}$ on 12 scenes from \ETH dataset, using the Poselib framework and SP+LG matches.}
    \setlength{\tabcolsep}{4.8pt}
    \resizebox{1.0\linewidth}{!}{
\begin{tabular}{ c | r r c | c c c c c | c c | c}
    \toprule
    & & & & \multicolumn{7}{c}{Poselib - \ETH - \textit{Scenario A} \textit{Wild}} \\
        \midrule
    & Minimal & Refinement & Sample & AVG $(^\circ)$ $\downarrow$ & MED $(^\circ)$ $\downarrow$ & AUC@5 $\uparrow$ & @10 & @20 & AVG $\epsilon(\lambda)$ $\downarrow$ & MED $\epsilon(\lambda)$ $\downarrow$ & Time (ms) $\downarrow$ \\
    \midrule
    \multirow{10}{*}{\rotatebox[origin=c]{90}{$\lambda_1 = \lambda_2$}}     & 7pt \F & \F & 0 & 37.97 & 15.97 & 0.15 & 0.24 & 0.36 & 0.87 & 0.87 & \phantom{11}\textbf{32.69} \\
    & 7pt \F & \Fk & $0$ & 23.30 & \phantom{1}1.38 & 0.56 & 0.61 & 0.67 & 0.26 & \underline{0.03} & \phantom{1}125.28 \\
    & 7pt \F & \Fkk & $0$ & 25.40 & \phantom{1}1.79 & 0.54 & 0.59 & 0.65 & 0.35 & 0.07 & \phantom{1}126.50 \\
    & 7pt \F & \Fk & $\{0.0, -0.6, -1.2\}$ & \underline{18.42} & \phantom{1}\underline{0.76} & \underline{0.65} & \underline{0.70} & \underline{0.74} & \textbf{0.14} & \bf{0.02} & \phantom{1}259.43 \\
    & 7pt \F & \Fkk & $\{0.0, -0.6, -1.2\}$ & 19.97 & \phantom{1}0.96 & 0.63 & 0.68 & 0.73 & \underline{0.24} & 0.04 & \phantom{1}733.40 \\
    \cmidrule{2-12}
    & 8pt \Fk & \Fk & \ding{55} & \textbf{18.38} & \phantom{1}\textbf{0.68} & \textbf{0.66} & \textbf{0.71} & \textbf{0.75} & \bf{0.14} & \textbf{0.02} & 1161.38 \\
    & 9pt \Fk & \Fk & \ding{55} & 22.94 & \phantom{1}1.34 & 0.56 & 0.62 & 0.67 & 0.26 & \underline{0.03} & \phantom{1}335.45 \\
    \cmidrule{2-12}
    & 9pt \Fkk & \Fkk & \ding{55} & 28.68 & \phantom{1}3.27 & 0.48 & 0.53 & 0.60 & 0.42 & 0.13 & 2733.10 \\
    & 10pt \Fkk & \Fkk & \ding{55} & 23.71 & \phantom{1}1.81 & 0.54 & 0.59 & 0.66 & 0.28 & 0.06 & \phantom{1}\underline{125.03} \\
    \midrule
    \midrule
    \multirow{5}{*}{\rotatebox[origin=c]{90}{$\lambda_1 \neq \lambda_2$}}     & 7pt \F & \F & 0 & 45.01 & 23.81 & 0.09 & 0.14 & 0.25 & 0.87 & 0.87 & \phantom{11}\textbf{47.94} \\
    & 7pt \F & \Fkk & $0$ & 29.03 & \phantom{1}3.43 & 0.47 & 0.53 & 0.59 & 0.38 & 0.12 & \phantom{1}160.31 \\
    & 7pt \F & \Fkk & $\{0.0, -0.6, -1.2\}$ & \textbf{21.18} & \phantom{1}\textbf{1.01} & \textbf{0.61} & \textbf{0.67} & \textbf{0.72} & \textbf{0.25} & \textbf{0.04} & \phantom{1}765.85 \\
    \cmidrule{2-12}
    & 9pt \Fkk & \Fkk & \ding{55} & 28.29 & \phantom{1}3.52 & 0.47 & 0.52 & 0.59 & 0.41 & 0.13 & 2733.84 \\
    & 10pt \Fkk & \Fkk & \ding{55} & \underline{23.77} & \phantom{1}\underline{1.78} & \underline{0.54} & \underline{0.60} & \underline{0.66} & \underline{0.28} & \underline{0.07} & \phantom{1}\underline{132.51} \\
    \bottomrule
    \toprule
    & & & & \multicolumn{7}{c}{Poselib - \ETH - \textit{Scenario B} \textit{Small distortion}} \\
    \midrule
    & Minimal & Refinement & Sample & AVG $(^\circ)$ $\downarrow$ & MED $(^\circ)$ $\downarrow$ & AUC@5 $\uparrow$ & @10 & @20 & AVG $\epsilon(\lambda)$ $\downarrow$ & MED $\epsilon(\lambda)$ $\downarrow$ & Time (ms) $\downarrow$ \\
    \midrule
    \multirow{8}{*}{\rotatebox[origin=c]{90}{$\lambda_1 = \lambda_2$}}     & 7pt \F & \F & 0 & 20.22 & 3.20 & 0.45 & 0.58 & 0.68 & 0.15 & 0.15 & \phantom{11}\textbf{21.63} \\
    & 7pt \F & \Fk & $0$ & \textbf{17.34} & \underline{0.70} & \textbf{0.66} & \textbf{0.71} & \textbf{0.76} & \textbf{0.07} & \bf{0.01} & \phantom{1}\underline{111.57} \\
    & 7pt \F & \Fkk & $0$ & 19.20 & 0.87 & \underline{0.64} & \underline{0.70} & 0.74 & 0.17 & 0.03 & \phantom{1}112.44 \\
    \cmidrule{2-12}
    & 8pt \Fk & \Fk & \ding{55} & \underline{18.04} & \textbf{0.67} & \bf{0.66} & \underline{0.70} & \underline{0.75} & \underline{0.14} & \textbf{0.01} & 1242.30 \\
    & 9pt \Fk & \Fk & \ding{55} & 21.02 & 1.06 & 0.60 & 0.66 & 0.71 & \underline{0.14} & \underline{0.02} & \phantom{1}356.98 \\
    
    \cmidrule{2-12}
    & 9pt \Fkk & \Fkk & \ding{55} & 23.15 & 1.48 & 0.56 & 0.62 & 0.67 & 0.22 & 0.05 & 2575.97 \\
    & 10pt \Fkk & \Fkk & \ding{55} & 24.91 & 1.81 & 0.54 & 0.59 & 0.65 & 0.28 & 0.06 & \phantom{1}135.44 \\
    \midrule
    \midrule
    \multirow{4}{*}{\rotatebox[origin=c]{90}{$\lambda_1 \neq \lambda_2$}}     & 7pt \F & \F & 0 & 23.31 & 4.66 & 0.34 & 0.48 & 0.60 & \textbf{0.15} & 0.15 & \phantom{11}\textbf{25.48} \\
    & 7pt \F & \Fkk & $0$ & \textbf{19.86} & \textbf{1.00} & \textbf{0.62} & \textbf{0.68} & \textbf{0.73} & \underline{0.18} & \textbf{0.03} & \phantom{1}\underline{117.94} \\
    \cmidrule{2-12}
    & 9pt \Fkk & \Fkk & \ding{55} & \underline{22.43} & \underline{1.60} & \underline{0.55} & \underline{0.61} & \underline{0.67} & 0.20 & \underline{0.05} & 2538.66 \\
    & 10pt \Fkk & \Fkk & \ding{55} & 24.67 & 1.82 & 0.54 & 0.59 & 0.65 & 0.28 & 0.06 & \phantom{1}138.26 \\
    \bottomrule
        \toprule
    & & & & \multicolumn{7}{c}{Poselib - \ETH - \textit{Scenario C} \textit{Visible distortion}} \\
    \midrule
    & Minimal & Refinement & Sample & AVG $(^\circ)$ $\downarrow$ & MED $(^\circ)$ $\downarrow$ & AUC@5 $\uparrow$ & @10 & @20 & AVG $\epsilon(\lambda)$ $\downarrow$ & MED $\epsilon(\lambda)$ $\downarrow$ & Time (ms) $\downarrow$ \\
    \midrule
    \multirow{10}{*}{\rotatebox[origin=c]{90}{$\lambda_1 = \lambda_2$}}     & 7pt \F & \F & 0 & 44.12 & 20.78 & 0.07 & 0.13 & 0.26 & 1.10 & 1.09 & \phantom{11}\textbf{36.27} \\
    & 7pt \F & \Fk & $-0.9$ & 18.63 & \phantom{1}0.84 & 0.63 & 0.69 & 0.73 & \underline{0.12} & 0.02 & \phantom{1}\underline{102.13} \\
    & 7pt \F & \Fkk & $-0.9$ & 20.80 & \phantom{1}1.17 & 0.60 & 0.66 & 0.71 & 0.24 & 0.05 & \phantom{1}107.32 \\
    & 7pt \F & \Fk & $\{-0.6, -0.9, -1.2\}$ & \textbf{17.74} & \phantom{1}\textbf{0.67} & \textbf{0.67} & \textbf{0.72} & \textbf{0.76} & \textbf{0.11} & \bf{0.02} & \phantom{1}249.32 \\
    & 7pt \F & \Fkk & $\{-0.6, -0.9, -1.2\}$ & 19.74 & \phantom{1}0.92 & 0.63 & 0.69 & 0.74 & 0.22 & 0.04 & \phantom{1}726.06 \\
    \cmidrule{2-12}
    & 8pt \Fk & \Fk & \ding{55} & \underline{18.08} & \phantom{1}\underline{0.70} & \underline{0.66} & \underline{0.71} & \underline{0.75} & 0.14 & \textbf{0.02} & 1137.91 \\
    & 9pt \Fk & \Fk & \ding{55} & 23.47 & \phantom{1}1.51 & 0.55 & 0.61 & 0.66 & 0.27 & \underline{0.04} & \phantom{1}330.17 \\
    \cmidrule{2-12}
    & 9pt \Fkk & \Fkk & \ding{55} & 31.12 & \phantom{1}5.26 & 0.44 & 0.49 & 0.55 & 0.48 & 0.19 & 2777.24 \\
    & 10pt \Fkk & \Fkk & \ding{55} & 24.36 & \phantom{1}1.84 & 0.53 & 0.59 & 0.66 & 0.30 & 0.07 & \phantom{1}124.10 \\
    \midrule
    \midrule
    \multirow{5}{*}{\rotatebox[origin=c]{90}{$\lambda_1 \neq \lambda_2$}}     & 7pt \F & \F & 0 & 48.82 & 26.36 & 0.05 & 0.10 & 0.21 & 1.10 & 1.10 & \phantom{11}\textbf{44.53} \\
    & 7pt \F & \Fkk & $-0.9$ & \underline{23.66} & \phantom{1}1.66 & \underline{0.55} & \underline{0.61} & \underline{0.67} & \underline{0.26} & \underline{0.06} & \phantom{1}\underline{124.89} \\
    & 7pt \F & \Fkk & $\{-0.6, -0.9, -1.2\}$ & \textbf{19.46} & \phantom{1}\textbf{0.94} & \textbf{0.63} & \textbf{0.68} & \textbf{0.73} & \textbf{0.22} & \textbf{0.04} & \phantom{1}741.01 \\
    \cmidrule{2-12}
    & 9pt \Fkk & \Fkk & \ding{55} & 29.20 & \phantom{1}4.54 & 0.45 & 0.50 & 0.57 & 0.47 & 0.17 & 2813.36 \\
    & 10pt \Fkk & \Fkk & \ding{55} & 23.85 & \phantom{1}\underline{1.62} & 0.54 & 0.60 & 0.66 & 0.27 & \underline{0.06} & \phantom{1}129.30 \\
    \bottomrule
    \end{tabular}
    }
    \label{tab:poselib_eth3d_scenaria}
\end{table*}

\begin{table*}[ht]
    \centering
    \caption{Results for (\textit{Scenario A}) \textit{Wild} on 5k image pairs on each of Sacre Coeur, St. Peter's Square and Temple Nara Japan scenes from the \Phototourism dataset, using the GC-RANSAC framework with SP+LG matches.}
    \setlength{\tabcolsep}{4.8pt}
    \resizebox{1.0\linewidth}{!}{
    \begin{tabular}{ c | r r c | c c c c c | c c | c}
    \toprule
        & & & & \multicolumn{8}{c}{GC-RANSAC - \Phototourism - Sacre Coeur} \\
        \midrule
        & Minimal & Non-Minimal & Sample & AVG $(^\circ)$ $\downarrow$ & MED $(^\circ)$ $\downarrow$ & AUC@5 $\uparrow$ & @10 & @20 & AVG $\epsilon(\lambda)$ $\downarrow$ & MED $\epsilon(\lambda)$ $\downarrow$ & Time (ms) $\downarrow$ \\
        \midrule
        \multirow{11}{*}{\rotatebox[origin=c]{90}{$\lambda_1 = \lambda_2$}} & 7pt \F & 7pt \F & 0 & 13.11 & 5.72 & 0.21 & 0.38 & 0.55 & 0.86 & 0.85 & \phantom{1}90.04 \\
        & 7pt \F & 8pt \F & 0 & 12.83 & 6.15 & 0.20 & 0.36 & 0.55 & 0.86 & 0.85 & \phantom{1}\bf{50.12}\\
        & 7pt \F & 9pt \Fk & 0 & 10.49 & 3.56 & 0.28 & 0.47 & 0.63 & 0.53 & 0.36 & \phantom{1}\underline{51.43} \\
        & 7pt \F & 12pt \Fkk & 0 & 13.71 & 6.44 & 0.17 & 0.34 & 0.52 & 0.85 & 0.83 & \phantom{1}70.35 \\
        & 7pt \F & 9pt \Fk & $\{-1.2, -0.6, 0\}$ & \phantom{1}\underline{7.88} & \underline{2.62} & \textbf{0.35} & \bf{0.55} & \textbf{0.70} & 0.29 & \underline{0.21} & \phantom{1}62.47 \\
        & 7pt \F & 12pt \Fkk & $\{-1.2, -0.6, 0\}$ & \phantom{1}8.47 & 3.16 & 0.31 & 0.51 & 0.68 & \bf{0.27} & 0.22 & \phantom{1}60.94 \\
        \cmidrule{2-12}
        & 8pt \Fk & 9pt \Fk & \ding{55} & \phantom{1}\bf{7.71} & \bf{2.57} & \bf{0.35} & \textbf{0.55} & \bf{0.70} & 0.29 & \bf{0.18} & 150.03 \\
        & 8pt \Fk & 12pt \Fkk & \ding{55} & \phantom{1}7.97 & 3.07 & 0.31 & \underline{0.52} & \underline{0.68} & \underline{0.28} & \bf{0.18} & 160.09 \\
        \cmidrule{2-12}
        & 9pt \Fk & 9pt \Fk & \ding{55} & \phantom{1}8.78 & 3.05 & \underline{0.32} & 0.51 & 0.67 & 0.36 & 0.24 & \phantom{1}70.22\\
        & 9pt \Fk & 12pt \Fkk & \ding{55} & \phantom{1}9.50 & 3.78 & 0.26 & 0.46 & 0.64 & 0.38 & 0.27 & \phantom{1}80.83\\
        \midrule
        \midrule
        \multirow{6}{*}{\rotatebox[origin=c]{90}{$\lambda_1 \neq \lambda_2$}} & 7pt \F & 7pt \F & 0 & 15.14 & 7.04 & 0.15 & 0.32 & 0.50 & 0.86 & 0.86 & \phantom{1}40.78\\
        & 7pt \F & 8pt \F & 0 & 14.88 & 7.44 & 0.14 & 0.31 & 0.50 & 0.86 & 0.86 & \phantom{1}\bf{20.69} \\
        & 7pt \F & 12pt \Fkk & $\{-1.2, -0.6, 0\}$ & \phantom{1}\bf{9.21} & \bf{3.43} & \bf{0.29} & \bf{0.49} & \bf{0.66} & \bf{0.34} & \bf{0.29} & \phantom{1}90.91\\
        \cmidrule{2-12}
        & 9pt \Fkk & 12pt \Fkk & \ding{55} & \underline{13.07} & \underline{5.67} & \underline{0.19} & \underline{0.37} & \underline{0.55} & \underline{0.46} & \underline{0.39} & 130.03 \\
        & 10pt \Fkk & 12pt \Fkk & \ding{55} & 13.45 & 5.99 & 0.18 & 0.36 & 0.54 & 0.49 & 0.42 & \phantom{1}\underline{20.75} \\
        \bottomrule
        \toprule
        & & & & \multicolumn{8}{c}{GC-RANSAC - \Phototourism - St. Peter's Square} \\
        \midrule
        & Minimal & Non-Minimal & Sample & AVG $(^\circ)$ $\downarrow$ & MED $(^\circ)$ $\downarrow$ & AUC@5 $\uparrow$ & @10 & @20 & AVG $\epsilon(\lambda)$ $\downarrow$ & MED $\epsilon(\lambda)$ $\downarrow$ & Time (ms) $\downarrow$ \\
        \midrule
        \multirow{11}{*}{\rotatebox[origin=c]{90}{$\lambda_1 = \lambda_2$}} & 7pt \F & 7pt \F & 0 & 30.22 & 22.80 & 0.05 & 0.11 & 0.23 & 0.85 & 0.84 & \phantom{1}70.37\\
        & 7pt \F & 8pt \F & 0 & 31.13 & 24.42 & 0.03 & 0.09 & 0.20 & 0.85 & 0.84 & \phantom{1}\underline{40.64}\\
        & 7pt \F & 9pt \Fk & 0 & 30.62 & 24.09 & 0.04 & 0.10 & 0.21 & 0.65 & 0.51 & \phantom{1}41.03 \\
        & 7pt \F & 12pt \Fkk & 0 & 31.91 & 25.23 & 0.03 & 0.08 & 0.19 & 0.82 & 0.80 & \phantom{1}\bf{40.12}\\
        & 7pt \F & 9pt \Fk & $\{-1.2, -0.6, 0\}$ & 25.78 & 17.14 & \bf 0.07 & \underline{0.16} & 0.30 & 0.37 & 0.23 & \phantom{1}50.11\\
        & 7pt \F & 12pt \Fkk & $\{-1.2, -0.6, 0\}$ & 25.90 & 16.89 & \underline{0.06} & 0.15 & 0.30 & \underline{0.27} & 0.22 & \phantom{1}52.28 \\
        \cmidrule{2-12}
        & 8pt \Fk & 9pt \Fk & \ding{55} & \bf{24.62} & \bf{15.71} & \bf{0.07} & \bf{0.17} & \bf{0.32} & 0.33 & \underline{0.17} & \phantom{1}70.62 \\
        & 8pt \Fk & 12pt \Fkk & \ding{55} & \underline{24.89} & \underline{15.73} & \bf{0.07} & \bf{0.17} & \underline{0.31} & \bf{0.22} & \bf{0.13} & \phantom{1}81.35 \\
        \cmidrule{2-12}
        & 9pt \Fk & 9pt \Fk & \ding{55} & 26.06 & 18.21 & \underline{0.06} & 0.14 & 0.28 & 0.46 & 0.29 & \phantom{1}53.29\\
        & 9pt \Fk & 12pt \Fkk & \ding{55} & 26.55 & 18.20 & 0.05 & 0.13 & 0.27 & 0.41 & 0.28 & \phantom{1}54.94\\
        \midrule
        \midrule
        \multirow{6}{*}{\rotatebox[origin=c]{90}{$\lambda_1 \neq \lambda_2$}} & 7pt \F & 7pt \F & 0 & 31.36 & 25.39 & 0.03 & 0.09 & 0.20 & 0.86 & 0.86 & \phantom{1}41.03 \\
        & 7pt \F & 8pt \F & 0 & 32.03 & 25.70 & 0.03 & 0.08 & 0.18 & 0.86 & 0.86 & \phantom{1}\bf{20.70}\\
        & 7pt \F & 12pt \Fkk & $\{-1.2, -0.6, 0\}$ & \bf{23.00} & \bf{14.39} & \bf{0.07} & \bf{0.18} & \bf{0.33} & \bf{0.35} & \bf{0.30} & \phantom{1}91.87 \\
        \cmidrule{2-12}
        & 9pt \Fkk & 12pt \Fkk & \ding{55} & \underline{28.13} & \underline{20.03} & \underline{0.04} & \underline{0.12} & \underline{0.26} & \underline{0.40} & \underline{0.33} & 100.80\\
        & 10pt \Fkk & 12pt \Fkk & \ding{55} & 30.28 & 22.22 & 0.03 & 0.10 & 0.23 & 0.44 & 0.37 & \phantom{1}\underline{21.31}\\
        \bottomrule
        \toprule
        & & & & \multicolumn{8}{c}{GC-RANSAC - \Phototourism - Temple Nara Japan} \\
        \midrule
        & Minimal & Non-Minimal & Sample & AVG $(^\circ)$ $\downarrow$ & MED $(^\circ)$ $\downarrow$ & AUC@5 $\uparrow$ & @10 & @20 & AVG $\epsilon(\lambda)$ $\downarrow$ & MED $\epsilon(\lambda)$ $\downarrow$ & Time (ms) $\downarrow$ \\
        \midrule
        \multirow{11}{*}{\rotatebox[origin=c]{90}{$\lambda_1 = \lambda_2$}} & 7pt \F & 7pt \F & 0 & 30.74 & 22.21 & 0.06 & 0.15 & 0.26 & 0.87 & 0.87 & \phantom{1}80.49 \\
        & 7pt \F & 8pt \F & 0 & 29.79 & 21.29 & 0.06 & 0.13 & 0.26 & 0.87 & 0.87 & \phantom{1}\bf 50.12\\
        & 7pt \F & 9pt \Fk & 0 & 28.05 & 18.43 & 0.07 & 0.17 & 0.30 & 0.51 & 0.31 & \phantom{1}\underline{50.24}\\
        & 7pt \F & 12pt \Fkk & 0 & 32.80 & 25.98 & 0.03 & 0.10 & 0.21 & 0.83 & 0.81 & \phantom{1}51.32 \\
        & 7pt \F & 9pt \Fk & $\{-1.2, -0.6, 0\}$ & \underline{20.55} & \underline{10.42} & \bf{0.11} & \underline{0.24} & \bf{0.41} & 0.29 & 0.18 & \phantom{1}60.93\\
        & 7pt \F & 12pt \Fkk & $\{-1.2, -0.6, 0\}$ & 22.68 & 13.93 & 0.07 & 0.19 & 0.34 & 0.27 & 0.21 & \phantom{1}60.02 \\
        \cmidrule{2-12}
        & 8pt \Fk & 9pt \Fk & \ding{55} & \bf 19.85 & \bf 10.32 & \bf 0.11 & \bf 0.25 & \bf 0.41 & \underline{0.26} & \underline{0.14} & \phantom{1}81.44 \\
        & 8pt \Fk & 12pt \Fkk & \ding{55} & 21.04 & 12.75 & 0.08 & 0.20 & 0.36 & \bf 0.22 & \bf 0.13 & \phantom{1}80.80\\
        \cmidrule{2-12}
        & 9pt \Fk & 9pt \Fk & \ding{55} &  21.23 & 11.76 & \underline{0.10} & 0.23 & \underline{0.39} & 0.37 & 0.21 & \phantom{1}52.91\\
        & 9pt \Fk & 12pt \Fkk & \ding{55} &  23.48 & 15.11 & 0.06 & 0.16 & 0.32 & 0.46 & 0.32 & \phantom{1}52.30\\
        \midrule
        \midrule
        \multirow{6}{*}{\rotatebox[origin=c]{90}{$\lambda_1 \neq \lambda_2$}} & 7pt \F & 7pt \F & 0 & 31.60 & 23.63 & 0.04 & 0.11 & 0.23 & 0.86 & 0.87 & \phantom{1}60.39 \\
        & 7pt \F & 8pt \F & 0 & 30.41 & 21.19 & 0.04 & 0.10 & 0.23 & 0.86 & 0.87 & \phantom{1}\underline{30.44} \\
        & 7pt \F & 12pt \Fkk & $\{-1.2, -0.6, 0\}$ & \bf 21.11 & \bf 12.42 & \bf 0.08 & \bf 0.20 & \bf 0.37 & \bf 0.35 & \bf 0.29 & 120.72 \\  
        \cmidrule{2-12}
        & 9pt \Fkk & 12pt \Fkk & \ding{55} & \underline{26.13} & \underline{17.40} & \underline{0.05} & \underline{0.14} & \underline{0.29} & \underline{0.41} & \underline{0.32} & 101.89\\
        & 10pt \Fkk & 12pt \Fkk & \ding{55} & 27.75 & 19.15 & 0.04 & 0.13 & 0.26 & 0.45 & 0.38 & \phantom{1}\bf 20.88\\
        \bottomrule
    \end{tabular}
    }
    \label{tab:phototourism_st_peter_square_main_splg}
\end{table*}

\begin{table*}[ht]
    \centering
    \caption{Results for (\textit{Scenario A}) \textit{Wild} on 5k image pairs for each of the Sacre Coeur, St. Peter's Square and Temple Nara Japan scenes from the \Phototourism dataset, using the GC-RANSAC framework with SIFT+LG matches.}
    \setlength{\tabcolsep}{4.8pt}
    \resizebox{1.0\linewidth}{!}{
    \begin{tabular}{ c | r r c | c c c c c | c c | c}
        \toprule
        & & & & \multicolumn{8}{c}{GC-RANSAC - \Phototourism - Sacre Coeur} \\
        \midrule
        & Minimal & Non-Minimal & Sample & AVG $(^\circ)$ $\downarrow$ & MED $(^\circ)$ $\downarrow$ & AUC@5 $\uparrow$ & @10 & @20 & AVG $\epsilon(\lambda)$ $\downarrow$ & MED $\epsilon(\lambda)$ $\downarrow$ & Time (ms) $\downarrow$ \\
        \midrule
        \multirow{11}{*}{\rotatebox[origin=c]{90}{$\lambda_1 = \lambda_2$}} & 7pt \F & 7pt \F & 0 & 17.24 & 6.45 & 0.21 & 0.36 & 0.51 & 0.86 & 0.85 & \phantom{1}60.49\\
        & 7pt \F & 8pt \F & 0 & 16.91 & 6.93 & 0.19 & 0.34 & 0.51 & 0.86 & 0.85 & \phantom{1}\textbf{40.22}\\
        & 7pt \F & 9pt \Fk & 0 & 14.73 & 4.75 & 0.25 & 0.42 & 0.57 & 0.55 & 0.39 & \phantom{1}\underline{40.79}\\
        & 7pt \F & 12pt \Fkk & 0 & 18.01 & 7.72 & 0.15 & 0.31 & 0.47 & 0.84 & 0.83 & \phantom{1}50.90\\
        & 7pt \F & 9pt \Fk & $\{-1.2, -0.6, 0\}$ & \underline{12.55} & \textbf{3.58} & \textbf{0.30} & \textbf{0.47} & \textbf{0.62} & \underline{0.34} & \underline{0.24} & \phantom{1}60.66\\
        & 7pt \F & 12pt \Fkk & $\{-1.2, -0.6, 0\}$ & 13.28 & 4.43 & 0.25 & 0.43 & \underline{0.59} & \textbf{0.33} & 0.26 & \phantom{1}60.73\\
        \cmidrule{2-12}
        & 8pt \Fk & 9pt \Fk & \ding{55} & \textbf{12.28} & \textbf{3.58} & \textbf{0.30} & \textbf{0.47} & \textbf{0.62} & 0.36 & \textbf{0.23} & 210.34\\
        & 8pt \Fk & 12pt \Fkk & \ding{55} & 13.16 & 4.49 & 0.25 & 0.42 & \underline{0.59} & 0.36 & \underline{0.24} & 230.50 \\
        \cmidrule{2-12}
        & 9pt \Fk & 9pt \Fk & \ding{55} & 13.96 & \underline{4.27} & \underline{0.26} & \underline{0.44} & \underline{0.59} & 0.43 & 0.30 & \phantom{1}90.55\\
        & 9pt \Fk & 12pt \Fkk & \ding{55} & 14.94 & 5.56 & 0.20 & 0.38 & 0.55 & 0.47 & 0.35 & 102.08\\
        \midrule
        \midrule
        \multirow{6}{*}{\rotatebox[origin=c]{90}{$\lambda_1 \neq \lambda_2$}} & 7pt \F & 7pt \F & 0 & 18.78 & 8.04 & 0.16 & 0.31 & 0.47 & 0.86 & 0.86 & \phantom{1}61.19\\
        & 7pt \F & 8pt \F & 0 & 18.31 & 8.18 & 0.15 & 0.30 & 0.47 & 0.86 & 0.86 & \phantom{1}\textbf{40.20} \\
        & 7pt \F & 12pt \Fkk & $\{-1.2, -0.6, 0\}$ & \textbf{13.42} & \textbf{4.63} & \textbf{0.24} & \textbf{0.42} & \textbf{0.58} & \textbf{0.38} & \textbf{0.30} & 120.01 \\
        \cmidrule{2-12}
        & 9pt \Fkk & 12pt \Fkk & \ding{55} & 14.89 & 5.45 & 0.21 & 0.38 & 0.54 & 0.45 & 0.34 & 870.90\\
        & 10pt \Fkk & 12pt \Fkk & \ding{55} & \underline{14.87} & \underline{5.39} & 0.21 & 0.38 & 0.55 & 0.47 & 0.36 & \phantom{1}\underline{60.47}\\
        \bottomrule
        \toprule
        & & & & \multicolumn{8}{c}{GC-RANSAC - \Phototourism - St. Peter's Square} \\
        \midrule
        & Minimal & Non-Minimal & Sample & AVG $(^\circ)$ $\downarrow$ & MED $(^\circ)$ $\downarrow$ & AUC@5 $\uparrow$ & @10 & @20 & AVG $\epsilon(\lambda)$ $\downarrow$ & MED $\epsilon(\lambda)$ $\downarrow$ & Time (ms) $\downarrow$ \\
        \midrule
        \multirow{11}{*}{\rotatebox[origin=c]{90}{$\lambda_1 = \lambda_2$}} & 7pt \F & 7pt \F & 0 & 29.09 & 21.46 & 0.05 & 0.12 & 0.24 & 0.86 & 0.84 & \phantom{1}50.03\\
        & 7pt \F & 8pt \F & 0 & 28.51 & 21.20 & 0.04 & 0.11 & 0.24 & 0.86 & 0.84 & \phantom{1}\textbf{30.42}\\
        & 7pt \F & 9pt \Fk & 0 & 28.51 & 21.03 & 0.05 & 0.13 & 0.25 & 0.65 & 0.51 & \phantom{1}42.27\\
        & 7pt \F & 12pt \Fkk & 0 & 29.91 & 22.77 & 0.03 & 0.10 & 0.22 & 0.84 & 0.81 & \phantom{1}\underline{40.19}\\
        & 7pt \F & 9pt \Fk & $\{-1.2, -0.6, 0\}$ & \underline{25.10} & \underline{16.56} & \textbf{0.08} & \textbf{0.17} & \textbf{0.31} & 0.41 & 0.28 & \phantom{1}50.37\\
        & 7pt \F & 12pt \Fkk & $\{-1.2, -0.6, 0\}$ & 25.74 & 17.17 & \underline{0.07} & \underline{0.16} & \underline{0.30} & \textbf{0.32} & \underline{0.25} & \phantom{1}50.80\\
        \cmidrule{2-12}
        & 8pt \Fk & 9pt \Fk & \ding{55} & \textbf{24.98} & \textbf{16.15} & \textbf{0.08} & \textbf{0.17} & \textbf{0.31} & \underline{0.40} & \underline{0.25} & 190.91\\
        & 8pt \Fk & 12pt \Fkk & \ding{55} & 25.43 & 16.69 & \underline{0.07} & \underline{0.16} & \textbf{0.31} & \textbf{0.32} & \textbf{0.21} & 200.38\\
        \cmidrule{2-12}
        & 9pt \Fk & 9pt \Fk & \ding{55} & 26.29 & 17.44 & 0.06 & 0.15 & 0.29 & 0.49 & 0.34 & \phantom{1}90.00\\
        & 9pt \Fk & 12pt \Fkk & \ding{55} & 26.18 & 17.94 & 0.06 & 0.14 & 0.28 & 0.47 & 0.34 & \phantom{1}92.41\\
        \midrule
        \midrule
        \multirow{6}{*}{\rotatebox[origin=c]{90}{$\lambda_1 \neq \lambda_2$}} & 7pt \F & 7pt \F & 0 & 31.74 & 24.68 & 0.03 & 0.09 & 0.20 & 0.86 & 0.85 & \phantom{1}\underline{50.01}\\
        & 7pt \F & 8pt \F & 0 & 31.00 & 24.28 & 0.02 & 0.08 & 0.20 & 0.86 & 0.85 & \phantom{1}\textbf{30.08}\\
        & 7pt \F & 12pt \Fkk & $\{-1.2, -0.6, 0\}$ & \textbf{25.51} & \textbf{16.86} & \textbf{0.06} & \textbf{0.15} & \textbf{0.30} & \textbf{0.38} & \textbf{0.30} & 100.17\\
        \cmidrule{2-12}
        & 9pt \Fkk & 12pt \Fkk & \ding{55} & 26.76 & 18.29 & \underline{0.05} & \underline{0.14} & \underline{0.28} & \underline{0.43} & \underline{0.31} & 730.38\\
        & 10pt \Fkk & 12pt \Fkk & \ding{55} & \underline{26.66} & \underline{18.14} & \underline{0.05} & \underline{0.14} & \underline{0.28} & 0.47 & 0.35 & \phantom{1}60.40\\
        \bottomrule
        \toprule
        & & & & \multicolumn{8}{c}{GC-RANSAC - \Phototourism - Temple Nara Japan} \\
        \midrule
        & Minimal & Non-Minimal & Sample & AVG $(^\circ)$ $\downarrow$ & MED $(^\circ)$ $\downarrow$ & AUC@5 $\uparrow$ & @10 & @20 & AVG $\epsilon(\lambda)$ $\downarrow$ & MED $\epsilon(\lambda)$ $\downarrow$ & Time (ms) $\downarrow$ \\
        \midrule
        \multirow{11}{*}{\rotatebox[origin=c]{90}{$\lambda_1 = \lambda_2$}} & 7pt \F & 7pt \F & 0 & 36.74 & 31.81 & 0.05 & 0.10 & 0.19 & 0.86 & 0.85 & \phantom{1}40.01 \\
        & 7pt \F & 8pt \F & 0 & 33.81 & 26.93 & 0.04 & 0.11 & 0.21 & 0.86 & 0.85 & \phantom{1}\textbf{20.27}\\
        & 7pt \F & 9pt \Fk & 0 & 30.92 & 22.94 & 0.07 & 0.16 & 0.27 & 0.49 & 0.31 & \phantom{1}\underline{30.13} \\
        & 7pt \F & 12pt \Fkk & 0 & 37.85 & 32.70 & 0.02 & 0.07 & 0.16 & 0.81 & 0.78 & \phantom{1}30.92\\
        & 7pt \F & 9pt \Fk & $\{-1.2, -0.6, 0\}$ & \underline{25.59} & \underline{15.06} & \textbf{0.10} & \textbf{0.20} & \textbf{0.34} & 0.32 & 0.20 & \phantom{1}42.42\\
        & 7pt \F & 12pt \Fkk & $\{-1.2, -0.6, 0\}$ & 29.09 & 20.91 & 0.05 & 0.13 & 0.26 & \underline{0.27} & 0.22 & \phantom{1}40.98\\
        \cmidrule{2-12}
        & 8pt \Fk & 9pt \Fk & \ding{55} & \textbf{24.82} & \textbf{15.01} & \underline{0.09} & \textbf{0.20} & \textbf{0.34} & 0.28 & \underline{0.15} & \phantom{1}70.90 \\
        & 8pt \Fk & 12pt \Fkk & \ding{55} & 27.21 & 18.05 & 0.06 & 0.15 & 0.29 & \textbf{0.24} & \textbf{0.13} & \phantom{1}70.77\\
        \cmidrule{2-12}
        & 9pt \Fk & 9pt \Fk & \ding{55} & 27.37 & 17.55 & 0.08 & \underline{0.18} & \underline{0.31} & 0.40 & 0.25 & \phantom{1}40.45\\
        & 9pt \Fk & 12pt \Fkk & \ding{55} & 29.72 & 21.84 & 0.04 & 0.12 & 0.24 & 0.50 & 0.36 & \phantom{1}43.90\\
        \midrule
        \midrule
        \multirow{6}{*}{\rotatebox[origin=c]{90}{$\lambda_1 \neq \lambda_2$}} & 7pt \F & 7pt \F & 0 & 39.80 & 35.68 & 0.02 & \underline{0.06} & 0.13 & 0.86 & 0.86 & \phantom{1}40.71\\
        & 7pt \F & 8pt \F & 0 & 38.17 & 32.94 & 0.02 & \underline{0.06} & 0.15 & 0.86 & 0.86 & \phantom{1}\textbf{30.07}\\
        & 7pt \F & 12pt \Fkk & $\{-1.2, -0.6, 0\}$ & \textbf{29.52} & \textbf{21.06} & \textbf{0.05} & \textbf{0.12} & \textbf{0.25} & \textbf{0.36} & \underline{0.28} & \phantom{1}70.93\\
        \cmidrule{2-12}
        & 9pt \Fkk & 12pt \Fkk & \ding{55} & 29.80 & \underline{21.56} & \underline{0.04} & \textbf{0.12} & \underline{0.24} & \underline{0.39} & \textbf{0.27} & 220.05\\
        & 10pt \Fkk & 12pt \Fkk & \ding{55} & \underline{29.55} & 21.79 & \underline{0.04} & \textbf{0.12} & \textbf{0.25} & 0.43 & 0.32 & \phantom{1}\underline{30.13}\\
        \bottomrule
    \end{tabular}
    }
    \label{tab:phototourism_sacre_coeur_main_siftlg}
\end{table*}

\begin{table*}[ht]
    \centering
    \caption{Results on 2038 image pairs of 12 scenes from \ETH dataset, using GC-RANSAC framework, for SIFT+LG matches.}
    \setlength{\tabcolsep}{4.8pt}
    \resizebox{1.0\linewidth}{!}{
    \begin{tabular}{ c | r r c | c c c c c | c c | c}
        \toprule
        & & & & \multicolumn{7}{c}{GC-RANSAC - \ETH} \\
        \midrule
        & Minimal & Non-Minimal & Sample & AVG $(^\circ)$ $\downarrow$ & MED $(^\circ)$ $\downarrow$ & AUC@5 $\uparrow$ & @10 & @20 & AVG $\epsilon(\lambda)$ $\downarrow$ & MED $\epsilon(\lambda)$ $\downarrow$ & Time (ms) $\downarrow$ \\
        \midrule
        \multirow{11}{*}{\rotatebox[origin=c]{90}{$\lambda_1 = \lambda_2$}} & 7pt \F & 7pt \F & 0 & 25.16 & 16.80 & 0.08 & 0.17 & 0.31 & 0.88 & 0.88 & \phantom{1}120.22\\
        & 7pt \F & 8pt \F & 0 & 24.73 & 15.80 & 0.08 & 0.18 & 0.32 & 0.88 & 0.88 & \phantom{11}\underline{90.17}\\
        & 7pt \F & 9pt \Fk & 0 & 20.75 & \phantom{1}9.77 & 0.21 & 0.32 & 0.44 & 0.53 & 0.33 & \phantom{11}\textbf{80.91}\\
        & 7pt \F & 12pt \Fkk & 0 & 25.59 & 18.53 & 0.07 & 0.15 & 0.29 & 0.84 & 0.85 & \phantom{1}113.00\\
        & 7pt \F & 9pt \Fk & $\{-1.2, -0.6, 0\}$ & \underline{16.12} & \phantom{1}\underline{5.53} & \underline{0.26} & \underline{0.40} & \underline{0.54} & 0.31 & 0.17 & \phantom{1}120.11 \\
        & 7pt \F & 12pt \Fkk & $\{-1.2, -0.6, 0\}$ & 17.26 & \phantom{1}7.38 & 0.19 & 0.33 & 0.49 & 0.30 & 0.22 & \phantom{1}130.93\\
        \cmidrule{2-12}
        & 8pt \Fk & 9pt \Fk & \ding{55} & \textbf{15.36} & \phantom{1}\textbf{4.37} & \textbf{0.29} &\textbf{0.43} & \textbf{0.56} & \underline{0.29} & \underline{0.12} & \phantom{1}380.04 \\
        & 8pt \Fk & 12pt \Fkk & \ding{55} & 16.49 & \phantom{1}5.57 & 0.24 & 0.39 & 0.53 & \textbf{0.25} & \textbf{0.11} & \phantom{1}370.55 \\
        \cmidrule{2-12}
        & 9pt \Fk & 9pt \Fk & \ding{55} & 17.92 & \phantom{1}6.45 & 0.24 & 0.37 & 0.50 & 0.40 & 0.20 & \phantom{1}140.81\\
        & 9pt \Fk & 12pt \Fkk & \ding{55} & 19.33 & \phantom{1}9.43 & 0.16 & 0.29 & 0.45 & 0.48 & 0.32 & \phantom{1}150.43\\
        \midrule
        \midrule
        \multirow{6}{*}{\rotatebox[origin=c]{90}{$\lambda_1 \neq \lambda_2$}} & 7pt \F & 7pt \F & 0 & 32.47 & 24.35 & 0.04 & 0.10 & 0.21 & 0.86 & 0.84 & \phantom{1}203.38\\
        & 7pt \F & 8pt \F & 0 & 32.52 & 25.24 & 0.04 & 0.10 & 0.21 & 0.86 & 0.84 & \phantom{1}\underline{160.89}\\
        & 7pt \F & 12pt \Fkk & $\{-1.2, -0.6, 0\}$ & 20.96 & \phantom{1}9.99 & \underline{0.14} & 0.27 & 0.43 & \underline{0.39} & 0.30 & \phantom{1}420.77 \\
        \cmidrule{2-12}
        & 9pt \Fkk & 12pt \Fkk & \ding{55} & \underline{20.25} & \phantom{1}\underline{9.40} & \textbf{0.17} & \underline{0.30} & \underline{0.45} &\textbf{0.38} & \textbf{0.23} & 1380.09\\
        & 10pt \Fkk & 12pt \Fkk & \ding{55} & \textbf{19.95} & \phantom{1}\textbf{8.69} & \textbf{0.17} & \textbf{0.31} & \textbf{0.46} & \underline{0.39} & \underline{0.26} & \phantom{11}\textbf{80.12}\\
        \bottomrule
    \end{tabular}
    }
    \label{tab:eth3d_all_main_siftlg}
\end{table*}

\begin{table*}[ht]
    \centering
    \caption{Results of synthetic experiment for the \textbf{Wide-angle lens} setup on 12 scenes from the \ETH dataset, using the GC-RANSAC framework and SP+LG matches.}
    \setlength{\tabcolsep}{4.8pt}
    \resizebox{1.0\linewidth}{!}{
    \begin{tabular}{ c | r r c | c c c c c | c c | c}
        \toprule
        & & & & \multicolumn{7}{c}{GC-RANSAC - \ETH - \textbf{Wide-angle lens}} \\
        \midrule
        & Minimal & Non-Minimal & Sample & AVG $(^\circ)$ $\downarrow$ & MED $(^\circ)$ $\downarrow$ & AUC@5 $\uparrow$ & @10 & @20 & AVG $\epsilon(\lambda)$ $\downarrow$ & MED $\epsilon(\lambda)$ $\downarrow$ & Time (ms) $\downarrow$ \\
        \midrule
        \multirow{7}{*}{\rotatebox[origin=c]{90}{$\lambda_1 = \lambda_2$}} & 7pt \F & 9pt \Fk & $-1.5$ & \underline{15.85} & \underline{2.43} & \textbf{0.38} & \underline{0.52} & \textbf{0.64} & \underline{0.13} & \underline{0.08} & \phantom{1}\bf 170.05\\
        & 7pt \F & 12pt \Fkk & $-1.5$ & 16.06 & 2.80 & 0.34 & 0.51 & \underline{0.63} & \bf 0.09 & 0.09 & \phantom{1}\underline{180.12}\\
        \cmidrule{2-12}
        & 8pt \Fk & 9pt \Fk & \ding{55} & \bf 15.80 & \bf 2.36 & \bf 0.38 & \bf 0.53 & \bf 0.64 & 0.21 & \bf 0.07 & 1750.29\\
        & 8pt \Fk & 12pt \Fkk & \ding{55} & 16.15 & 2.87 & \underline{0.35} & 0.50 & 0.62 & 0.21 & \underline{0.08} & 1759.42\\
        \cmidrule{2-12}
        & 9pt \Fk & 9pt \Fk & \ding{55} & 18.94 & 4.20 & 0.30 & 0.44 & 0.56 & 0.38 & 0.15 & \phantom{1}603.35\\
        & 9pt \Fk & 12pt \Fkk & \ding{55} & 20.51 & 6.17 & 0.23 & 0.37 & 0.51 & 0.47 & 0.23 & \phantom{1}641.93\\
        \bottomrule
    \end{tabular}
    }
    \label{tab:eth_bias_wide}
\end{table*}

\begin{table*}[ht]
    \centering
    \caption{Results of synthetic experiment for the \textbf{Medium-angle lens} setup on 12 scenes from the \ETH dataset, using the GC-RANSAC framework and SP+LG matches.}
    \setlength{\tabcolsep}{4.8pt}
    \resizebox{1.0\linewidth}{!}{
    \begin{tabular}{ c | r r c | c c c c c | c c | c}
        \toprule
        & & & & \multicolumn{7}{c}{GC-RANSAC - \ETH - \textbf{Medium-angle lens} } \\
        \midrule
        & Minimal & Non-Minimal & Sample & AVG $(^\circ)$ $\downarrow$ & MED $(^\circ)$ $\downarrow$ & AUC@5 $\uparrow$ & @10 & @20 & AVG $\epsilon(\lambda)$ $\downarrow$ & MED $\epsilon(\lambda)$ $\downarrow$ & Time (ms) $\downarrow$ \\
        \midrule
        \multirow{7}{*}{\rotatebox[origin=c]{90}{$\lambda_1 = \lambda_2$}} & 7pt \F & 9pt \Fk & $-0.7$ & 16.64 & \underline{2.58} & \underline{0.37} & \underline{0.51} & \underline{0.62} & \underline{0.17} & \underline{0.08} & \phantom{1}\bf 205.99 \\
        & 7pt \F & 12pt \Fkk & $-0.7$ & \bf 15.98 & 3.11 & 0.32 & 0.49 & \underline{0.62} & \bf 0.10 & 0.09 & \phantom{1}\underline{220.47}\\
        \cmidrule{2-12}
        & 8pt \Fk & 9pt \Fk & \ding{55} & 16.33 & \bf 2.10 & \bf 0.39 & \bf 0.53 & \bf 0.63 & 0.22 & \bf 0.06 & 1933.12\\
        & 8pt \Fk & 12pt \Fkk & \ding{55} & \underline{16.10} & 3.02 & 0.35 & 0.49 & \underline{0.62} & 0.21 & \underline{0.08} & 1930.89\\
        \cmidrule{2-12}
        & 9pt \Fk & 9pt \Fk & \ding{55} & 18.56 & 3.79 & 0.32 & 0.45 & 0.57 & 0.30 & 0.14 & \phantom{1}625.29\\
        & 9pt \Fk & 12pt \Fkk & \ding{55} & 19.21 & 5.95 & 0.24 & 0.38 & 0.52 & 0.37 & 0.25 & \phantom{1}660.03\\
        \bottomrule
    \end{tabular}
    }
    \label{tab:eth3d_bias_medium}
\end{table*}

%
%
%
